\documentclass[11pt]{article}

\usepackage[preprint]{acl}
\usepackage{times}
\usepackage{latexsym}

\usepackage[T1]{fontenc}
\usepackage[utf8]{inputenc}

\usepackage{microtype}

\usepackage{inconsolata}

\usepackage{graphicx}

\usepackage{subcaption,booktabs,multirow,comment,graphicx} %nireak
\usepackage{tabularx}
\usepackage{xcolor,soul}
\sethlcolor{lightgray}

\usepackage{tcolorbox}
\usepackage{fancyvrb}
\usepackage{enumitem}
\usepackage{hyperref}

\newcommand{\flag}[1]{\includegraphics[height=9pt]{latex/flags/#1.pdf}}
\newcommand{\llm}[1]{\includegraphics[height=10pt]{latex/llm/#1.png}}
\newcommand{\llmb}[1]{\includegraphics[height=8pt]{latex/llm/#1.png}}
\newcommand{\topics}[1]{\includegraphics[height=8pt]{latex/topics/#1.png}}

\title{Why are all LLMs Obsessed with Japanese Culture? \\ On the Hidden Cultural and Regional Biases of LLMs 
        \thanks{This work has been accepted at EMNLP 2026 Findings}
        }

\author{Joseba Fernandez de Landa\textsuperscript{1} \quad Carla Perez-Almendros\textsuperscript{2} \quad Jose Camacho-Collados\textsuperscript{2} \\
  \textsuperscript{1} HiTZ Center - Ixa, University of the Basque Country EHU \\
  \textsuperscript{2} Cardiff University \\
  \texttt{joseba.fernandezdelanda@ehu.eus, CamachoColladosJ@cardiff.ac.uk} \\ 
  }

\begin{document}
\maketitle
\begin{abstract}
LLMs have limitations when it comes to cultural coverage and competence, and in some cases, show specific cultural biases. Although prior studies have examined the cultural capabilities of LLMs, none have specifically investigated their regional preferences in generic culture-related questions. In this work, we propose a new dataset based on a comprehensive taxonomy of Culture-Related Open Questions (CROQ), with questions available in 24 languages. We evaluate LLMs by prompting them to answer questions from CROQ and provide a sample location. The results show that, contrary to previous cultural bias work, LLMs show a clear tendency towards countries such as Japan in their answers. Moreover, our results show that when prompting in languages such as English or other high-resource ones, LLMs tend to provide more diverse outputs. Low-resource languages, on the other hand, show more inclinations towards answering questions highlighting countries for which the input language is an official language. Finally, we also investigate at which point of LLM training this cultural bias emerges, with our results suggesting that the first clear signs appear after supervised fine-tuning, and not during pre-training. Dataset available at \url{https://huggingface.co/datasets/HiTZ/CROQ}.
\end{abstract}

\section{Introduction}

Specific cultural knowledge, values and behaviours are often rooted in certain regions or countries. In this sense, we would expect that people from different geographical locations will have different answers to the same cultural questions. However, the behaviour of Large Language Models (LLMs) to these types of culture-related input remains an open question. Extensive work has been done towards unveiling cultural biases in LLMs \cite{shi-etal-2024-culturebank, NEURIPS2024_9a16935b, 10.1162/COLI.a.583, myung2024blend, chiu-etal-2025-culturalbench, arora-etal-2025-calmqa,ousidhoum-etal-2026-semeval}, mainly showing biases in favour of Western or Anglocentric viewpoints, especially United States and Europe \cite{lee-etal-2024-exploring-cross, NEURIPS2024_9a16935b, hasan2024nativqa, chiu-etal-2025-culturalbench, arora-etal-2025-calmqa, 10.1162/COLI.a.583}. 

\begin{figure}
    \centering    
    \includegraphics[trim=0 410 780 0, 
    clip, width=0.8\linewidth]{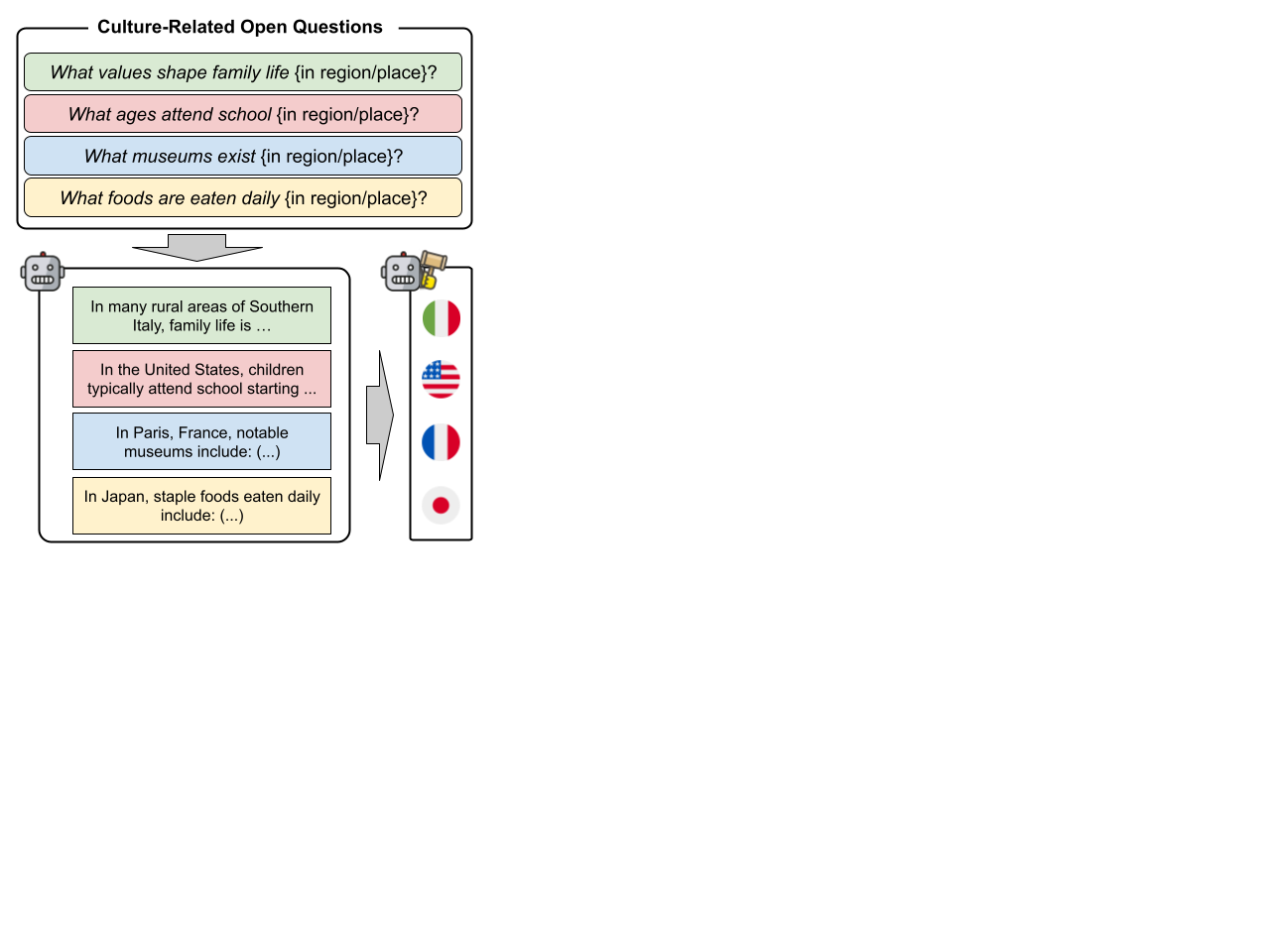}
    \caption{Framework for measuring cultural biases using CROQ dataset: (1) prompting models with cultural questions containing undefined locations, (2) collecting open responses from the specific model, and (3) identifying the referenced region(s) in the response.}
    \label{fig:proposal}
    \vspace{-12pt}
\end{figure}

Many of these works evaluate LLMs' responses against a set of predefined answers considered the gold standard \cite{romanou2024include, hasan2024nativqa, arora-etal-2025-calmqa, chiu-etal-2025-culturalbench}. Others evaluate the models' performance against human answers, covering a more or less wide spectrum of topics, cultures and countries/regions \cite{NEURIPS2024_9a16935b, shi-etal-2024-culturebank, myung2024blend, 10.1162/COLI.a.583}. In this work, we propose a novel framework to evaluate cultural regional biases in LLMs (Figure \ref{fig:proposal}), by prompting different models with open ambiguous cultural questions while masking the geographic information. This strategy requires the model not only to answer the cultural question, but also to geographically anchor that answer by selecting a location. By encouraging the models to link their responses to a specific region, our framework surfaces internal cultural priors and exposes inherent biases.

\noindent We address the following Research Questions (RQ):
\vspace{-10pt}

\begin{figure}[ht]
    \centering
\begin{tcolorbox}[left=-0.5em]
\vspace{-5pt}
\begin{tabular}{l} 
\textbf{1} What regional cultural biases do LLMs have? \\
\textbf{2} What is the impact of language in the bias? \\
\textbf{3} At what training stage does the bias emerge?
\end{tabular}
\vspace{-7pt}
    
\end{tcolorbox}

\end{figure}
\vspace{-10pt}

To support this analysis, we construct CROQ, a dataset of 31,680 open cultural questions spanning 24 languages, 11 major topics, and 66 subtopics. We then prompt LLMs to answer questions from CROQ and requiring them to choose a specific location, therefore unveiling hidden cultural and regional biases. Finally, we use an LLM as-a-judge on the generated answers of each model to extract the geographic reference chosen by each model, which is later manually validated. 

Our results indicate that regional preferences and model creativity are primarily determined by the input language. Notably, experiments across all models, languages and topics reveal a strong inclination for countries like Japan. Finally, further experiments suggest that these cultural and regional biases are induced predominantly during the post-training or instruction phase.

\section{Related Work}

Research on cultural knowledge and bias in LLMs is expanding rapidly \cite{adilazuarda-etal-2024-towards,liu-etal-2025-culturally,10.1162/COLI.a.14}, producing a variety of benchmarks, knowledge bases, and evaluation frameworks. While these resources provide important insights into multilingual and cultural capabilities, most are designed around closed-form answers and do not directly capture the leanings of models when responding to open-ended, ambiguous cultural questions.

Cultural knowledge is being benchmarked through question answering (QA) tasks. CulturalBench \cite{chiu-etal-2025-culturalbench} introduces human-authored multiple-choice questions covering 45 world regions and 17 topics, finding that LLMs fall far short of human performance, particularly in under-represented regions. Include \cite{romanou2024include} compiles nearly 200k exam-style questions in 44 languages, targeting multilingual and regional knowledge. BLEnD \cite{myung2024blend} emphasizes everyday cultural knowledge (e.g., food, family, leisure) across 16 countries and 13 languages, revealing strong disparities between high- and low-resource languages. Following this lead, SemEval-2026 Task 7 \cite{ousidhoum-etal-2026-semeval} expands this framework to include 17 additional language–culture pairs. Additionally, NativQA \cite{hasan2024nativqa} and CaLMQA \cite{arora-etal-2025-calmqa} use native-speaker based and culturally specific questions, respectively, across a wide range of topics, demonstrating heightened model failure rates for low-resource languages. Finally, Global MMLU \cite{singh-etal-2025-global} adapts MMLU benchmark by annotating questions for cultural sensitivity and expanding coverage to 42 languages. While these benchmarks reveal important weaknesses in factual cultural competence, their closed-answer format limits their ability to assess free-form generation.

Beyond QA, several works construct resources of cultural commonsense as knowledge bases. CANDLE \cite{10.1145/3543507.3583535} automatically extracts over one million assertions about food, rituals, and behaviours across 386 cultural groups, while CultureBank \cite{shi-etal-2024-culturebank} curates community-grounded descriptors from TikTok and Reddit to capture lived experiences. These knowledge bases are useful for training and grounding models, but their assertional structures are primarily designed for factual recall rather than for evaluating models’ tendencies in situations where multiple perspectives are valid.

A smaller but closely related line of work evaluates cultural adaptability and alignment. NormAd \cite{rao-etal-2025-normad} measures whether models can adapt to diverse cultural norms across 75 countries using situational vignettes, showing that even state-of-the-art systems perform far below human levels, especially in abstract or implicit norm settings. CulFiT \cite{feng-etal-2025-culfit} introduces a training paradigm and evaluation dataset (GlobalCultureQA) for open-ended cultural QA, using fine-grained reward modeling to enhance cultural alignment. Similarly, Makieval \cite{zhao2025makieval} proposes a framework for evaluating cultural awareness across 13 languages, 19 countries and regions, and six culturally salient topics, assessing open-ended generations by linking cultural entities in model outputs to Wikidata. While these approaches move beyond factual accuracy toward normative and generative evaluation, they often reduce outcomes to categorical labels (e.g., acceptable vs. unacceptable, aligned vs. misaligned) or optimize models toward a predefined target norm, rather than characterizing the full range of plausible cultural positions a model may express.

In contrast to prior work, our framework explicitly targets open cultural questions for which no single correct answer exists, prompting the model to choose among all regions in the world. By eliciting multiple model generations, we evaluate the leaning of LLMs along diverse cultural topics and languages. This complements existing factual and normative benchmarks by shifting the focus from accuracy to distributional analysis, enabling a more nuanced understanding of cultural bias and alignment in LLMs. Moreover, we attempt to cover a wide range of cultural topics, encompassing areas covered in previous work into a single framework, as we detail in the following section.%\footnote{See Appendix for more details on the coverage of cultural topics by our taxonomy and previous work.}}

\section{Dataset Generation: CROQ}
\label{sec:dataset}

We introduce Culture-Related Open Questions (CROQ), a multilingual dataset designed to uncover the cultural tendencies of LLMs through open-ended yet culturally grounded questions. CROQ consists of culturally relevant prompts spanning 11 broad topics and 66 finer-grained subtopics, constructed consistently across topics to preserve structural uniformity and regional ambiguity. The dataset has been translated into 24 languages, covering a wide range of language families, typologies, and resource levels. It contains 1,320 questions (20 per subtopic) which, after translation, results in a total of 31,680 questions (details about the translation and validation  in Appendix \ref{sec:translation-anex}).

\paragraph{Cultural Taxonomy.}
To make the open questions diverse and representative, we construct a taxonomy of 66 cultural subtopics grouped into 11 higher-level domains: \textit{\textbf{Beliefs}, Values, and Identity} (\topics{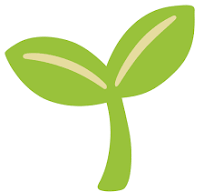}); \textit{\textbf{Social} Structure and Daily Life} (\topics{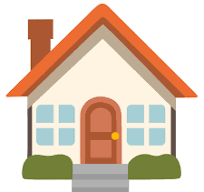}); \textit{Knowledge, Communication, and \textbf{Education}} (\topics{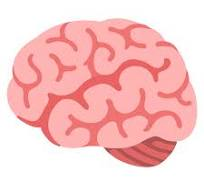}); \textit{Cultural Expression and the \textbf{Arts}} (\topics{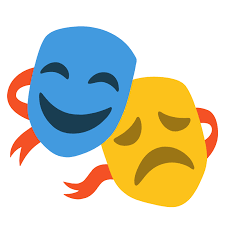}); \textit{\textbf{Food}, Drink, and Leisure} (\topics{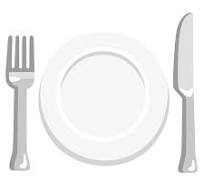}); \textit{\textbf{Geographic} Aspects} (\topics{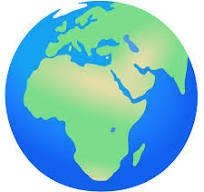}); \textit{\textbf{Political} Aspects} (\topics{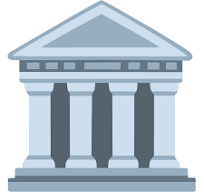}); \textit{\textbf{Health} and Wellness} (\topics{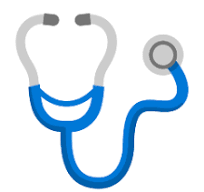}); \textit{\textbf{Media} and Entertainment} (\topics{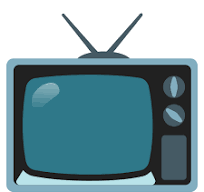}); \textit{\textbf{History}} (\topics{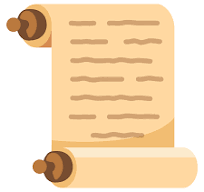}); and \textit{\textbf{Economy} and Industry} (\topics{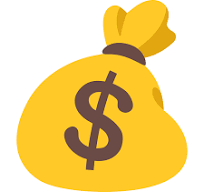}). This taxonomy is intended to capture a broad and representative spectrum of cultural topics, encompassing as many culturally relevant dimensions as possible. It was developed by aggregating the topics from the datasets discussed in the related work and expanding them into a more comprehensive collection (more detail in Appendix \ref{sec:taxonomy-anex}).

\paragraph{Open Question Generation.}
A central design choice in CROQ is the deliberate open-ended nature of each question. For each of the 66 cultural subtopics, we generate 20 questions, resulting in 1,320 queries in total. 
All questions follow the same basic pattern, avoiding direct references to specific countries or cultures, while remaining culturally meaningful. 
Questions are generated semi-automatically with \textit{GPT-5.1} and manually reviewed to remove repetitions or accidental location references. 
Example questions per topic are shown below.

%\vspace{-2pt}
\begin{figure}[ht]
    \centering
        \begin{tcolorbox}[left=0em]
\small
\vspace{-3pt}
\begin{tabular}{ll} 
 \textbf{\topics{t1} Beliefs:} & What legends explain the land?  \\ 
 \textbf{\topics{t2} Social:} & What is the role of neighbors? \\ 
 \textbf{\topics{t3} Education:} & What subjects are most valued? \\
 \textbf{\topics{t4} Arts:} & What traditional dances exist? \\ 
 \textbf{\topics{t5} Food:} & What foods are for daily meal? \\ 
 \textbf{\topics{t6} Geography:} & What rivers influence settlements? \\ 
 \textbf{\topics{t7} Politics:}  & What political protests are common? \\ 
 \textbf{\topics{t8} Health:} & What exercise routines are common? \\ 
 \textbf{\topics{t9} Media:} & What film industries exist? \\ 
 \textbf{\topics{t10} History:} & What historical tours exist? \\ 
 \textbf{\topics{t11} Economy:} & What commodity markets exist?
\end{tabular}
\vspace{-5pt}

    \end{tcolorbox}
    \vspace{-8pt}

\label{fig:question-examples}
\end{figure}
\vspace{-3pt}

\paragraph{Multilingual Coverage.} 
The dataset was initially constructed in English and then automatically translated and post-edited into 23 additional languages to ensure broad coverage across speaker populations, resource levels, and language families (full list in Appendix \ref{sec:lang_anex}). As a starting point, we included the 13 languages from the BLEND dataset \cite{myung2024blend}, which range from low-resource (Amharic -am-, Assamese -as-, Azerbaijani -az-, Hausa -ha- and Sundanese -su-) to mid- and high-resource (Greek -el-, Indonesian -id-, Korean -ko-, Persian -fa-, Arabic -ar-, Chinese -zh-, Spanish -es-, English -en-) languages and provide wide geographical coverage.

To further enhance diversity, we added several of the world’s most widely spoken languages with varying degrees of representation in CommonCrawl (Hindi -hi-, French -fr-, Bengali -bn-, Portuguese -pt-, Russian -ru-, German -de-, Japanese -ja-, Swahili -sw-). Additionally, we included all co-official but low-resource languages of Spain (Basque -eu-, Galician -gl-, Catalan -ca-)\footnote{We use ISO 639-1:2002 codes for each language.}, which have relatively small speaker populations.

\section{Evaluation Framework}
\label{sec:method}

We propose a novel method to uncover cultural biases in LLMs. We achieve this by prompting models with CROQ and requiring them to explicitly select a country or region. Subsequently, a secondary model is employed to systematically extract the geographic information inferred from the responses, allowing us to analyze potential cultural biases (see Figure \ref{fig:proposal}).

\subsection{Regional Question Grounding}

To probe cultural priors in LLMs, we leverage the set of culture-related open questions (Section \ref{sec:dataset}) and require the model to openly generate responses to these questions. Each question contains a locational placeholder (\textit{{in region/place}}) that the model must implicitly resolve when producing an answer. We further prompt the model to be brief and to select a specific location (example below), thereby requiring it to tacitly choose a region or cultural background to ground its response:

 %\vspace{-4pt}
\begin{figure}[ht]
    \centering
        \begin{tcolorbox}[left=.5em,top=.2em,bottom=.1em]
%\begin{Verbatim}[fontsize=\small]
\large
\textit{What values shape family life} \\ \{in region/place\}?\\
\\ Be brief. Choose yourself the place.
%\end{Verbatim}
    \end{tcolorbox}
    \vspace{-12pt}
    %\caption{Input example given to the models.}
    \label{fig:asking_prompt}
\end{figure}
 %\vspace{-4pt}

Each LLM evaluated receives the same set of 1,320 open questions in different languages (see Section \ref{sec:dataset} for more details). The prompt used to guide the model varies by language. To reduce prompt variability, all questions follow a standardized template, and no explicit regional cues are provided. This ensures that any regional or cultural assumptions arise from the model’s internal priors rather than prompt design.

\subsection{Processing Model Responses}
\label{sec:processing_model_responses}

To process the open-ended answers generated by the LLMs, we employ a secondary model as-a-judge to interpret the responses and extract the relevant information required to associate each question with a specific set of countries. This process is designed to uncover potential cultural biases by tasking the judge model with identifying up to five regions or places mentioned or implied in each response; if no such information can be reliably determined, the model is prompted to explicitly indicate that no inference is possible (non-answered responses). Following an evaluation of various prompting strategies, we selected the optimal approach, which achieved an accuracy of 98\% over 264 evaluation items, the details of which, including the prompts and additional technical information, can be found in Appendix \ref{sec:eval_det_anex}.

\subsection{Comparison of Frontier LLMs}
\label{sec:compared_llms}

For our evaluation, we compare a wide range of multilingual LLMs, including both frontier closed- and open-weight models. All models are accessed through OpenRouter (temp.=0.7 and top-p=0.9) and evaluated across the 24 languages in our dataset. 
The models include: \llmb{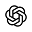} gpt-4o-mini, \llmb{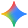} gemini-2.5-flash, \llmb{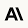} claude-3.5-haiku, \llmb{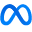} llama-4-maverick, \llmb{cohere} command-r-08-2024, \llmb{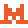} magistral-small-2506, \llmb{deepseek} deepseek-v3.2-exp, \llmb{qwen} qwen3-next-80b-a3b-instruct.

\setlength{\tabcolsep}{3.8pt}%{4.5pt}

\begin{table*}[ht!]
\centering
\small
\renewcommand{\arraystretch}{0.8}
\begin{tabular}{@{}lrrrrrrrrrrrrrr|rr@{}}

\toprule
\textbf{Model} & \textbf{Own} (\%) & \textbf{NA} & \flag{JP} & \flag{US} & \flag{IN} & \flag{CN} & \flag{FR} & \flag{IT} & \flag{GB} & \flag{DE} & \flag{MX} & \flag{BR} & \flag{RU} & \flag{EG}  & \textbf{Div}& \textbf{Ent}\\
\midrule
\llm{openai} GPT        & 24,763 (.78)& 1,136 & 811   & \textbf{944} & 451 & 237 & 289 & 265 & 128 & 150 & 134 & 234 & 53 & 87 & 82 & 0.40 \\
\llm{gemini} Gemini     & 20,172 (.64)& 1,853 & \textbf{1,493} & \textbf{1,493} & 673 & 473 & 439 & 344 & 310 & 271 & 164 & 125 & 75 & 240 & 113 & 0.50\\
\llm{anthropic} Claude  & 20,585 (.65)& 2,063 & \textbf{1,601} & 1,200 & 510 & 560 & 210 & 218 & 138 & 172 & 340 & 349 & 122 & 141 & 95 & 0.48\\
\llm{llama} Llama       & 20,775 (.66)& 887   & \textbf{2,701} & 1,074 & 453 & 524 & 587 & 265 & 190 & 173 & 75 & 222 & 61 & 132 & 102 & 0.49\\
\llm{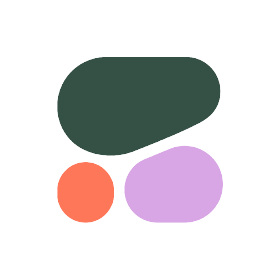} Command-r  & 13,707 (.43)& 4,815 & 936   & \textbf{2,064} & 567 & 512 & 504 & 389 & 515 & 309 & 209 & 259 & 113 & 142 & \textbf{120} & \textbf{0.59}\\
\llm{mistral} Magistral & 23,040 (.73)& 2,485 & \textbf{1,754} & 1,254 & 496 & 560 & 632 & 392 & 264 & 390 & 315 & 339 & 163 & 211 & 88 & 0.47\\
\llm{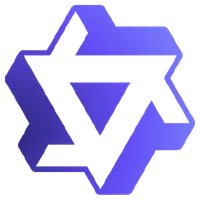} Qwen         & 22,572 (.71)& 1,502 & \textbf{1,861} & 483 & 249 & 434 & 191 & 170 & 60 & 158 & 222 & 67 & 58 & 60 & 79 & 0.40\\
\llm{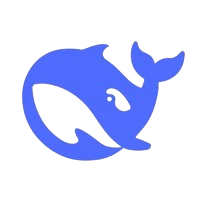} DeepSeek & 21,907 (.69)& 877   & \textbf{2,104} & 1,006 & 541 & 608 & 351 & 415 & 390 & 499 & 308 & 205 & 110 & 267 & 116 & 0.52\\
\bottomrule
\end{tabular}
\caption{Frontier Model outputs for 24 languages. \textbf{Own} counts references to countries in which the language is an official language. \textbf{NA} denotes missing responses. Top referenced countries are Japan, USA, India, China, France, Italy, United Kingdom, Germany, Mexico, Brazil, Russia, Egypt. \textbf{Div} and \textbf{Ent} represent the average for diversity and entropy by model. Bold values indicate the highest count for each model, excluding self-references, as well as the highest diversity and entropy values (two right-most columns).}
\vspace{-4pt}
\label{tab:frontier_out_bymodel}
\end{table*}

\setlength{\tabcolsep}{6pt}

\subsection{Analysis Metrics}
\label{sec:analysis_metrics}

For our subsequent analysis, we devise three metrics that can help analyse the output of each LLM.

\paragraph{Diversity.}
We define diversity as the number of distinct countries or regions referenced across all model outputs. Higher values indicate broader geographic coverage of cultural references.

\paragraph{Entropy.}
We compute normalized entropy over the distribution of referenced places to capture how evenly cultural references are distributed, with higher scores indicating greater balance.

\paragraph{Raw counts.}
We additionally report country or region frequency counts to provide complementary and interpretable evidence.

\section{RQ1: What Regional Cultural Biases Do LLMs Have?}

\paragraph{Models favor input language countries:}
Table~\ref{tab:frontier_out_bymodel} summarizes the regional distribution of country references across models for the 24 languages. Across all models, references overwhelmingly concentrate on \textit{own} countries (ranging from 43\% for Command-r to 78\% for GPT), confirming a strong alignment between the output language and its associated regions. This pattern is consistent and robust, suggesting that language choice alone strongly constrains the cultural priors adopted by the models. 

\paragraph{Exogenous references are dominated by a small set of regions:}
The data reveal that for non-own country references, all models display a similar pattern of global salience. On average, Japan (preferred country for six out of the eight evaluated models) and the United States are the most frequently referenced nations, followed by India, China, and France. References to other countries are considerably less frequent and vary by model.

\paragraph{Differences among models:}
 Although the distribution across mentioned exogenous countries (Japan, United States, India, and China) is similar, some differences emerge. Five models favor Japan, while two emphasize the US, and one is tied. Command-R stands out as the most US-biased model and the one most likely to refuse an answer, producing in general the most diverse outputs with the highest entropy. Command-R, DeepSeek, and Gemini have higher output diversity (over 113 different countries mentioned), whereas Qwen, GPT, and Magistral show lower diversity and entropy (fewer than 88 countries mentioned in their outputs -- more details on individual model responses in Appendix \ref{sec:div_anex}).

\paragraph{Topic Analysis:}
Table \ref{fig:to_by_taxonomy} presents the top countries mentioned by the models across the 11 general topics in our dataset. Japan and the US remain salient across all categories, followed by other frequently mentioned countries. However, some variation occurs depending on the topic which may reflect global trends. For example, in \textit{Geography} and \textit{Economy}, there is a noticeable shift in prominence between the US and Japan, where US takes the lead in number of references, and \textit{Politics} and \textit{History} are the only two topics where Japan is relegated to third and fourth positions. Countries such as Greece, China and France stand out in very specific topics, namely \textit{Beliefs}, \textit{Politics} and \textit{History}, respectively. Or other countries such as South Korea, which is barely mentioned in most topics, emerge as the third most mentioned country in \textit{Media and Entertainment}. Mentions of other countries also fluctuate across topics, but their overall counts remain substantially lower (data in Appendix \ref{sec:top_anex}).

\begin{table}[h!]
\centering
%\small
\begin{tabular}{@{}l l@{}}
\toprule
 \textbf{Topic} & \textbf{Top Countries} \\
\midrule
 \topics{t1} 
    1. Beliefs
    & \flag{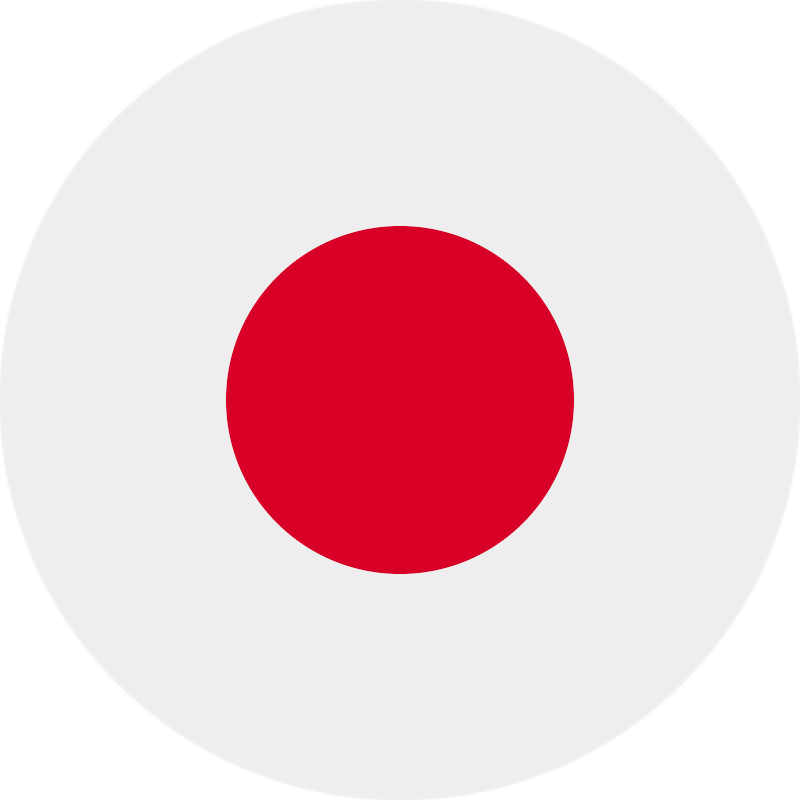} \flag{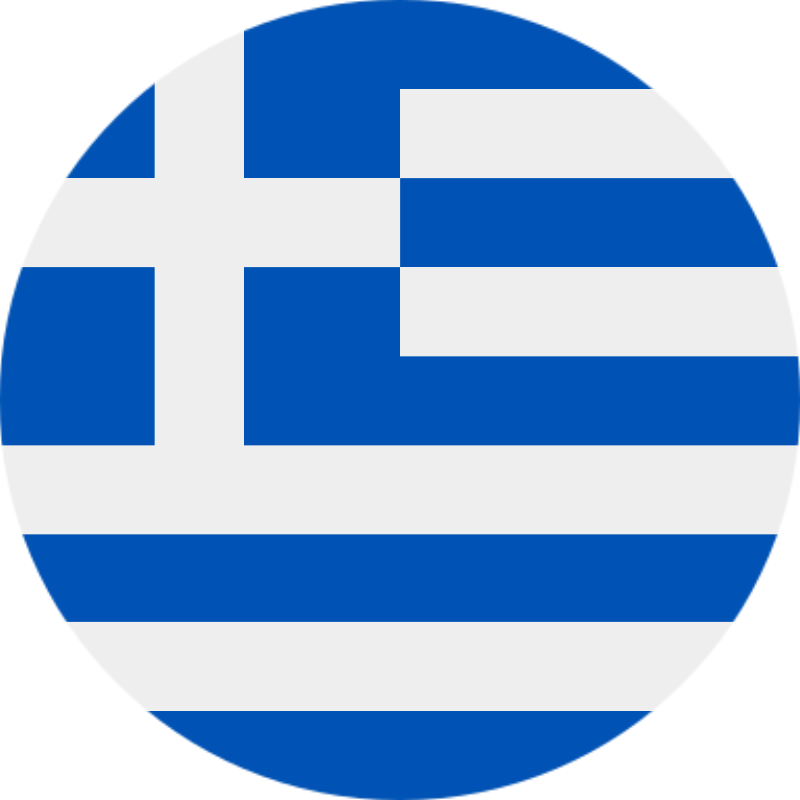} \flag{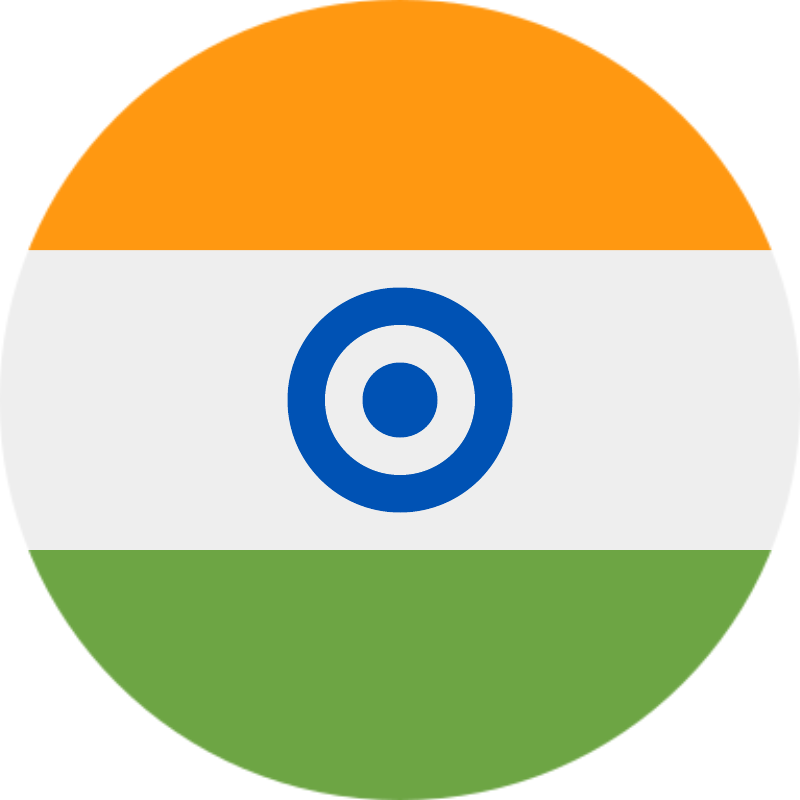} \flag{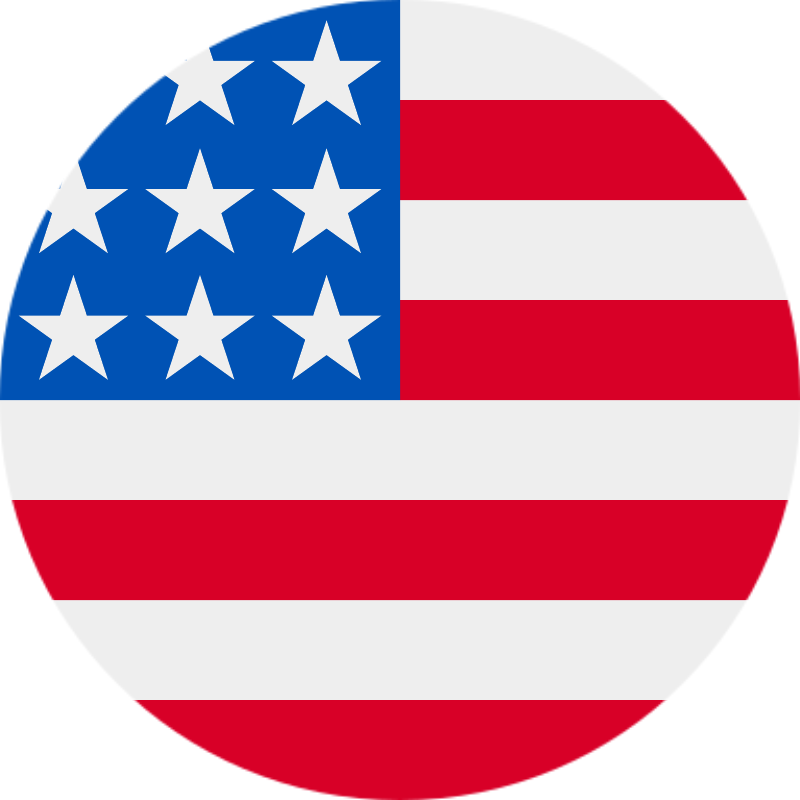} \flag{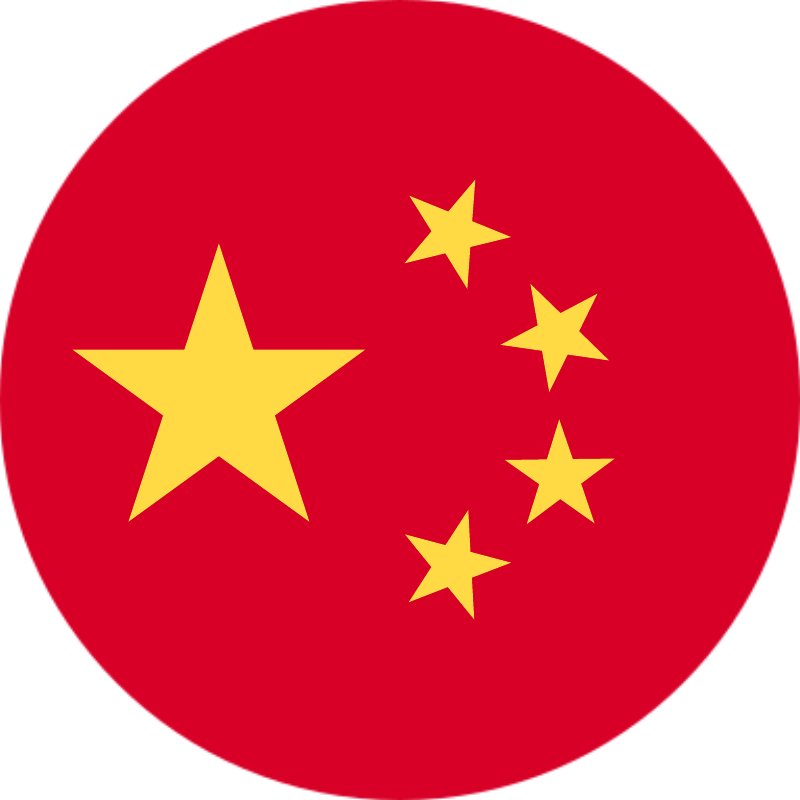} \flag{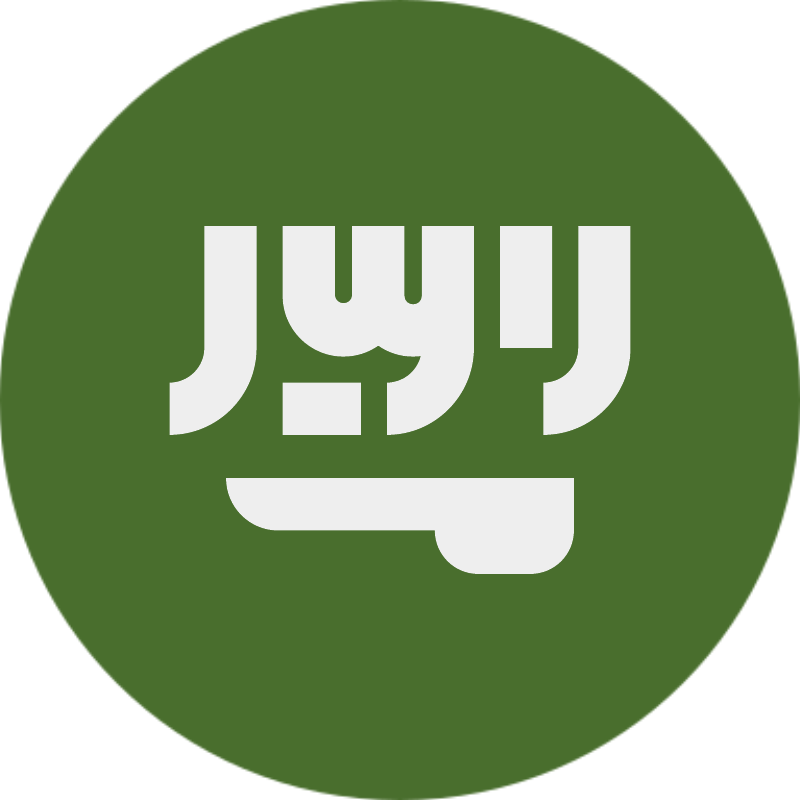} \flag{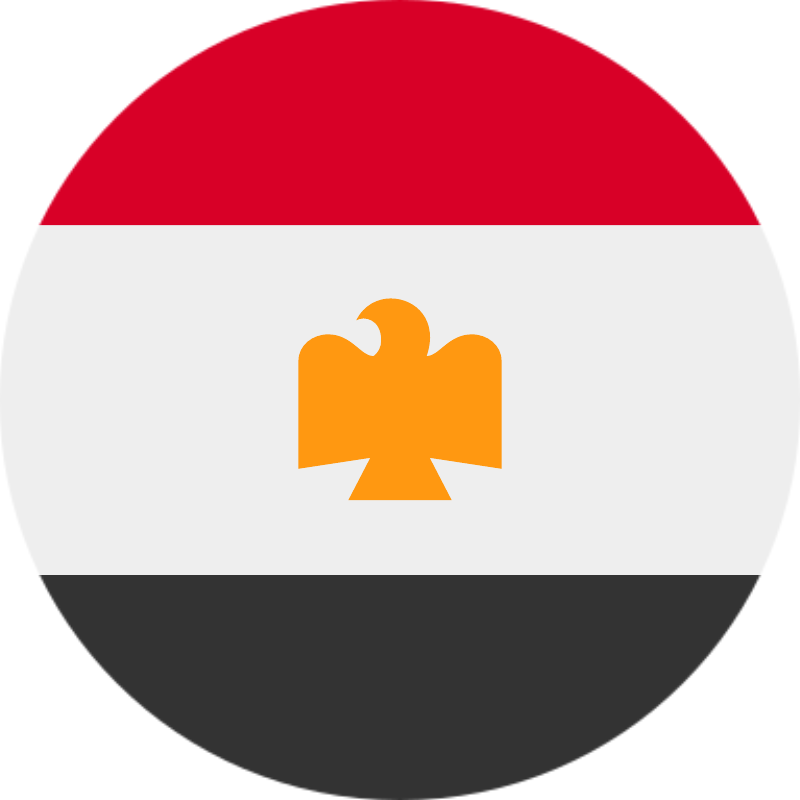} \flag{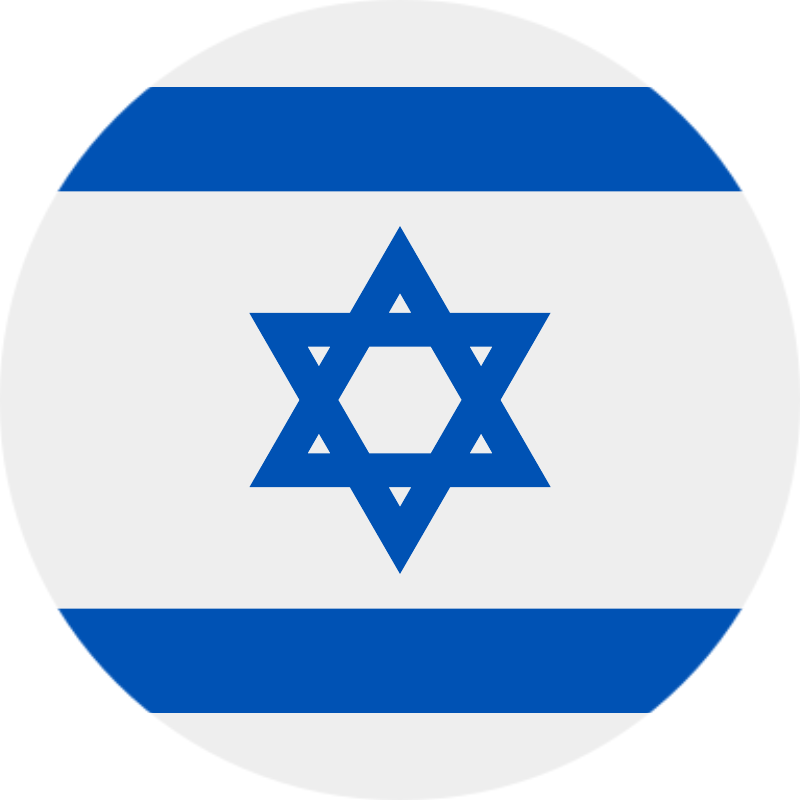} \flag{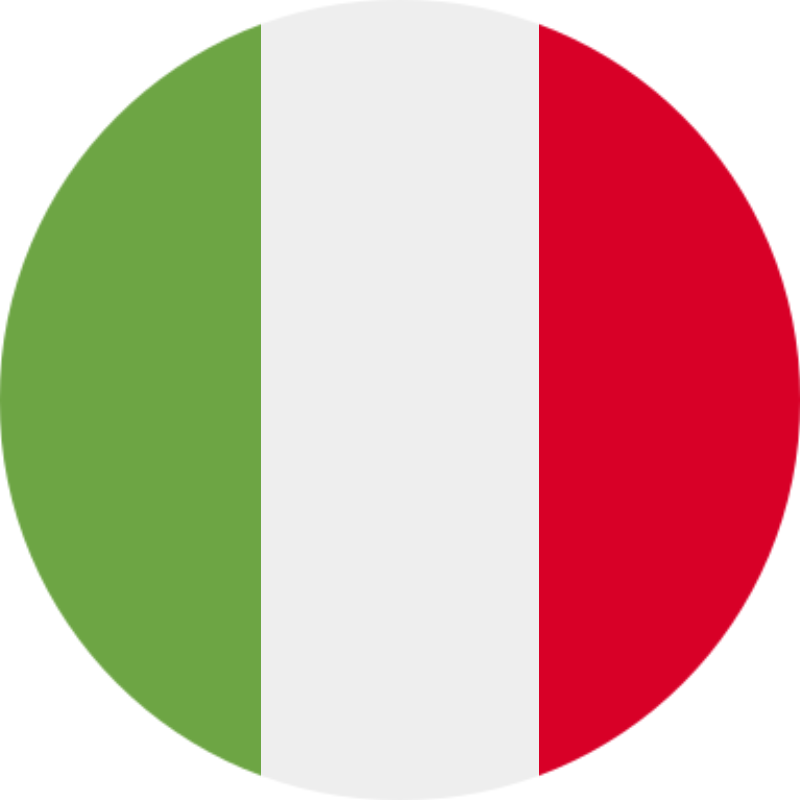} \flag{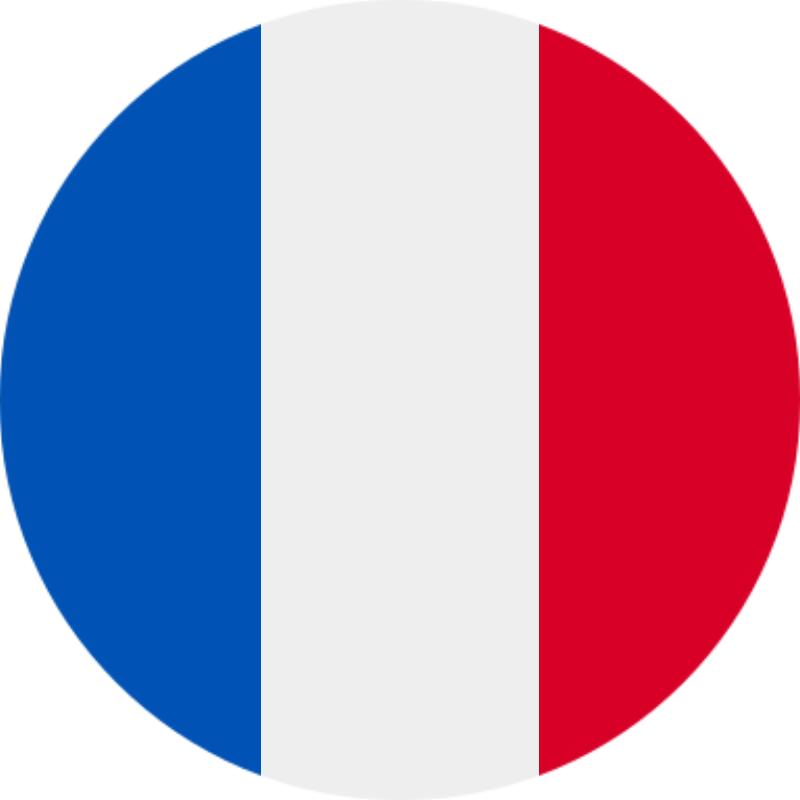} \\
 \topics{t2} 
    2. Social
    & \flag{jp} \flag{us} \flag{il} \flag{it} \flag{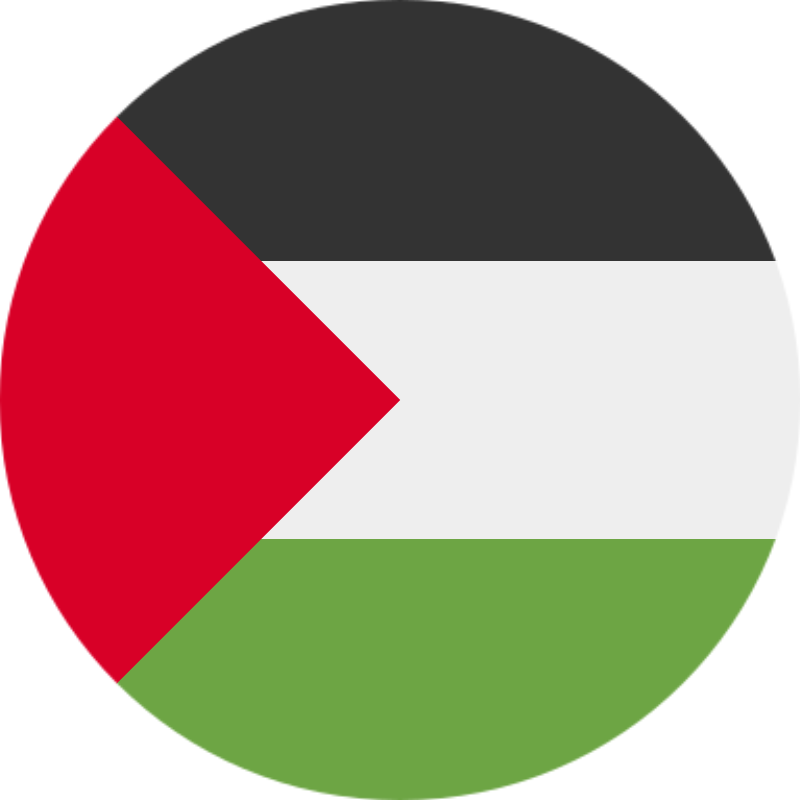} \flag{cn} \flag{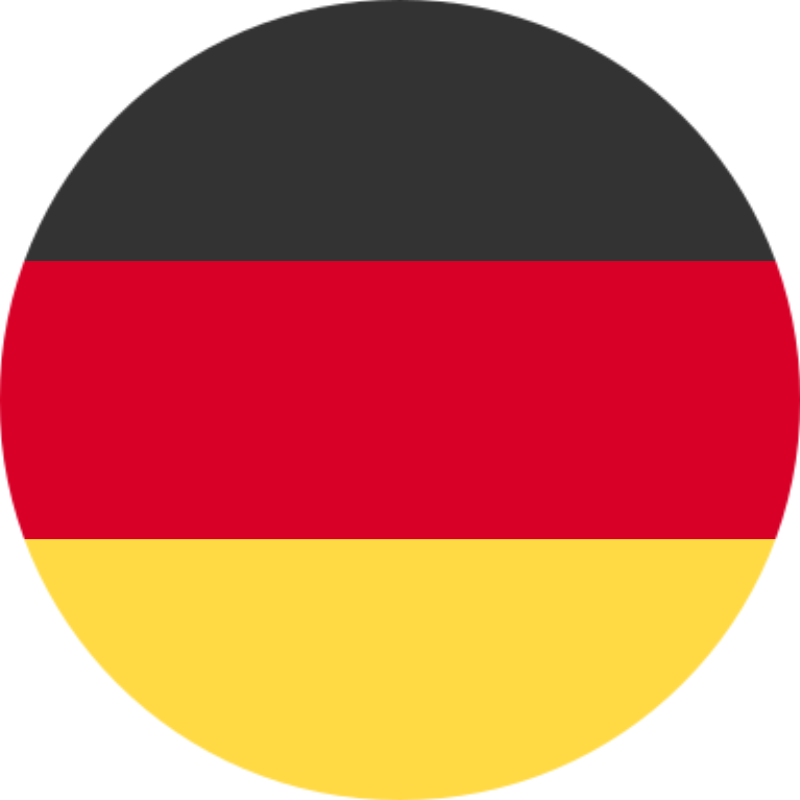} \flag{fr} \flag{in} \flag{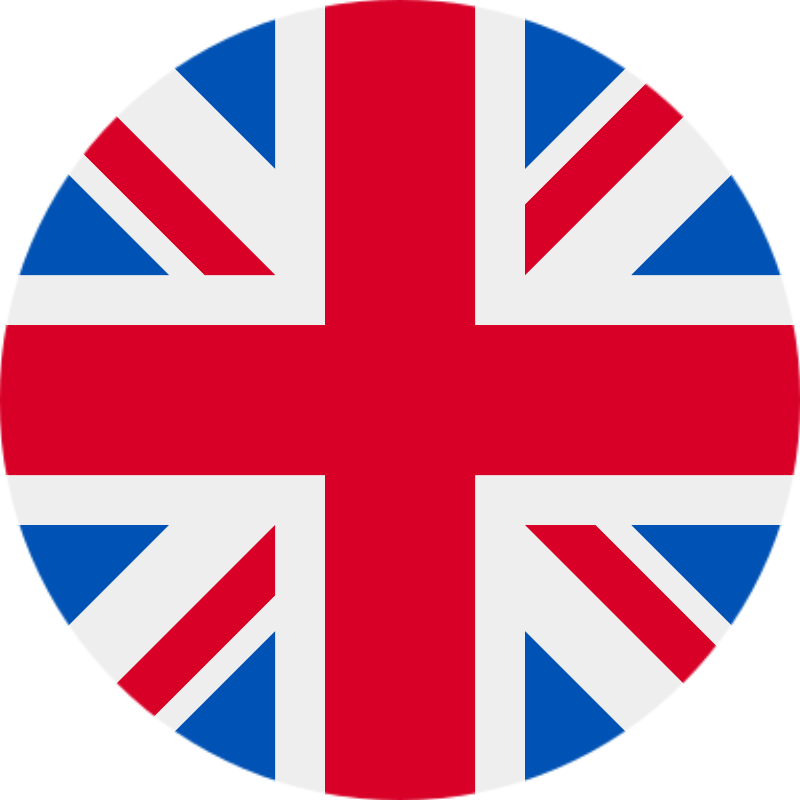} \\
 \topics{t3}
    3. Education 
    & \flag{jp} \flag{us} \flag{gr} \flag{cn} \flag{fr} \flag{in} \flag{de} \flag{eg} \flag{it} \flag{gb}\\
 \topics{t4} 
    4. Arts 
    & \flag{jp} \flag{us} \flag{in} \flag{fr} \flag{it} \flag{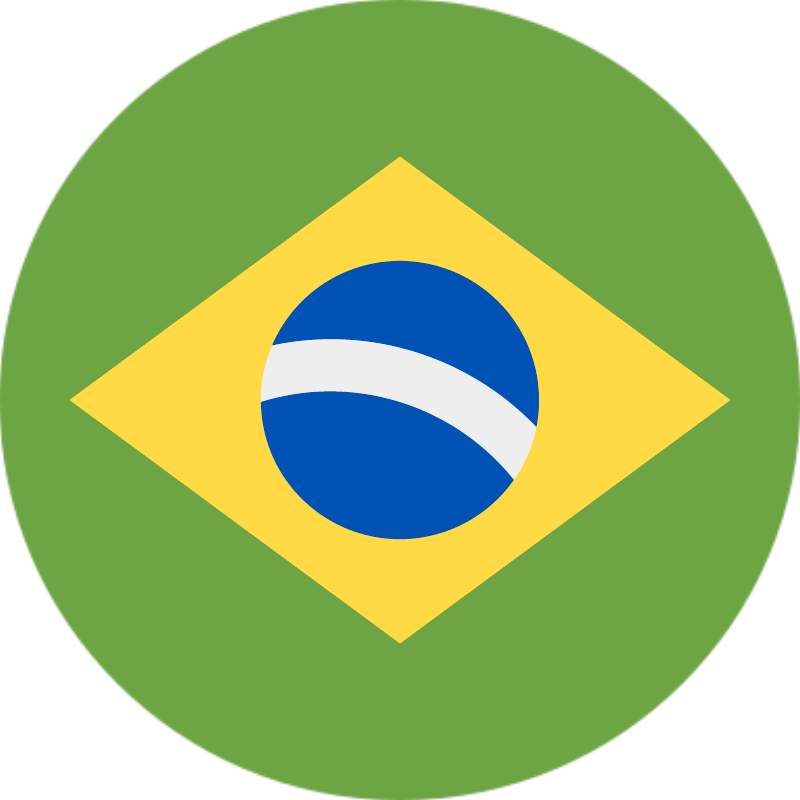} \flag{cn} \flag{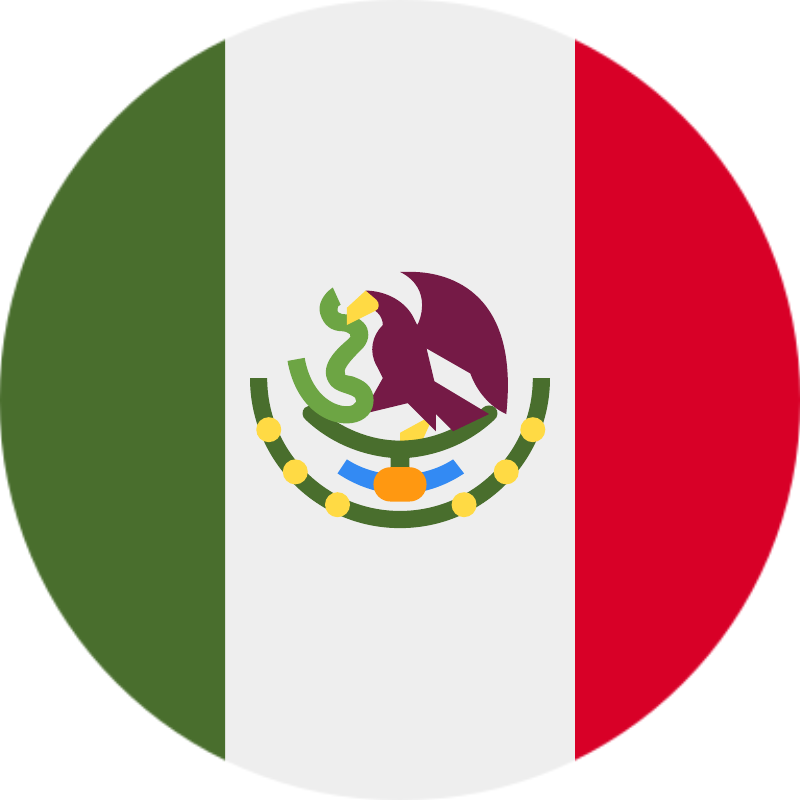} \flag{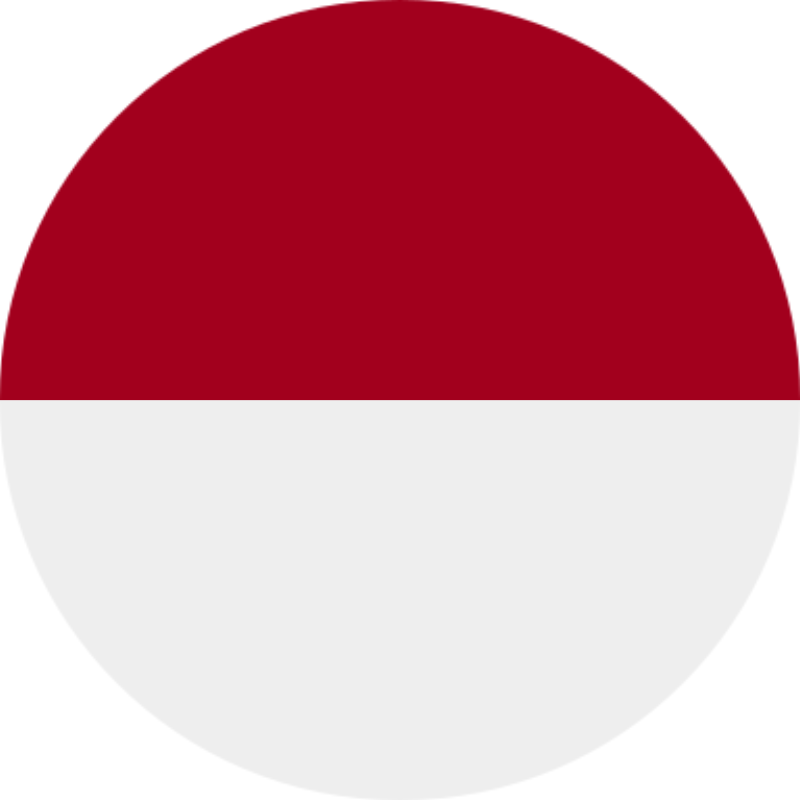} \flag{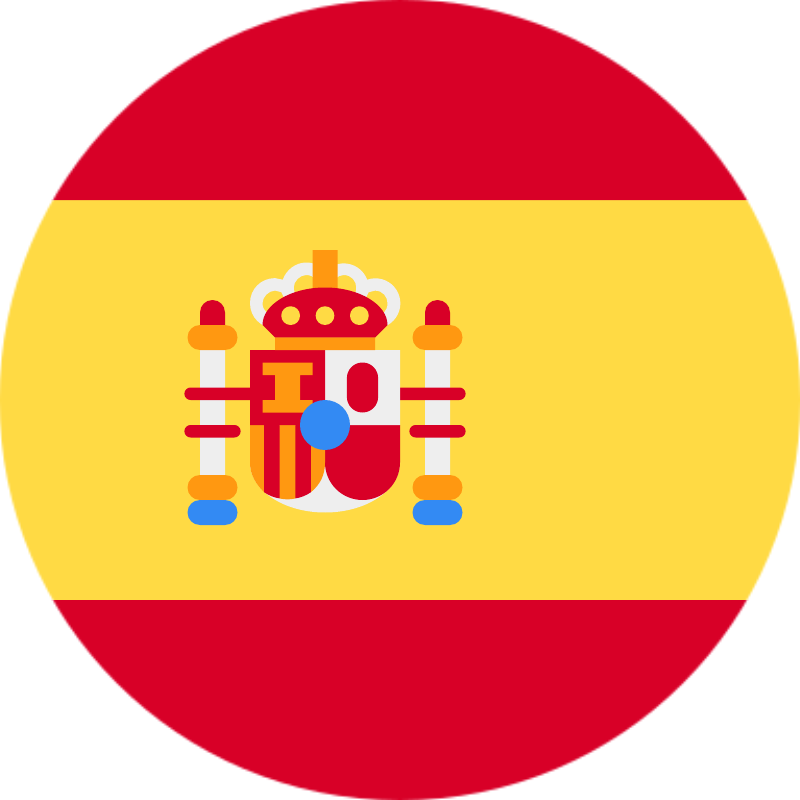}\\
 \topics{t5} 
    5. Food
    & \flag{jp} \flag{us} \flag{in} \flag{mx} \flag{cn} \flag{it} \flag{br} \flag{es} \flag{fr} \flag{id}\\
 \topics{t6} 
    6. Geography
    & \flag{us} \flag{jp} \flag{in} \flag{cn} \flag{fr} \flag{it} \flag{mx} \flag{br} \flag{de} \flag{il}\\
 \topics{t7} 
    7. Politics
    & \flag{us} \flag{cn} \flag{jp} \flag{fr} \flag{gb} \flag{de} \flag{in} \flag{sa} \flag{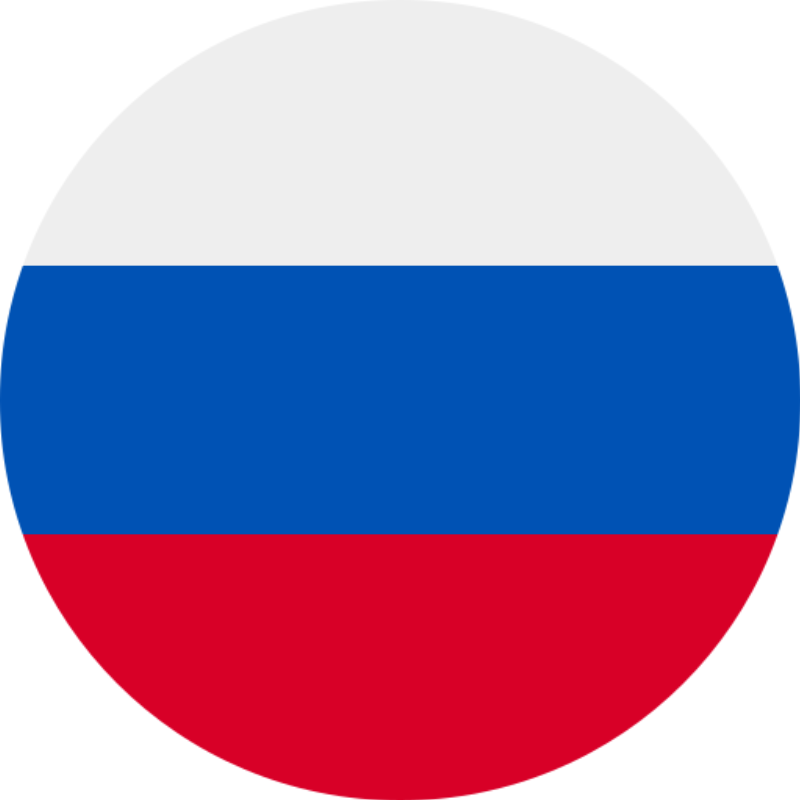} \flag{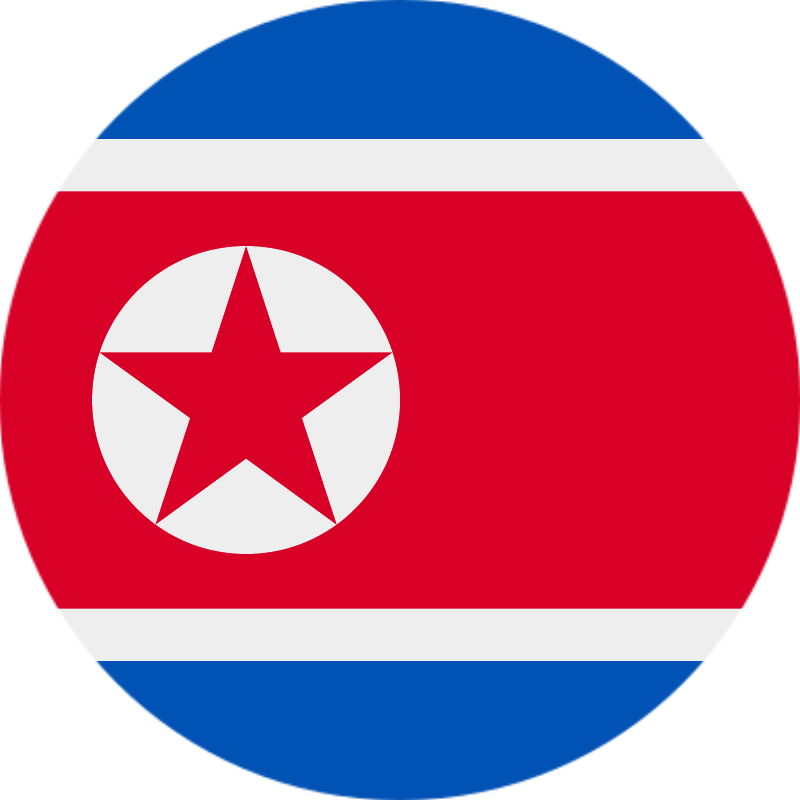}\\ 
 \topics{t8} 
    8. Health
    & \flag{jp} \flag{us} \flag{cn} \flag{in} \flag{id} \flag{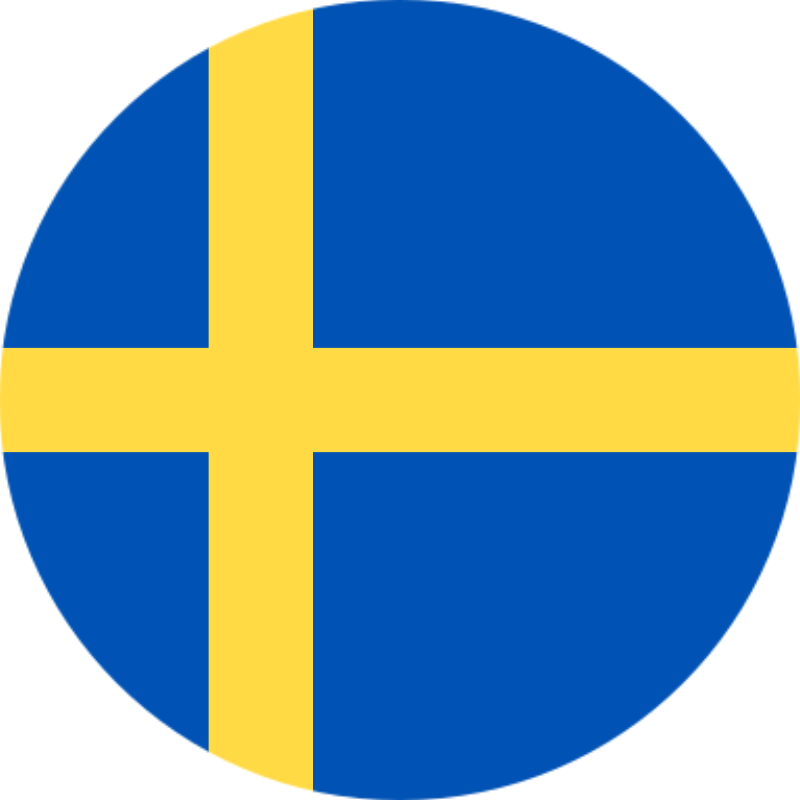} \flag{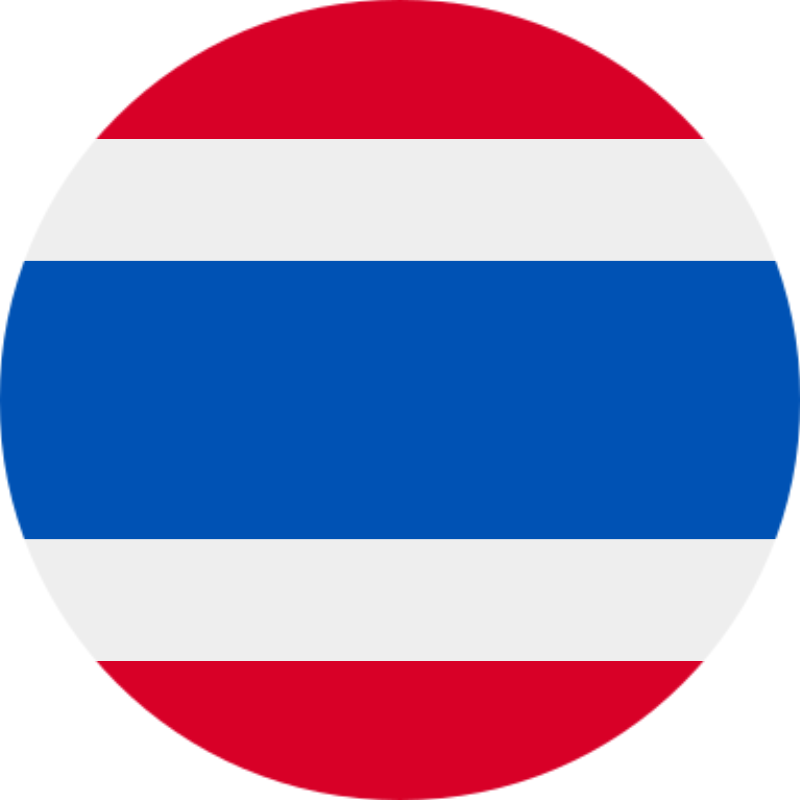} \flag{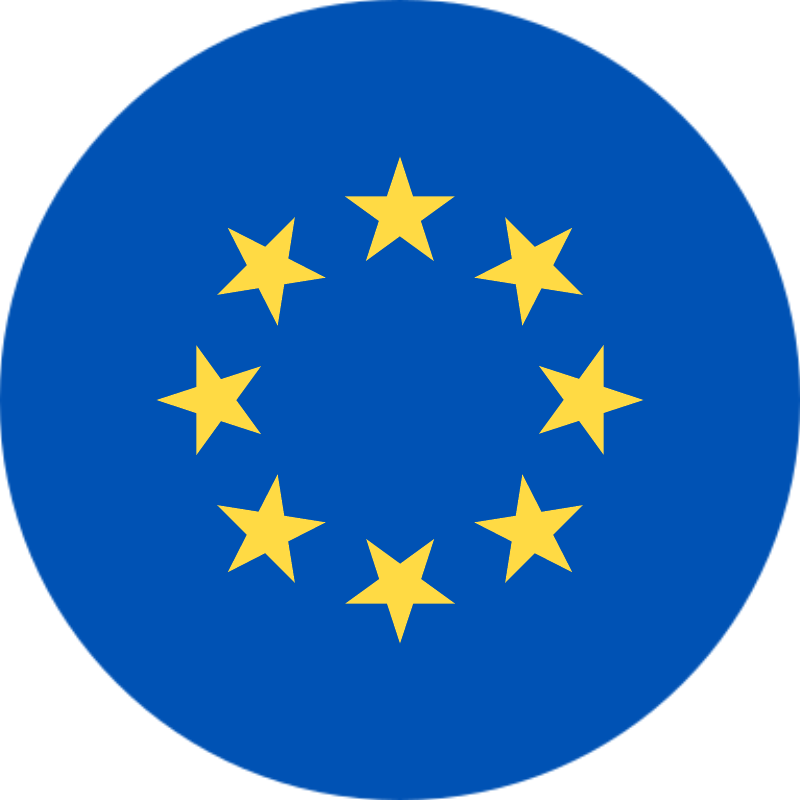} \flag{br} \flag{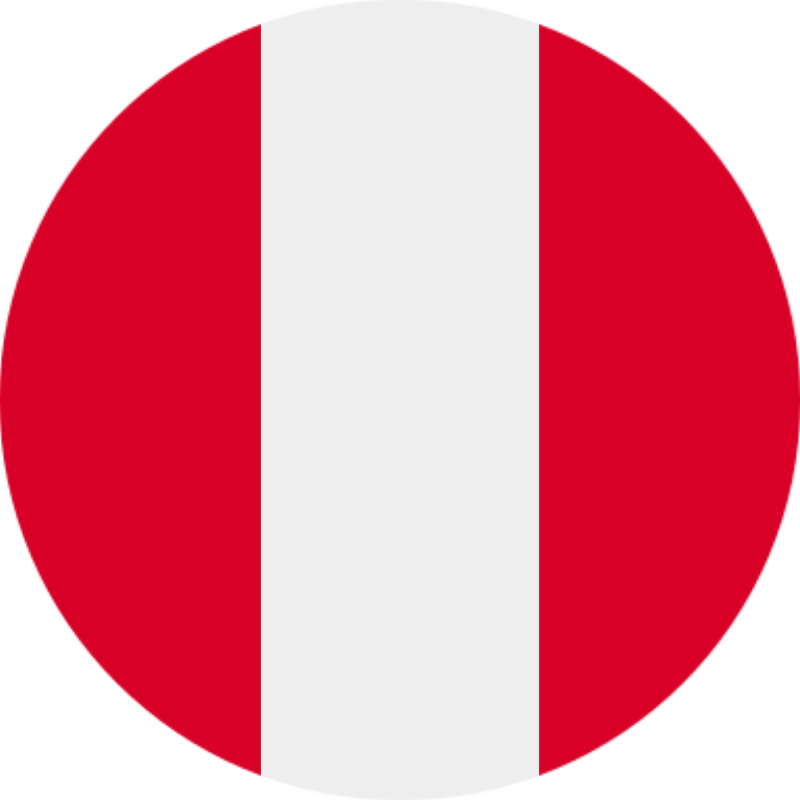}\\
 \topics{t9} 
    9. Media
    & \flag{jp} \flag{us} \flag{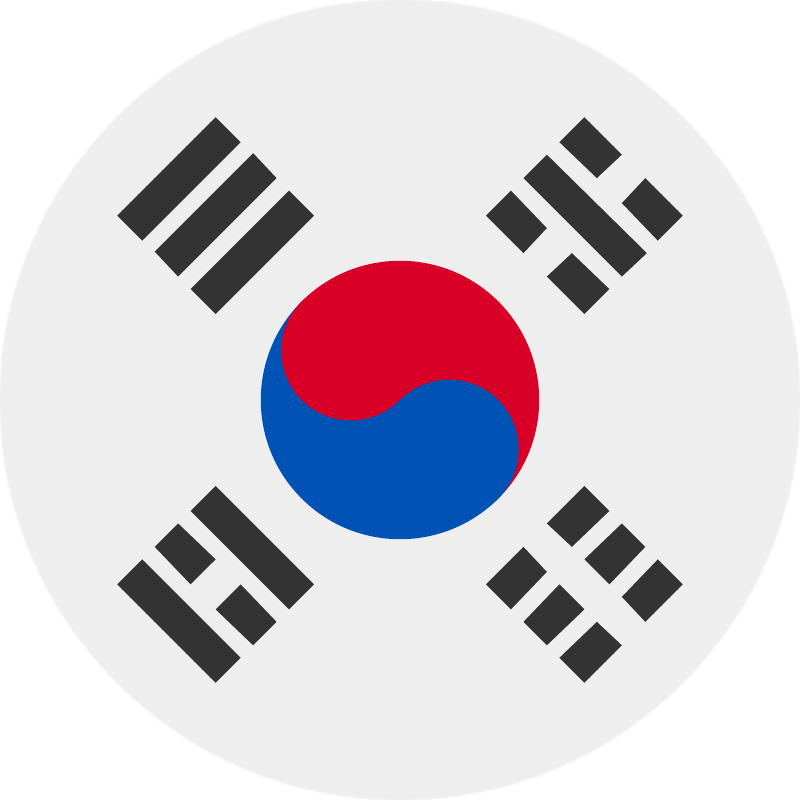} \flag{gb} \flag{cn} \flag{fr} \flag{de} \flag{br} \flag{in} \flag{eu}\\
 \topics{t10} 
    10. History 
    & \flag{us} \flag{fr} \flag{in} \flag{jp} \flag{it} \flag{gb} \flag{cn} \flag{gr} \flag{de} \flag{eg}\\
 \topics{t11} 
    11. Economy
    & \flag{us} \flag{jp} \flag{cn} \flag{eu} \flag{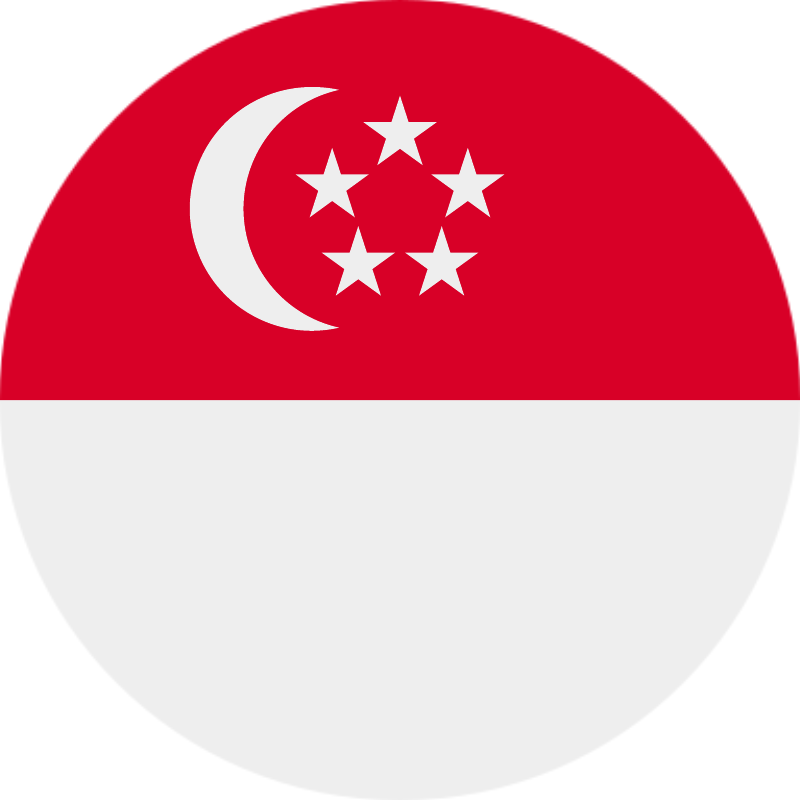} \flag{de} \flag{in} \flag{kr} \flag{gb} \flag{fr}  \\
\bottomrule
\end{tabular}
    \caption{Top 10 countries most frequently referenced by all Frontier API models for each topic in the taxonomy. 24 languages, excluding own country mentions.}
    \label{fig:to_by_taxonomy}
\end{table}

\paragraph{Summary of findings and discussion.} 
All eight frontier models exhibit a clear bias toward languages referring to their own regions of origin. When associations to own language-region pairs are isolated, this bias becomes more pronounced. Consequently, all models consistently favour Japan or the United States, while regions such as India and China receive comparatively less emphasis, and references to other regions are negligible. This pattern suggests an uneven regional representation in frontier model outputs, with a strong concentration on a small set of dominant regions.

\section{RQ2: What is the Impact of Language in Regional Cultural Bias?}

\paragraph{Salient regions are overrepresented across all languages:}
As established in the previous section, models strongly associate languages with their primary regions or countries of origin. To quantify this effect and explore its variation, we conducted a per-language analysis. Table~\ref{tab:frontier_out_lang} shows that beyond self-referential patterns, the models exhibit a stable hierarchy of non-own country references. Japan and the United States are the most frequently referenced countries across languages. Notably, these references appear even in languages with limited cultural or geographic ties to these countries. This finding aligns with prior observations that LLMs tend to overrepresent culturally salient entities.

\setlength{\tabcolsep}{4.5pt}

\begin{table}[h!]
\centering
\small
\begin{tabular}{@{}l rrrrr|rr@{}}
\toprule
 & \textbf{Own} & \flag{jp} & \flag{us} & \flag{in} & \flag{cn} & \textbf{Div} & \textbf{Ent} 
 \\ 
\cmidrule{2-8}
en & \textcolor{olive}{31\%}& \textbf{2,893} & * & * & 367 & \textcolor{olive}{157} & \textcolor{olive}{0.69}\\ 
zh & \textcolor{olive}{48\%}& \textbf{2,064} & 1460 & 642 & * & 142 & 0.57\\ 
hi & 70\%& 94 & \textbf{104} & * & 92  & 56 & 0.14\\ 
es & 66\%&  \textbf{843} & 588 & 130 & 173 & 136 & \textcolor{olive}{0.65}\\ 
ar & 69\%&  \textbf{504} & 396 & 182 & 85 & 108 & \textcolor{olive}{0.71}\\ 
fr & 56\%& \textbf{832} & 557 & 215 & 236 & \textcolor{olive}{149} & 0.54\\ 
bn & \textcolor{red}{79\%}& 73 & \textbf{151} & * & 59 & 68 & 0.20\\ 
pt & 73\%& \textbf{479} & 442 & 167 & 161 & 114 & 0.42\\ 
ru & \textcolor{olive}{36\%}& \textbf{1,423} & 657 & 265 & 431 & \textcolor{olive}{146} & 0.60\\ 
id & 76\%& \textbf{411} & 246 & 129 & 133 & 78 & 0.26\\ 
de & 70\%& \textbf{485} & 237 & 169 & 203 & 122 & 0.38\\ 
ja & 59\%& * & \textbf{736} & 434 & 219 & 114 & 0.36\\ 
ko & 59\%& 1,064 & \textbf{1,308} & 574 & 663 & 112 & 0.47\\ 
fa & 57\%& 473 & \textbf{544} & 263 & 194 & 117 & 0.48\\ 
sw & 78\%& 230 & \textbf{389} & 85 & 79 & 89 & 0.50\\ 
ha & 51\%& 30 & \textbf{99} & 30 & 44 & \textcolor{red}{49} & \textcolor{red}{0.10}\\ 
az & 75\%& 123 & \textbf{212} & 81 & 64 & 78 & 0.43\\ 
am & 77\%& 43 & \textbf{106} & 17 & 19 & \textcolor{red}{38} & \textcolor{red}{0.13}\\ 
su & \textcolor{red}{79\%}& \textbf{167} & 146 & 49 & 76 & 62 & \textcolor{red}{0.11}\\ 
el & 70\%& \textbf{330} & 307 & 161 & 135 & 106 & 0.36\\ 
as & \textcolor{red}{83\%}& 41 & \textbf{53} & * & 42 & \textcolor{red}{41} & 0.30\\ 
ca & 72\%& 213 & \textbf{339} & 170 & 224 & 121 & 0.49\\ 
gl & 75\%& \textbf{314} & 297 & 129 & 143 & 113 & 0.37\\ 
eu & 78\%& 132 & \textbf{144} & 48 & 66 & 68 & 0.37\\ 
\bottomrule
\end{tabular}
\caption{Model outputs for the 24 languages. \textbf{Own}: percentage of references to countries in which the language is an official language; \textbf{Flags}: references to each country (top 4 displayed); * mark references to own countries; \textbf{Div} and \textbf{Ent} represent the average for diversity and entropy for all models. Colours indicate the \textcolor{olive}{top three} and \textcolor{red}{bottom three} results.
}
\label{tab:frontier_out_lang}
\end{table}

\setlength{\tabcolsep}{6pt}

 % by LANG

\paragraph{Bias is consistent on exogenous references across languages:}
When isolating non-own-country references, a consistent global hierarchy emerges across all languages. Japan and the United States remain the most frequently cited countries, followed by India, China, and France.
This suggests that cross-country mentions are driven more by a model's inherent priors of cultural salience than by contextual relevance to the specific language, underscoring a systematic overrepresentation of a narrow set of culturally dominant entities.

\paragraph{Low-resource languages produce more self-referential outputs:}
A key factor moderating this bias is the resource level of the language. Comparing languages by the volume of available training data reveals a clear pattern. Higher-resource languages (e.g., \textit{en, zh, es, fr, ru}) show a lower proportion of own-country references and higher output diversity, with a significant skew toward globally dominant regions like Japan and the United States. Lower-resource languages (e.g., \textit{su, as, am, ha, eu}), demonstrate a markedly higher proportion of own-country references (or an increase in non-responses) and lower diversity scores. This pattern suggests a link between training data volume and cultural framing: for languages with less comprehensive corpora, models default to safer, more self-referential, or less diverse outputs (more detail in Appendix \ref{sec:front-model-analysis-anex_lang}).

\begin{figure}
    \centering    
    \includegraphics[
    clip,width=1\linewidth]{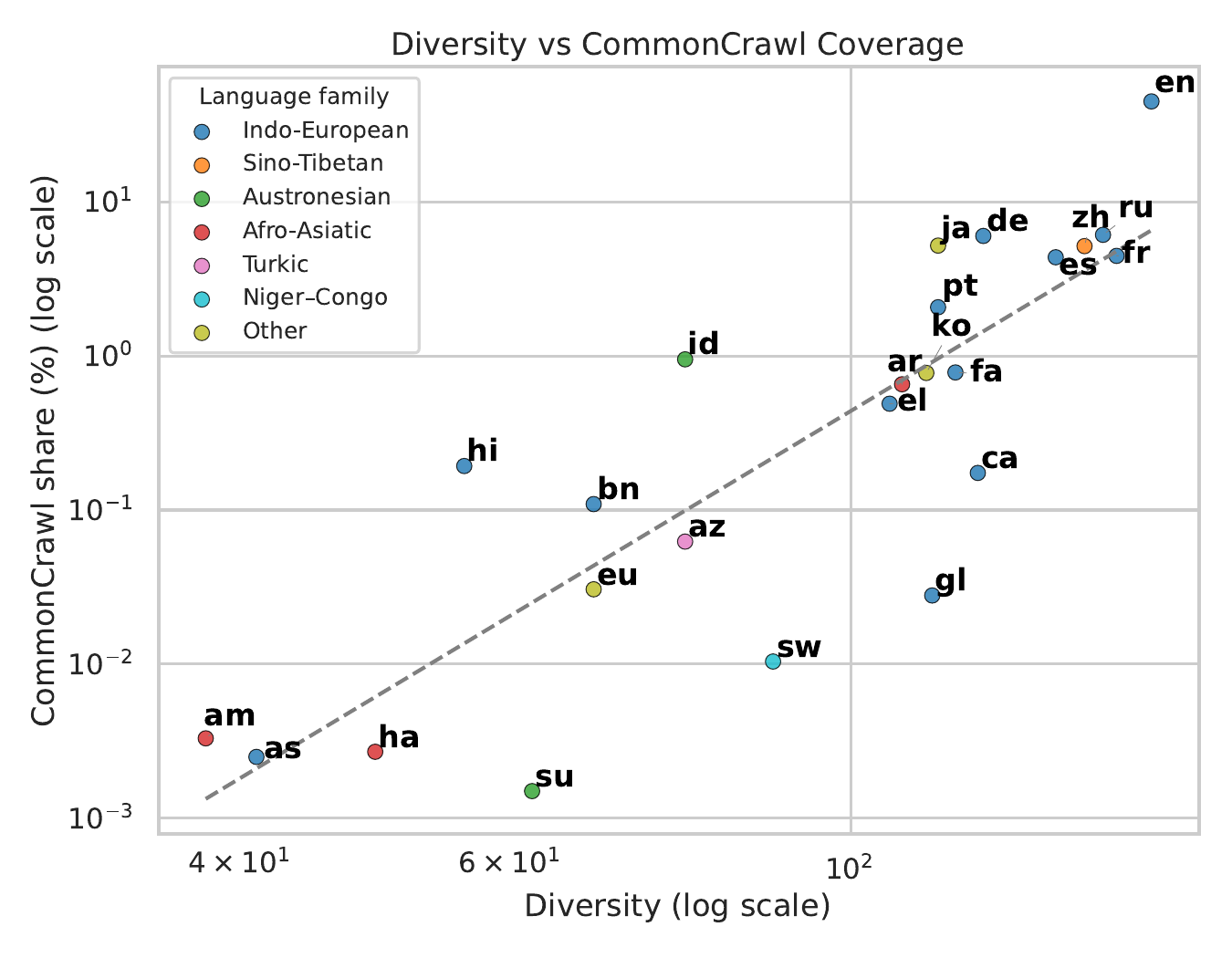}
    \caption{Correlation between CommonCrawl data share and output diversity (mean of models) across 24 languages (log-scale).}
    \label{fig:cc-div-fig}
\end{figure}

\paragraph{Pretraining data and diversity metrics correlate:}
To quantify the relationship between output diversity and language resource availability, we perform a statistical analysis correlating output diversity with the amount of CommonCrawl data available for each language, used as a proxy for pretraining resources (see Figure \ref{fig:cc-div-fig}). Across all frontier models, output diversity correlates positively with CommonCrawl coverage. 
Spearman’s correlation score, $r_s=0.843$, reveals a strong rank-based association; being statistically significant (p-value < 0.001). Analyses at the individual model level further confirm this trend for the majority of systems (see Appendix~\ref{sec:cc-div_anex}). Together, these results indicate that higher-resource languages benefit from more varied cultural representations, whereas lower-resource languages tend to elicit more self-referential or constrained outputs, reflecting reduced diversity in generated content.

\paragraph{Summary of findings and discussion.}  Overall, these findings demonstrate that language models exhibit strong language-aware referencing behaviour. However, once self-references are excluded, models consistently rely on the same narrow set of countries across languages. A particularly notable result is the disproportionate prominence of Japan, which emerges as one of the most frequently referenced exogenous countries across nearly all languages. United States is also highly salient, although this is less surprising given results from previous work and its global influence and the dominance of English as a high-resource language. Taken together, these patterns suggest that the composition of training data, either during pretraining or post-training, plays a central role in shaping models’ culturally biased referencing behaviour.

\begin{figure*}[t]
\centering
    \includegraphics[trim=12 160 26 160, clip,width=0.49\linewidth]{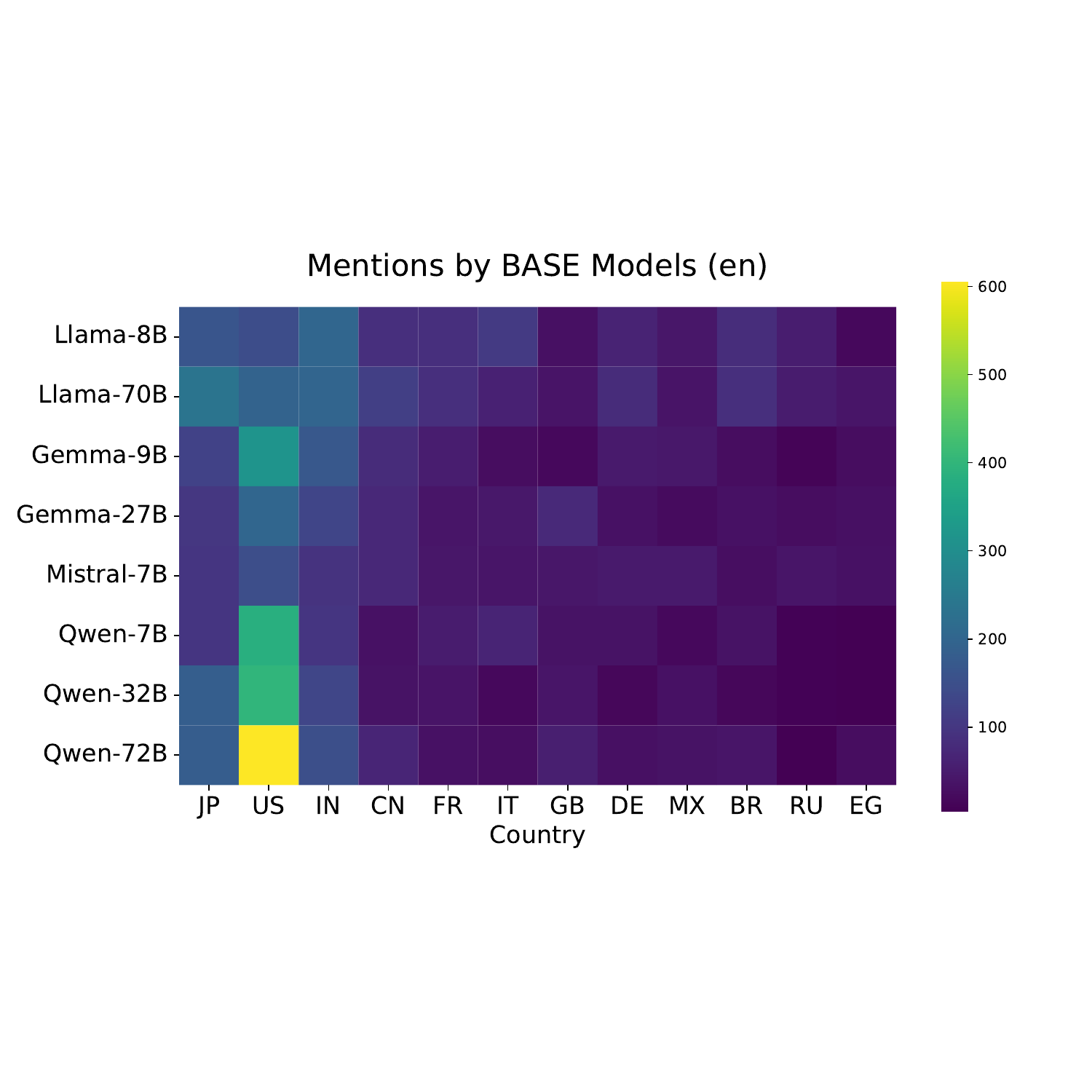}
    \includegraphics[trim=12 160 26 160, clip,width=0.49\linewidth]{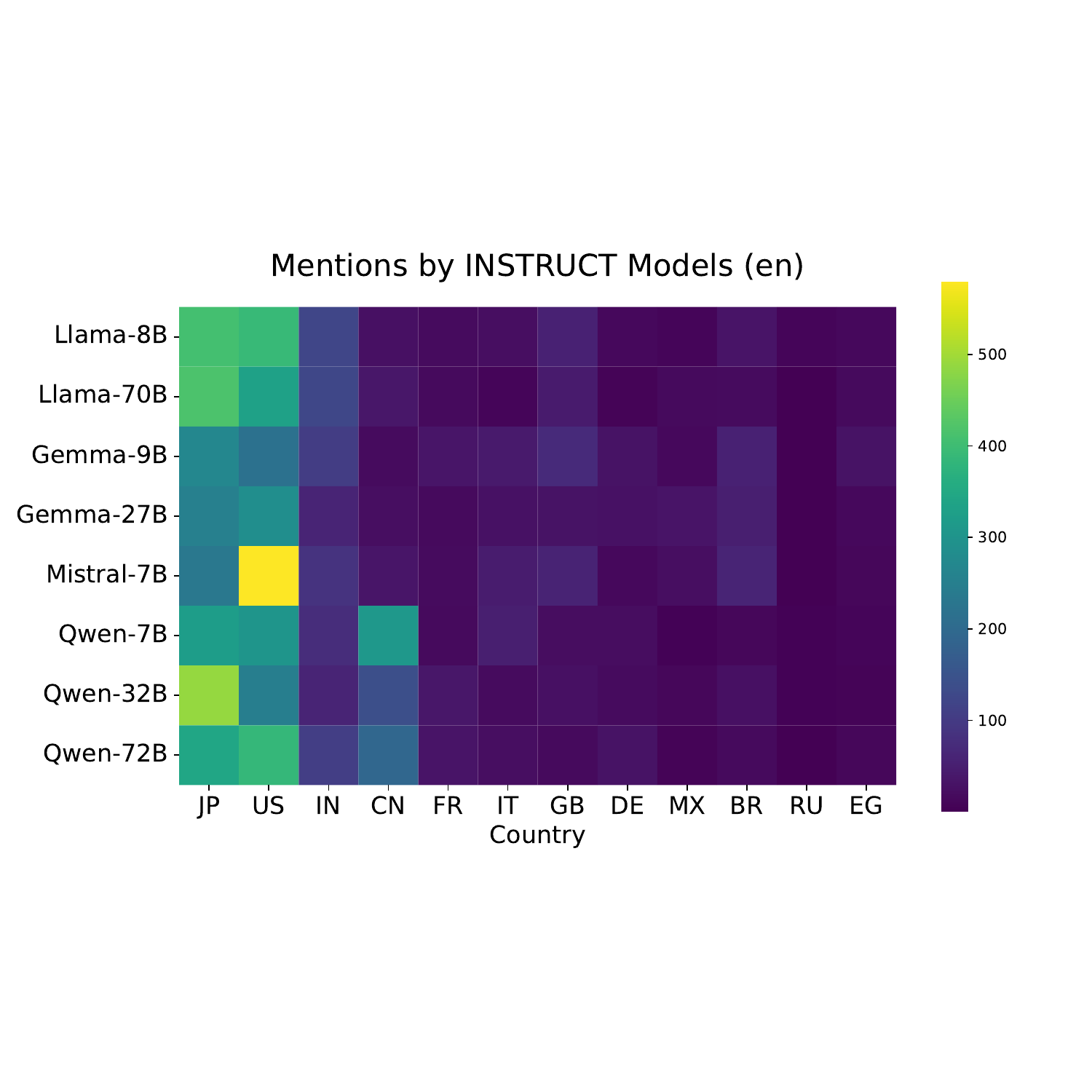}  \hfil 
    \caption{Number of outputs from Base and Instruct models of Llama-3.1, Gemma-2, Mistral and Qwen2.5 variants over the English questions. Countries included: Japan (JP), USA (US), India (IN), China (CN), France (FR), Italy (IT),  UK (GB), Germany (DE), Mexico (MX), Brazil (BR), Russia (RU), Egypt (EG).}
    \vspace{-10pt}
    \label{fig:base_vs_ins_hm} \hfil
\end{figure*}

\section{RQ3: At What Stage of LLM Training Do Regional Cultural Biases Emerge?}

To identify the training stage at which regional cultural biases emerge, we extend our analysis to open-weight \emph{base} and \emph{instruction-tuned} LLMs. By comparing models before and after post-training, we aim to disentangle biases learned during large-scale pretraining from those introduced or amplified during supervised fine-tuning and alignment. 

\subsection{Experimental Setting}

\paragraph{Comparison Models.} To assess whether instruction tuning itself introduces or amplifies bias, we analyze open models providing both base and instruct variants across multiple enterprises and parameter scales, using default settings. This evaluation is performed only on English and includes Llama-3.1 (8B, 70B) \cite{dubey2024llama}, Qwen2.5 (7B, 32B, 72B) \cite{qwen2.5}, Gemma-2 (9B, 27B) \cite{team2024gemma} and Mistral-7B-v0.3 (7B) \cite{jiang2023mistral}. In addition, we analyze OLMo-2 (7B, 32B) \cite{olmo2} and OLMo-3 (7B, 7B-Think) \cite{olmo3}, which provide base, supervised fine-tuned (SFT), and fully aligned instruction models. This allows us to disentangle the effects of instruction tuning from those of subsequent supervised fine-tuning. 

\paragraph{Prompting.} Due to model availability, we restrict all experiments to English prompts. Given the nature of base models that are explicitly trained to predict the next tokens, we rephrase the question prompt utilised in our experiments so the sentence can be completed rather than answered. Everything else remains unchanged and both base and instruct models are given the same questions as input.\footnote{In Appendix \ref{sec:prompt-analysis-anex}, we provide the prompt and an analysis of the impact of the prompt in base and instruct models.}

\begin{table}[t]
\centering
\small
\setlength{\tabcolsep}{4pt}
\begin{tabular}{@{}lcccccc@{}}
\toprule
 & \multicolumn{3}{c}{\textbf{Diversity}} & \multicolumn{3}{c}{\textbf{Entropy}} \\
\cmidrule(lr){2-4} \cmidrule(lr){5-7}
\textbf{Model} & \textbf{Base} & SFT & \textbf{Inst.} & \textbf{Base} & SFT &\textbf{Inst.} \\
\midrule
Llama-3.1-8B   & 109 &-& \textbf{174} & \textbf{0.78} &-& 0.66 \\
Llama-3.1-70B  & 117 &-& \textbf{174} & \textbf{0.79} &-& 0.65 \\

gemma-2-9b     &  95 &-& \textbf{136} & \textbf{0.70} &-& 0.71 \\
gemma-2-27b    &  80 &-& \textbf{154} & \textbf{0.77} &-& 0.73 \\

Mistral-7B     &  91 &-& \textbf{160} & \textbf{0.81} &-& 0.63 \\

Qwen2.5-7B     & 147 &-& \textbf{150} & \textbf{0.71} &-& 0.62 \\
Qwen2.5-32B    & \textbf{142} &-& 118 & \textbf{0.65} &-& 0.61 \\
Qwen2.5-72B    & \textbf{144} &-& 127 & \textbf{0.66} &-& 0.60 \\

\midrule
Olmo-3-7B   & 114 & \textbf{145} & 124 & \textbf{0.74} & 0.66 &  0.64 \\
Olmo-3-7B-Think & 114 & 121 & \textbf{127} & \textbf{0.74} & 0.66 & 0.68 \\
OLMo-2-7B & 102 & 151 & \textbf{158} & \textbf{0.84} & 0.63 & 0.64 \\
OLMo-2-32B & 88 & 151 & \textbf{161} & \textbf{0.77} & 0.65 & 0.71 \\
\bottomrule
\end{tabular}
\caption{Diversity and normalized entropy scores for base, SFT (for OLMo) and instruct model variants. For each model, the highest diversity and entropy are bolded.}

\label{tab:div_ent_base_inst_only}
\end{table}
\setlength{\tabcolsep}{6pt}

\subsection{Findings}

Figure~\ref{fig:base_vs_ins_hm} contrasts the distributions of country references produced by base and instruction-tuned models, while Table \ref{tab:div_ent_base_inst_only} shows the diversity and entropy scores for each model variant.

\paragraph{Cultural distributions are more balanced on Base Models:}
Base models (Figure~\ref{fig:base_vs_ins_hm}, left) exhibit a more balanced geographic coverage. While the United States remains prominent, substantial references are also made to Japan, India, China, and several European countries. This pattern is consistent across architectures and model sizes, suggesting that pretraining alone yields a comparatively diffuse set of cultural associations. This is further reflected on the entropy scores, which are consistently lower after instruction-tuning for all models.

\paragraph{Cultural leaning is stronger after Instruction-Tuning:}
Despite providing a higher number of countries overall, instruction-tuned models (Figure~\ref{fig:base_vs_ins_hm}, right) display a pronounced concentration of references to a narrow set of regions. Across all examined model families, instruction tuning sharply increases alignment with the United States and Japan while reducing references to most other countries. This convergence toward culturally dominant regions occurs even in models developed outside Western contexts, indicating that post-training induces a homogenization of cultural perspectives rather than merely reflecting model origin.

\begin{figure}
    \centering    
    \includegraphics[%trim=14 10 40 30,%\includegraphics[trim=14 10 40 30,
    clip,width=1\linewidth]{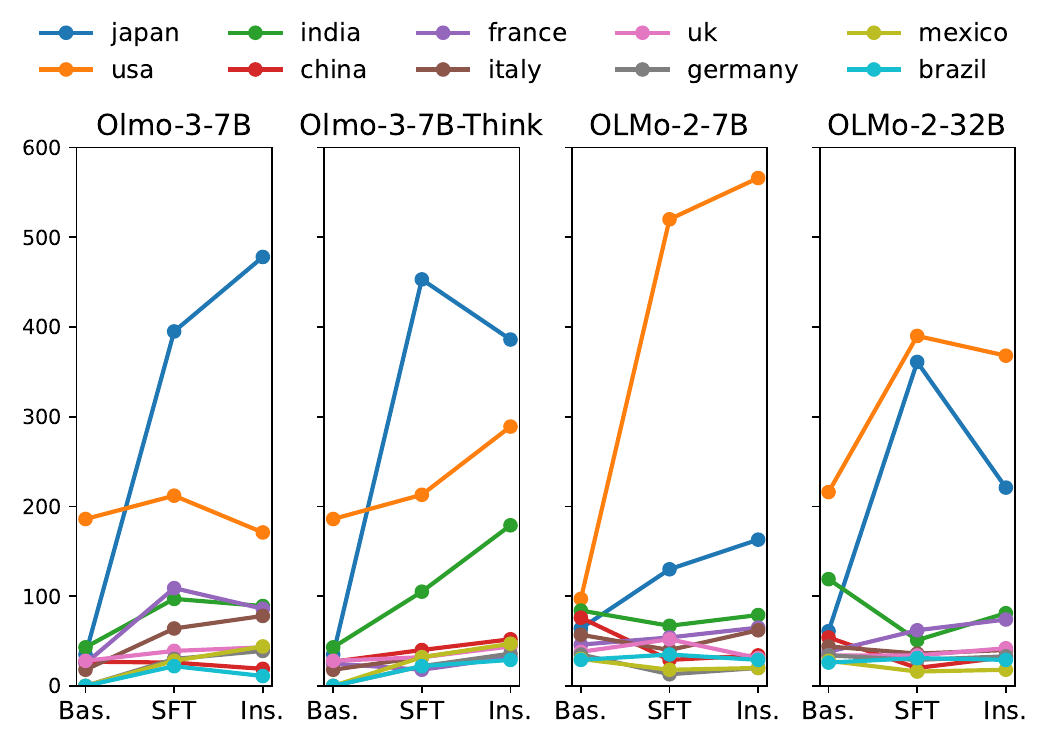}
    \caption{Number of outputs mentioning top-referenced countries from the Base, SFT, and Instruct variants of the named OLMo models.}
    %\vspace{-14pt}
    \label{fig:base-sft-ins}
\end{figure}

\paragraph{Supervised fine-tuning as a primary driver of cultural bias:}
To further assess whether instruction tuning itself introduces or amplifies cultural bias, we analyze OLMo models, which provide base, supervised fine-tuned (SFT), and instruction-aligned variants. In Figure \ref{fig:base-sft-ins} we observe that the most substantial shift in country reference distributions occurs during supervised fine-tuning: SFT sharply increases concentration on a small number of dominant regions (most notably the United States and Japan) while reducing entropy across all model outputs (see Table \ref{tab:div_ent_base_inst_only}). Subsequent instruction alignment only marginally mitigates these effects and does not recover the more balanced distributions observed in base models. These results indicate that alignment-induced cultural bias primarily originates from supervised fine-tuning data rather than pretraining, and that later instruction tuning largely preserves, rather than corrects, these learned cultural priors (see Appendix~\ref{sec:olmo-anex} for further details).

\paragraph{Summary of findings and discussion.}
Overall, base models distribute references more evenly across regions, exhibiting lower concentration of regions despite limited diversity. Instruction-tuned models, by contrast, show substantially more concentrated cultural biases. These findings suggest that post-training, rather than pretraining, plays a decisive role in shaping the dominant cultural perspectives expressed by LLMs (more details and results in Appendix \ref{sec:base-ins-anex}). The results from OLMO suggest that this behaviour starts appearing specifically after supervised fine-tuning. Finally, we checked the instructions provided to LLMs that could potentially amplify these biases without clear evidence (see Appendix \ref{sec:instruction-analysis-anex}).

\section{Conclusions}

Ensuring cultural awareness and diversity in LLMs is an open research question in NLP. However, measuring the impact of certain decisions is hard given all the cultural aspects that are involved in model responses. In this paper, we created a dataset of open questions based on a taxonomy of eleven cultural domains and sixty-six subtopics. We used this dataset to test eight frontier LLMs by prompting them in twenty-four languages. Instead of testing the models about their cultural knowledge, we let them answer the questions openly by only requiring them to provide regional examples of their choice.

The results show an inclination of models to provide cultural examples related to regions for which the prompted language is an official language. Not only that, but there is a similarity on the type of example that models choose, preferring countries such as Japan, United States or India overwhelmingly with respect to other choices across the world. When analysing the potential causes, the results indicate that this concentration of answers emerges during post-training. While this is a stage traditionally thought to guide models toward providing more diverse and unbiased responses, it appears to also increase this cultural bias into a different direction. For future work, it would be interesting to think of new methods to take this behaviour into account, in order to better understand the post-training effects in model output diversity as a whole, in particular when it comes to different languages and regions.

\section*{Limitations}

Our study has several limitations that should be considered when interpreting the results.

1) Due to the lack of open data in frontier multilingual LLMs, the comparison between base and instruction-tuned models is conducted only in English, which limits the extent to which these findings generalize to other languages. In addition, base and instruction-tuned models differ inherently in training objectives and interaction style, which may influence generation behaviour independently of the factors analyzed here. Nevertheless, this comparison provides a first step toward analyzing the intersection between base and instruction-tuned models, which we plan to explore further in future work.

2) All experiments are conducted using default generation settings, and we do not explore the robustness of our results to alternative decoding parameters. We deliberately chose this approach in order to focus on the settings employed by most users, enabling us to compare across different languages and models -- an additional exhaustive exploration of decoding parameter configurations would incur high generation and analysis costs.

3) We rely primarily on an automatic judge rather than human evaluation for all languages. While we validate the judge on a subset of languages, automatic evaluation may fail to capture subtle linguistic or cultural distinctions. Nevertheless, given that our expertise is limited to a small number of languages, conducting large-scale human evaluation would not be feasible, making automatic evaluation a necessary choice which we believe does not affect the reliability of the results as a whole.

4) Our evaluation covers only 24 languages, representing a limited subset of global linguistic diversity. We further restrict some analyses to official languages of countries, which may not accurately reflect real-world language use, as many widely spoken languages extend beyond official or national boundaries. Moreover, referring to countries as proxies for cultural contexts may lead to oversimplifications. 

5) Finally, we do not assess the impact of the observed behaviours on downstream tasks, leaving open the question of how these findings translate to applied or real-world scenarios.

\section*{Ethical considerations}

In this work, we use LLMs to semi-automatically generate candidate questions for CROQ (GPT-5.1, see Sec. \ref{sec:dataset} for more details) and as-a-judge models (gpt-4o-mini, Sec. \ref{sec:processing_model_responses}) to extract information from model responses. Because this process may introduce model biases, all generated questions underwent manual review to remove repetitions, accidental location references, and other artifacts. This human oversight was particularly important given that our work explicitly examines cultural biases in LLM outputs.

The cultural taxonomy underlying CROQ was developed by aggregating topics from existing benchmarks and expanding them into a comprehensive framework of 11 domains and 66 subtopics. Although we designed this taxonomy to capture a broad spectrum of culturally relevant dimensions, we acknowledge that cultural phenomena exist on a continuum and cannot be exhaustively categorized. This structural limitation means our analysis, while broad, remains grounded in particular cultural frameworks rather than offering a universal view of human culture.

In addition, we made deliberate efforts to ensure broad linguistic coverage by including 24 languages spanning different resource levels, language families, and geographical regions. This enables the study of cultural biases across varied linguistic contexts, including communities often underrepresented in NLP. However, 24 languages represent only a small fraction of the world's linguistic diversity, and our selection inevitably privileges certain communities and perspectives.

Consequently, this work remains affected by the broader inequalities that shape language technology research. We encourage future work to extend cultural evaluation frameworks beyond country-level language boundaries and better represent underrepresented communities.

Our results, particularly the concentration of model outputs towards culturally dominant regions like Japan and the United States, have implications for global users of these systems. Models trained to favour certain cultural perspectives may provide inadequate or inappropriate responses for users from other backgrounds. We hope this work contributes to awareness among both model developers and users about the need for more balanced and inclusive cultural training.

\section*{Acknowledgements}

Joseba Fernandez de Landa is supported by Grant DeepThought (PID2024-159202OB-C21), funded by MICIU/AEI/10.13039/501100011033 and ERDF, EU, as well as the ALIA Models Development Project, Resolution of the SEDIA 19/08/2024, within the framework of the National Language Technologies Plan – ENIA 2024, funded by MTDFP, PRTR, and the EU – NextGenerationEU.

Jose Camacho-Collados is supported by a UKRI Future Leaders Fellowship.

\bibliography{anthology,custom}

\appendix

\section{Validation of Dataset Translations}
\label{sec:translation-anex}

Given the scale of the CROQ dataset (31,680 total questions across 24 languages), we took deliberate steps to ensure the high quality and semantic integrity of the translations. Because the core 1,320 questions are intentionally designed to be short, direct, and structurally simple (as shown in Section 3 and Figure 1), the translation process is highly straightforward. Contemporary machine translation systems are highly effective at handling such short-form text without loss of context. Accordingly, we utilized \textit{GPT-5.1} for the translation process.

To verify this, we analyzed a representative sample of 6 (zh, ko, es, ca, gl, eu) out of the 23 translated languages, explicitly including low-resource languages, and conducted a validation check with native speakers. This involved a manual evaluation of 400 questions in total. The main conclusion from this analysis is that the automated translations are accurate and highly consistent across the dataset. While native speakers noted that a small fraction of the prompts exhibited minor artificiality (occasional phrasing that sounds slightly unnatural or different from how a native speaker would typically talk) the translations remain entirely correct and semantically faithful to the original prompts. This validation confirms that the minor stylistic variations do not impact the clarity of the questions, making the translation pipeline fully sufficient and reliable for the scope of this benchmark.

\section{Topics and subtopics in the Taxonomy }
\label{sec:taxonomy-anex}

\begin{table*}[h!]
\centering
\scriptsize

\begin{tabular}{@{}ll@{}}
\toprule
\textbf{Domains} & \textbf{Subtopics} \\ \midrule

\multirow{5}{*}{\textbf{1. Beliefs, Values, and Identity}}
& \textbf{Religion \& Spirituality}: Beliefs and practices about the divine, morality, and afterlife. \\ 
& \textbf{Values \& Ethics}: Core beliefs about right/wrong, community, and individual roles. \\ 
& \textbf{Symbolism}: Visual or material symbols that carry cultural meaning. \\ 
& \textbf{Mythology \& Folklore}: Narratives that explain origins, values, or natural phenomena. \\ 
& \textbf{Gender Roles}: Cultural expectations related to gender identity and expression. \\ \midrule

\multirow{5}{*}{\textbf{2. Social Structure and Daily Life}}
& \textbf{Family \& Social Roles}: Structure of family and communal relationships. \\ 
& \textbf{Community Life}: How people relate to neighbors and groups. \\ 
& \textbf{Manners \& Etiquette}: Everyday norms for respectful behavior and interaction. \\ 
& \textbf{Work Culture}: Attitudes and habits related to labor, time, and productivity. \\ 
& \textbf{Conflict Resolution}: How societies handle disputes or disagreements. \\ \midrule

\multirow{5}{*}{\textbf{3. Knowledge, Communication, and Education}}
& \textbf{Language \& Communication}: Spoken, written, and non-verbal systems of communication. \\ 
& \textbf{Education Systems}: Ways knowledge is transmitted and valued. \\ 
& \textbf{Technology \& Tools}: Use of tools and innovation in daily and cultural life. \\ 
& \textbf{Knowledge Transmission}: How wisdom and skills are preserved across generations. \\ 
& \textbf{Science \& Philosophy}: Intellectual traditions shaping discovery and thought. \\ \midrule

\multirow{9}{*}{\textbf{4. Cultural Expression and the Arts}}
& \textbf{Music}: Rhythmic and melodic expression of culture \\ 
& \textbf{Dance}: Movement as performance, ritual, or celebration \\ 
& \textbf{Art}: Visual cultural expression (painting, sculpture, etc.) \\ 
& \textbf{Literature \& Storytelling}: Written and oral traditions conveying culture and values \\ 
& \textbf{Theater \& Performance}: Live dramatic cultural expression \\ 
& \textbf{Film \& Cinema}: Storytelling through modern moving-image media \\ 
& \textbf{Crafts \& Handicrafts}: Traditional material culture and artisanal skills \\ 
& \textbf{Jewelry \& Body Decoration}: Ornamentation for identity, beauty, or tradition \\ 
& \textbf{Architecture \& Housing}: Building styles and structures shaped by culture and climate \\ \midrule

\multirow{6}{*}{\textbf{5. Food, Drink, and Leisure}}
& \textbf{Food \& Cuisine}: Traditional dishes, cooking styles, and dietary customs \\ 
& \textbf{Traditional Beverages}: Culturally significant drinks \\ 
& \textbf{Recreation \& Leisure}: Activities for enjoyment and relaxation \\ 
& \textbf{Sports \& Games}: Physical activities and competitions enjoyed culturally \\ 
& \textbf{Life-Cycle Rituals}: Ceremonies marking stages of life \\ 
& \textbf{Festivals \& Celebrations}: Seasonal, national, or spiritual gatherings expressing shared identity \\ \midrule

\multirow{7}{*}{\textbf{6. Geographic Aspects}}
& \textbf{Climate \& Environment}: How weather and landscape shape cultural practices, housing, and food \\ 
& \textbf{Topography \& Land Use}: How people interact with physical geography like mountains, rivers, or plains \\ 
& \textbf{Rural vs Urban Culture}: Differences in lifestyle, work, and values between city and countryside \\ 
& \textbf{Regional Identity}: Sub-national cultural traits tied to a specific region \\ 
& \textbf{Borders \& Territory}: How borders influence cultural blending or division \\ 
& \textbf{Migration \& Diaspora}: Movements shaping identity \\ 
& \textbf{Relationship with Nature}: Beliefs and practices related to land, animals, and ecology \\ \midrule

\multirow{7}{*}{\textbf{7. Political Aspects}}
& \textbf{Government Systems}: Political structures influencing law, rights, and daily life \\ 
& \textbf{Law \& Legal Traditions}: Cultural expectations of justice, rights, and punishment \\ 
& \textbf{Civic Values \& Participation}: How people engage with politics, voting, and national identity \\ 
& \textbf{Freedom of Expression}: Cultural and political limits on speech or art \\ 
& \textbf{Colonial History \& Influence}: Impact of colonization on culture, identity, and systems \\ 
& \textbf{Power \& Hierarchy}: How authority is structured culturally and politically \\ 
& \textbf{Human Rights \& Social Movements}: How cultures address equity, justice, and activism \\ \midrule

\multirow{4}{*}{\textbf{8. Health \& Wellness}}
& \textbf{Health Practices}: Traditional, spiritual, or modern approaches to healing and care \\ 
& \textbf{Public Health}: Collective strategies for health and safety \\ 
& \textbf{Wellbeing \& Lifestyle}: Balancing mental, physical, and social health \\ 
& \textbf{Birth \& Reproductive Health}: Fertility, childbirth, and parenting norms and practices \\ \midrule

\multirow{6}{*}{\textbf{9. Media \& Entertainment}}
& \textbf{Local Media}: Print, radio, and TV shaping cultural narratives \\ 
& \textbf{Digital Media}: Online platforms influencing communication, identity, and activism \\ 
& \textbf{Entertainment Industries}: Mass production of cultural content \\ 
& \textbf{News \& Information}: Flow of journalism and narratives in society \\ 
& \textbf{Popular Culture}: Icons, trends, and shared cultural references \\ 
& \textbf{Gaming \& Interactive Media}: Games, esports, and digital interactivity \\ \midrule

\multirow{6}{*}{\textbf{10. History}}
& \textbf{Historical Events}: Major turning points shaping collective identity \\ 
& \textbf{Historical Figures}: Leaders, thinkers, and cultural icons \\ 
& \textbf{Cultural Memory}: How societies remember, teach, and reinterpret their past \\ 
& \textbf{Colonialism \& Resistance}: Struggles of domination and liberation \\
& \textbf{National Narratives}: Shared stories of origin, destiny, and identity \\ 
& \textbf{Heritage \& Preservation}: Protecting, curating, and interpreting cultural pasts \\ \midrule

\multirow{6}{*}{\textbf{11. Economy \& Industry}}
& \textbf{Key Industries}: Dominant sectors shaping work and identity \\ 
& \textbf{Trade \& Exchange}: Movement of goods, ideas, and culture \\ 
& \textbf{Labor \& Employment}: Work systems, professions, and class structures \\ 
& \textbf{Wealth \& Inequality}: How resources and opportunities are distributed \\ 
& \textbf{Economic Growth \& Development}: Paths to modernization and sustainability \\ 
& \textbf{Globalization \& Cultural Economy}: Markets influencing culture and identity \\ \bottomrule

\end{tabular}

\caption{Overview of the taxonomy’s domains (11) and subtopics (66), with brief definitions for each category.}
\label{tab:taxonomy_explained}

\end{table*}

We construct a comprehensive taxonomy comprising 66 fine-grained cultural subtopics, which are organized into 11 higher-level cultural domains. The taxonomy was designed to capture a broad and diverse range of cultural phenomena while maintaining sufficient granularity to support detailed analysis. Each higher-level domain aggregates thematically related subtopics, enabling both coarse-grained and fine-grained examination of cultural dimensions. The complete taxonomy, including definitions and hierarchical relationships between domains and subtopics, is presented in Table \ref{tab:taxonomy_explained}.

\begin{figure} [ht]
    \centering    
    \includegraphics[width=1\linewidth]{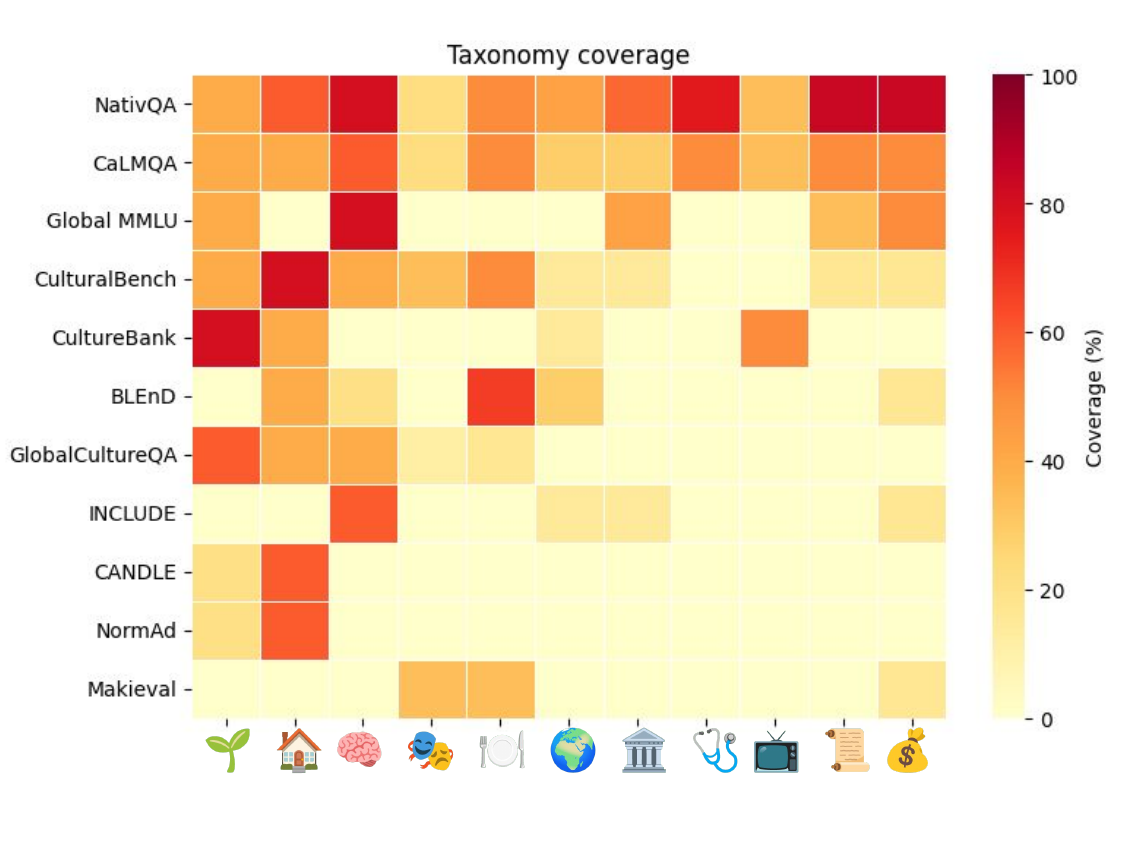}
    \caption{Taxonomy coverage for Related Work datasets. Percentage of subtopics covered in each higher-level domain proposed in our taxonomy for the named datasets.}
    \label{fig:taxonomy_cover}
\end{figure}

We conducted a semi-automatic evaluation of our proposed categories and subcategories, and we analyzed the datasets introduced in related work. Figure~\ref{fig:taxonomy_cover} illustrates the extent to which each general topic in our taxonomy is covered by existing datasets. The analysis shows that many datasets address only a limited subset of the general topics defined in our taxonomy. Although NativQA~\cite{hasan2024nativqa} and CaLMQA~\cite{arora-etal-2025-calmqa} cover all general topics, their coverage across subtopics is uneven, with notable imbalances within individual topics. Our goal is not to propose a definitive or exhaustive taxonomy of cultural topics, but rather to cover as broad a range of cultural topics as possible, in order to prompt LLMs with maximal cultural diversity.

\section{Selected languages analysis}
\label{sec:lang_anex}

Table \ref{tab:all_langs_anex} shows the coverage of languages in our work, detailing the distribution of the 24 languages included in our dataset. For each language, the table reports its relative proportion within the Common Crawl snapshot CC-MAIN-2025-43, together with an estimate of the total number of speakers (in millions), derived from publicly available Wikipedia statistics. In addition, we provide the corresponding language family for each language, as well as the ISO 639-1:2002 two-letter language code used consistently throughout the paper and accompanying resources. This information offers a concise yet comprehensive view of the linguistic diversity and representativeness of the dataset analyzed in this work.

\begin{table}[h!]
\centering
\scriptsize
\begin{tabular}{@{}lllrr@{}}
\toprule
\textbf{Code} & \textbf{Language} & \textbf{Family} & \textbf{Speakers (M)} & \textbf{CC \%} \\
\midrule
en & English    & Indo-European     & 1,528     & 44.8292 \\
zh & Chinese    & Sino-Tibetan      & 1,184     & 5.1598 \\
hi & Hindi      & Indo-European     & 609       & 0.1930 \\
es & Spanish    & Indo-European     & 558       & 4.3672 \\
ar & Arabic     & Afro-Asiatic      & 335       & 0.6549 \\
fr & French     & Indo-European     & 312       & 4.4605 \\
bn & Bengali    & Indo-European     & 284       & 0.1093 \\
pt & Portuguese & Indo-European     & 267       & 2.0696 \\
ru & Russian    & Indo-European     & 253       & 6.1083 \\
id & Indonesian & Austronesian      & 252       & 0.9505 \\
de & German     & Indo-European     & 134       & 6.0060 \\
ja & Japanese   & Other             & 126       & 5.2018 \\
ko & Korean     & Other             & 82        & 0.7754 \\
fa & Persian    & Indo-European     & 127       & 0.7814 \\
sw & Swahili    & Niger–Congo       & 87        & 0.0104 \\
ha & Hausa      & Afro-Asiatic      & 94        & 0.0027 \\
az & Azerbaijani& Turkic            & 24        & 0.0624 \\
am & Amharic    & Afro-Asiatic      & 60        & 0.0033 \\
su & Sundanese  & Austronesian      & 32        & 0.0015 \\
el & Greek      & Indo-European     & 13        & 0.4898 \\
as & Assamese   & Indo-European     & 24        & 0.0025 \\
ca & Catalan    & Indo-European     & 9         & 0.1741 \\
gl & Galician   & Indo-European     & 2.4       & 0.0279 \\
eu & Basque     & Other             & 0.8       & 0.0306 \\
\bottomrule
\end{tabular}
\caption{Distribution of the selected 24 languages (ISO 639-1:2002) included on our dataset by share of data in CommonCrawl (CC-MAIN-2025-43) and estimated number of total speakers (in millions) based on Wikipedia sources.}
\label{tab:all_langs_anex}
\end{table}

\begin{figure}
    \centering    
    \includegraphics[width=1\linewidth]{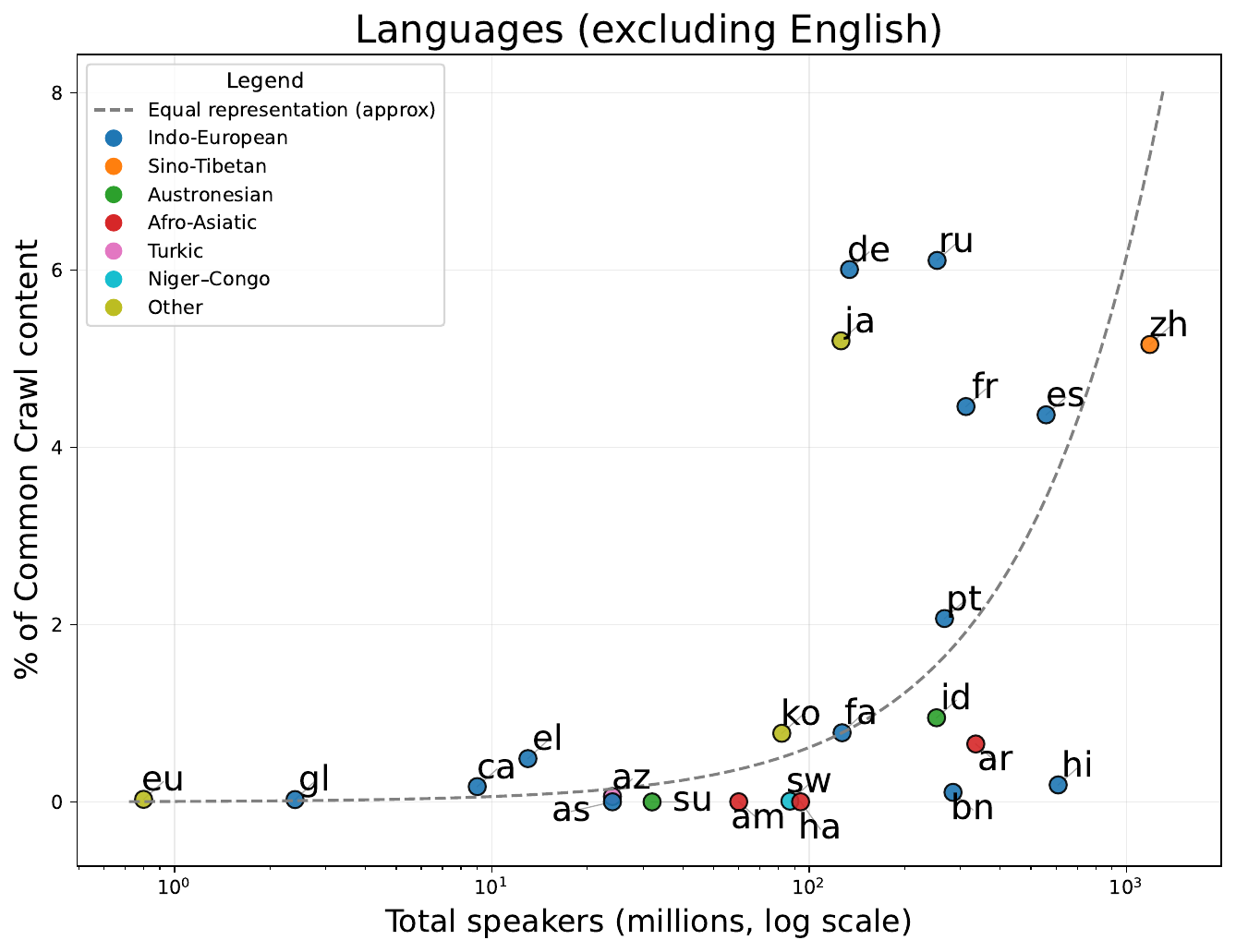}
    \caption{Distribution of the selected 23 languages (English excluded for a more clear visualisation) by share of data in CommonCrawl (CC-MAIN-2025-43) and estimated number of speakers based on Wikipedia sources.}
    \vspace{-10pt}
    \label{fig:languages}
\end{figure}

Figure~\ref{fig:languages} illustrates the relationship between the selected languages (excluding English), showing the correlation between number of speakers and their share in CommonCrawl. The broad dispersion across the plot demonstrates that CROQ covers a wide range of linguistic scenarios, spanning more than six distinct language families.

\section{Further Evaluation Details}
\label{sec:eval_det_anex}

To process the open-ended answers produced by the LLMs, we employ a secondary model as a judge to interpret the responses and extract the relevant information. The objective is to associate each question with a set of countries in order to uncover potential cultural biases. Therefore, the judge model is tasked with identifying up to five regions or places mentioned or implied in each answer. If no such information could be reliably determined, the model is prompted to explicitly indicate that no inference is possible (non-answered responses). 

\begin{table}[htbp]
\centering
\scriptsize
\begin{tabular}{@{}lrrr|rrr@{}}
\toprule
& \multicolumn{3}{c}{Basic eval.} & \multicolumn{3}{c}{Thorough eval.} \\
Lang. 
& \textit{Exp.(1)} & \textit{Inf.(2)} & \textit{Mid.(3)}
& \textit{Exp.(1)} & \textit{Inf.(2)} & \textit{Mid.(3)} \\
\midrule
en   & 93.94    & 92.42 & \textbf{100}         & 92.42 & 89.39 & \textbf{100}  \\
es   & 89.39  & 90.91 & \textbf{96.97}         & 77.27 & 78.79 & \textbf{96.97}  \\
ca   & 93.94    & 96.97  & \textbf{100}        & 56.06 & 46.97 & \textbf{96.97}  \\
eu   & 84.85  & \textbf{96.97} &\textbf{96.97} & 59.09 & 62.12 & \textbf{86.36}  \\
\midrule
\textbf{Avg.} & 90.53 & 94.32  & \textbf{98.48}  & 71.21 & 69.32 & \textbf{95.08}  \\
\bottomrule
\end{tabular}
\caption{Evaluation Accuracy results for LLM as-a-judge in Basic and Thorough evaluations. System prompt types: (1) \textit{Explicit}, (2) \textit{Infer} and (3) \textit{Middle}.}
\label{tab:judge_eval_results}
\end{table}

We evaluated three prompting strategies for the judge model: (1) \textit{Explicit}, in which only regions explicitly mentioned in the text were extracted; (2) \textit{Infer}, which allowed the model to extract regions that were either explicitly stated or reasonably inferred; and (3) \textit{Middle}, a balanced approach that prioritized explicit mentions while permitting limited, non-speculative inference when necessary (see Figures \ref{fig:system_prompt} and \ref{fig:system_prompt_discarded}). 

For the evaluation, we conducted a manual assessment of 264 items (one per subtopic across four languages), covering both high-resource (\textit{en}, \textit{es}) and low-resource (\textit{ca}, \textit{eu}) settings. Ground truth labels for these items are established via manual human annotation. Accuracy is measured as the percentage of instances where the judge model's extractions perfectly matches this human-annotated ground truth.

We considered two evaluation granularities within this setup:\\
\textit{Basic evaluation:} The objective is to extract the country or countries referenced in the text. If a country is explicitly mentioned, that country should be extracted. If a specific sub-national region is mentioned instead, the judge must extract either the general country name or the specific region, and both are considered correct.\\
\textit{Thorough evaluation:} The objective shifts toward finer-grained geographic specificity, and the model is required to adhere strictly to the geographic entity explicitly mentioned in the output; if a country is mentioned, the model must extract the country, and if a specific sub-national region is mentioned, the model must extract that specific sub-national region.\\

\vspace{-2pt}
\begin{figure}[ht]
    \centering
        \begin{tcolorbox}[left=0em]
\small
\vspace{-3pt}
\begin{tabular}{ll} 
  \multicolumn{2}{l}{Text: \textbf{``In Japan, staple foods eaten daily...''}}  \\ 
 Basic Eval: & Output must be -> Japan \\ 
 Thorough Eval: & Output must be -> Japan \\ 
\end{tabular}
\vspace{-2pt}

    \end{tcolorbox}
    \vspace{-2pt}

\label{fig:question-examples}
\end{figure}
\vspace{-2pt}

\vspace{-2pt}
\begin{figure}[ht]
    \centering
        \begin{tcolorbox}[left=0em]
\small
\vspace{-2pt}
\begin{tabular}{ll} 
 \multicolumn{2}{l}{Text: \textbf{``In Paris, notable museums include...''}}  \\ 
 Basic Eval: & Paris \textit{or} France are accepted. \\ 
 Thorough Eval: & Only Paris is accepted. 
\end{tabular}
\vspace{-2pt}

    \end{tcolorbox}
    \vspace{-2pt}

\label{fig:question-examples}
\end{figure}
\vspace{-2pt}

\vspace{-2pt}
\begin{figure}[ht!]
    \centering
        \begin{tcolorbox}[left=0em]
\small
\vspace{-2pt}
\begin{tabular}{ll} 
 \multicolumn{2}{l}{Text: \textbf{``In the Basque Country*, the food is...''}}  \\ 
 \multicolumn{2}{l}{\textit{*transboundary region between Spain and France.}} \\ [2pt]
 Basic Eval: & Spain, France \textit{or} Basque Country. \\ 
 Thorough Eval: & Only Basque Country. 
 
\end{tabular}
\vspace{-2pt}

    \end{tcolorbox}
    \vspace{-2pt}

\label{fig:question-examples}
\end{figure}
\vspace{-8pt}

To account for this finer-grained geographic specificity, we report the comparative results of both setups in Table \ref{tab:judge_eval_results}. The manual evaluation showed that the \textit{(3) Middle} prompting strategy yielded the most accurate and consistent results. Based on this assessment, we adopted this prompt in all subsequent analyses, using gpt-4o-mini from OpenRouter with default settings as the underlying judge model.

\begin{figure*}[ht]
    \centering
        \begin{tcolorbox}[left=.5em]
\begin{Verbatim}[fontsize=\small,commandchars=\\\{\}]
You are a precise geographical context classifier. Your task is to determine which nation, 
country or state(s) a given text is explicitly referring to. Follow these rules strictly:

   1. Do not infer nations, countries or states based on general knowledge, stereotypes, or 
   assumptions beyond what is stated in the text.

   2. If multiple nations, countries or states are mentioned, list all of them. Not more than 5.

Keep your answers concise and limited to the nation, country or state names. Translate the final 
answers into English.
\end{Verbatim}
    \end{tcolorbox}
    \caption{Chosen System prompt, \textit{Middle} (3), given to the Judge model}
    \label{fig:system_prompt}
\end{figure*}

\begin{figure*}[ht]

\begin{tcolorbox}[width=1\linewidth,left=.5em]
\begin{Verbatim}[fontsize=\scriptsize]
You are a geographical context classifier. Identify the nation, country, or state(s) explicitly mentioned in the text.				

Rules:			
1. Only list countries/states that appear in the text. Do not infer.			
2. If multiple countries/states are mentioned, list up to 5.			
3. Only list nations, countries, or states; skip regions or cities unless needed for clarity.

Answer with country names only, in English, separated by commas.			
\end{Verbatim}
\end{tcolorbox}

\begin{tcolorbox}[width=1\linewidth,left=.5em]
\begin{Verbatim}[fontsize=\scriptsize]
You are a geographical context classifier. Identify the nation, country, or state(s) the text refers to.		

Rules:		
1. If no country is mentioned, infer the most likely one from context.
2. If multiple countries/states are mentioned, list up to 5.		
3. Only list nations, countries, or states; skip regions or cities unless needed for clarity.	

Answer with country names only, in English, separated by commas.		
\end{Verbatim}
\end{tcolorbox}

\caption{System prompt given to the Judge model (top: \textit{Explicit (1)} ; bottom: \textit{Infer (2)} )}
\label{fig:system_prompt_discarded}

\end{figure*}

\section{Frontier models outputs for 24 languages}
\label{sec:front-model-analysis-anex}

As an initial analysis, we examined the outputs of frontier models across seven widely spoken languages and identified a pronounced concentration in their regional references. Specifically, we analyzed the top three most frequently mentioned countries or regions for each language-model pair (see Table \ref{tab:frontier_top3}). Models overwhelmingly anchor their references to the primary country associated with each language, such as the U.S. for English, China for Chinese, France or Canada for French, and Japan for Japanese. This shows that models correctly internalize the conventional linguistic–national associations. Secondary references vary somewhat across models, but still follow intuitive cultural or regional links: for example, Arabic prompts frequently bring up Egypt, Saudi Arabia, and the UAE, while Spanish prompts often reference Mexico, Spain, and Peru.

\begin{table*}[h!]
\centering
\small
\begin{tabular}{@{}l cccccccc@{}}
\toprule
& \multicolumn{8}{c}{\textbf{Models}}  \\

\textbf{Language} &
\llm{openai} &
\llm{gemini} &
\llm{anthropic} &
\llm{llama} &
\llm{Cohere} &
\llm{mistral} &
\llm{Qwen} &
\llm{DeepSeek} \\
\midrule
English (en)  
& \flag{us}\flag{jp}\flag{in} 
& \flag{us}\flag{jp}\flag{in} 
& \flag{jp}\flag{us}\flag{in}
& \flag{jp}\flag{us}\flag{sg}
& \flag{us}\flag{jp}\flag{gb}
& \flag{jp}\flag{us}\flag{in}
& \flag{jp}\flag{us}\flag{in}
& \flag{jp}\flag{us}\flag{in}\\
Chinese (zh) 
& \flag{cn}\flag{us}\flag{jp}
& \flag{cn}\flag{jp}\flag{us}
& \flag{cn}\flag{jp}\flag{us}
& \flag{jp}\flag{sg}\flag{us}
& \flag{cn}\flag{us}\flag{jp}
& \flag{cn}\flag{jp}\flag{us}
& \flag{cn}\flag{jp}\flag{us}
& \flag{cn}\flag{jp}\flag{us}\\
Spanish (es) 
& \flag{mx}\flag{es}\flag{pe}
& \flag{es}\flag{mx}\flag{pe}
& \flag{mx}\flag{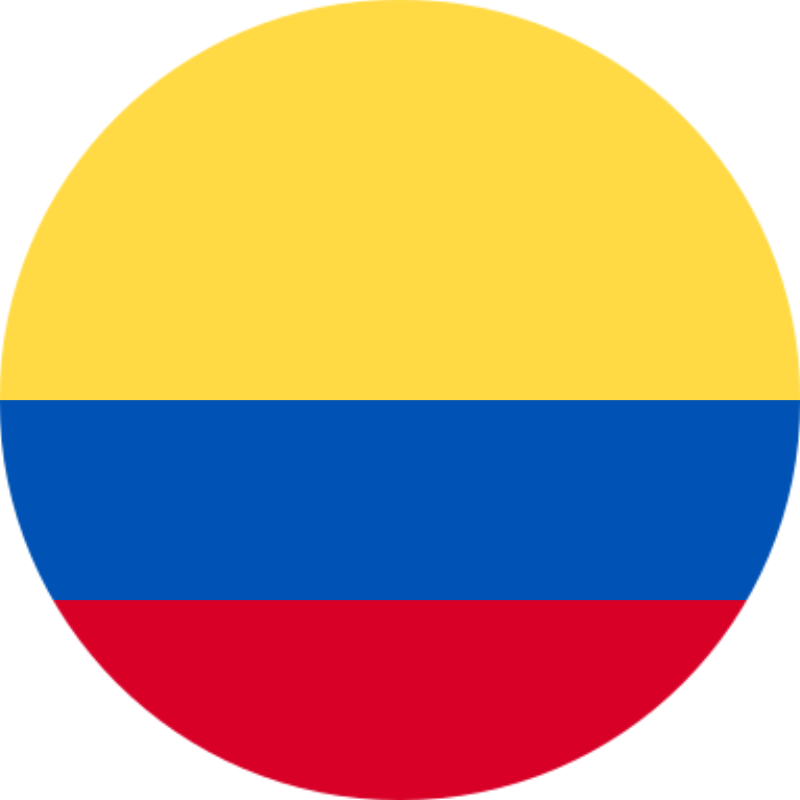}\flag{pe}
& \flag{es}\flag{mx}\flag{jp}
& \flag{es}\flag{us}\flag{jp}
& \flag{mx}\flag{es}\flag{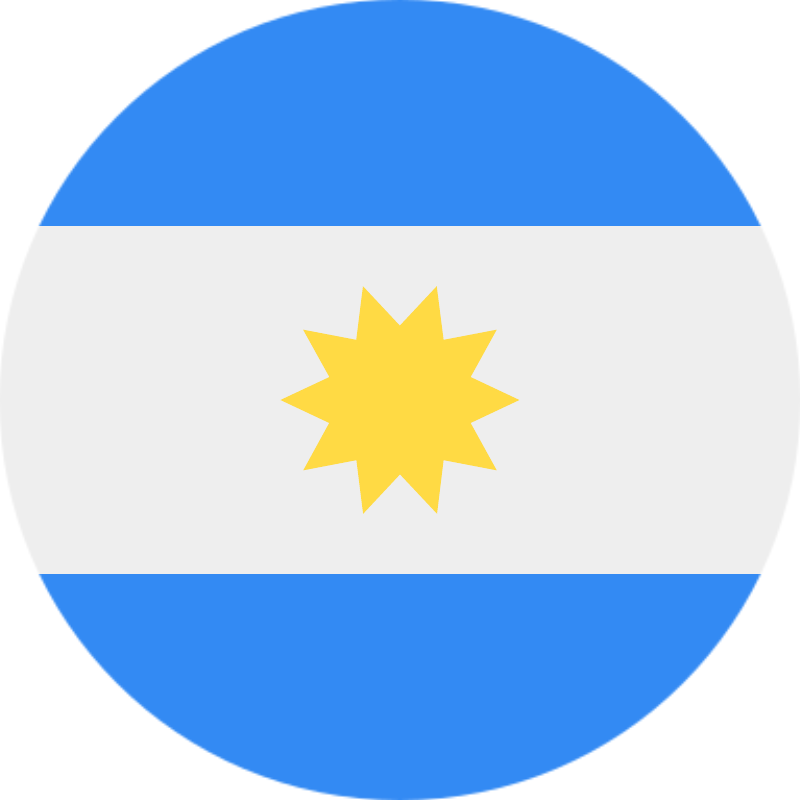}
& \flag{mx}\flag{jp}\flag{es}
& \flag{mx}\flag{es}\flag{jp}\\
Arabic (ar) 
& \flag{eg}\flag{sa}\flag{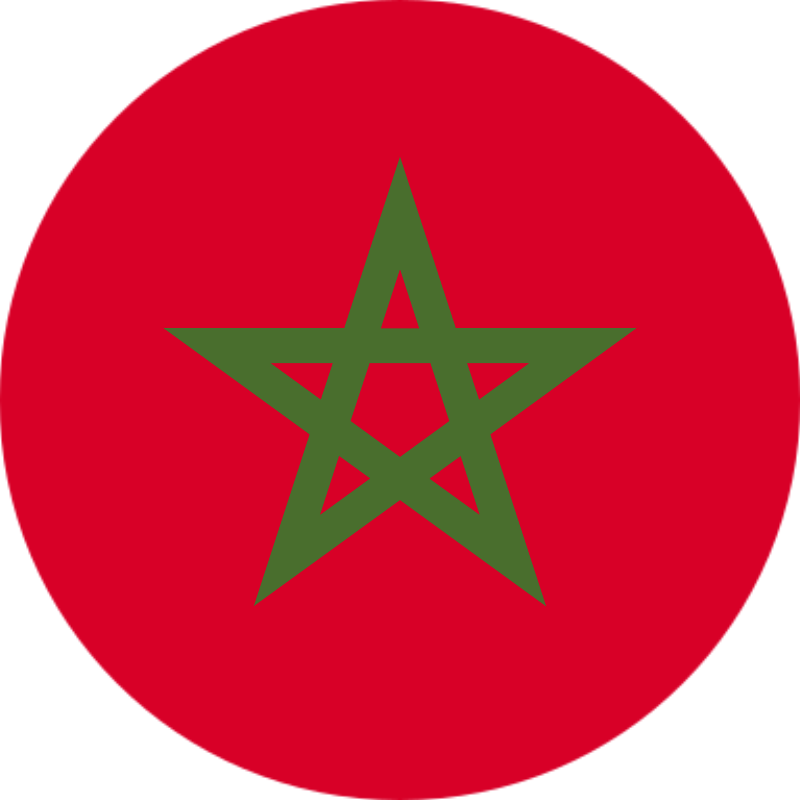}
& \flag{eg}\flag{sa}\flag{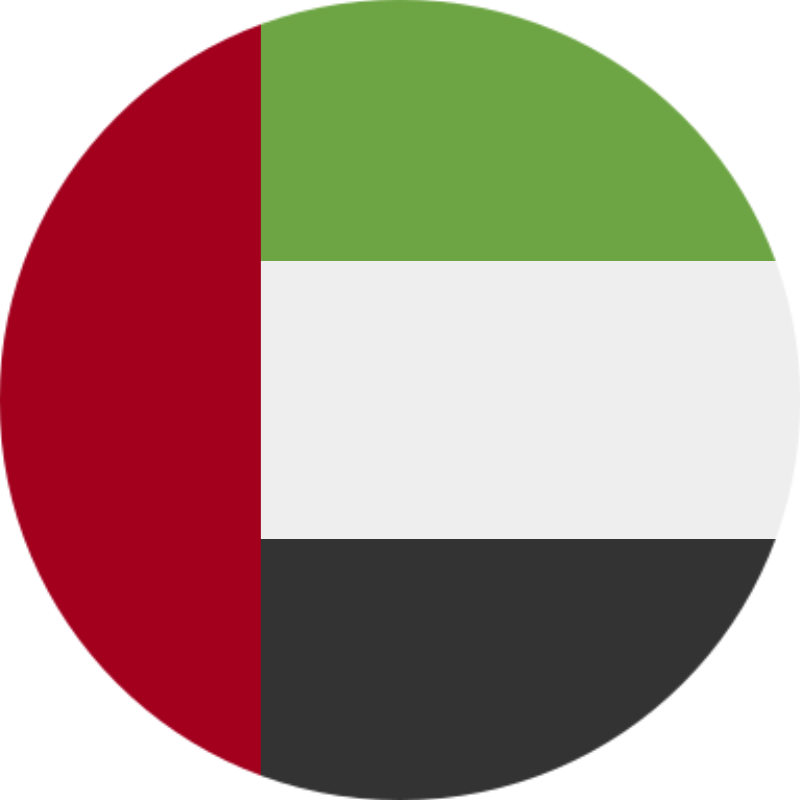}
& \flag{sa}\flag{eg}\flag{ae}
& \flag{ae}\flag{eg}\flag{sa}
& \flag{us}\flag{sa}\flag{eg}
& \flag{ae}\flag{sa}\flag{eg}
& \flag{sa}\flag{eg}\flag{ma}
& \flag{sa}\flag{eg}\flag{ae}\\
French (fr)
& \flag{fr}\flag{us}\flag{it}
& \flag{fr}\flag{jp}\flag{us}
& \flag{fr}\flag{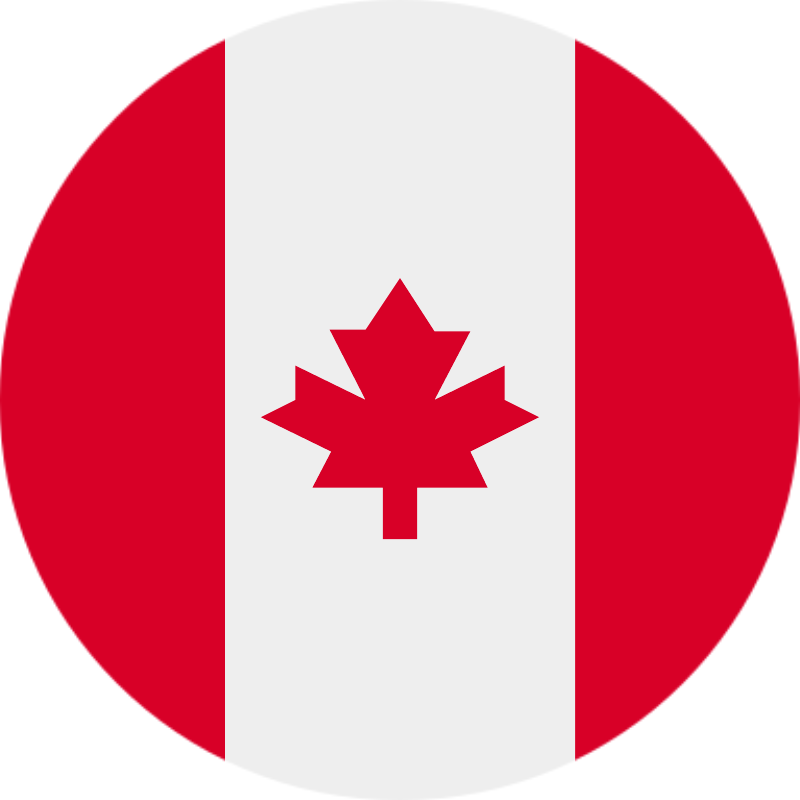}\flag{jp}
& \flag{fr}\flag{jp}\flag{ca}
& \flag{fr}\flag{us}\flag{ca}
& \flag{fr}\flag{jp}\flag{us}
& \flag{fr}\flag{jp}\flag{ca}
& \flag{fr}\flag{jp}\flag{it}\\
Russian (ru)
& \flag{ru}\flag{jp}\flag{us}
& \flag{ru}\flag{jp}\flag{us}
& \flag{ru}\flag{jp}\flag{us}
& \flag{jp}\flag{ru}\flag{us}
& \flag{ru}\flag{us}\flag{jp}
& \flag{ru}\flag{jp}\flag{us}
& \flag{ru}\flag{jp}\flag{us}
& \flag{ru}\flag{jp}\flag{us}\\
Japanese (ja)
& \flag{jp}\flag{us}\flag{in}
& \flag{jp}\flag{us}\flag{in}
& \flag{jp}\flag{in}\flag{us}
& \flag{jp}\flag{id}\flag{in}
& \flag{jp}\flag{us}\flag{gb}
& \flag{jp}\flag{us}\flag{fr}
& \flag{jp}\flag{us}\flag{in}
& \flag{jp}\flag{de}\flag{in}\\
\bottomrule
\end{tabular}
\caption{Top three countries most frequently referenced by model for each high-resourced language.}
\label{tab:frontier_top3}
\end{table*} %% TOP 3 - FRONTIER

\begin{figure*}[t!]
\centering
    \includegraphics[trim=12 120 26 120, clip,width=0.49\linewidth]{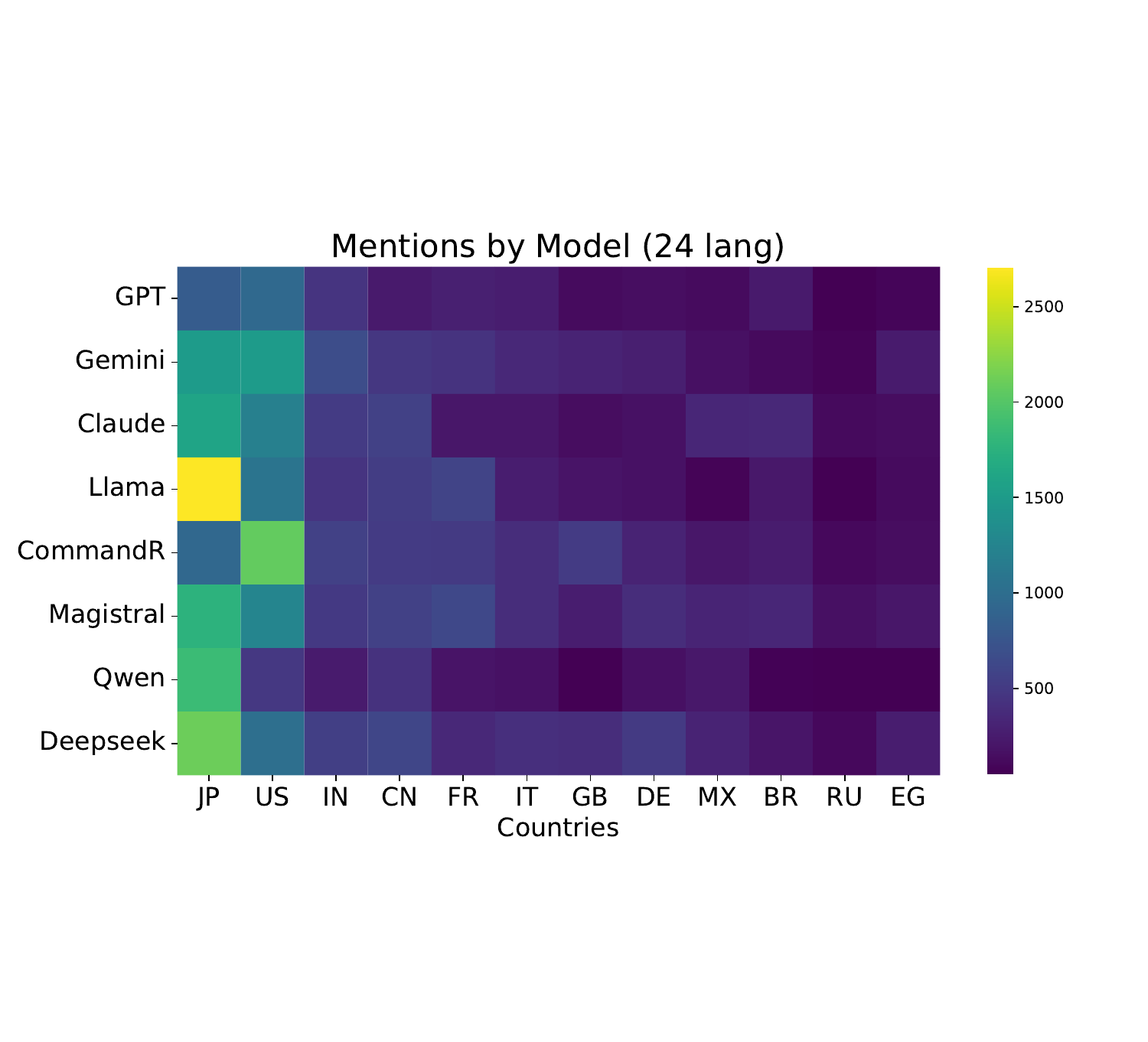}
    \includegraphics[trim=12 100 38 100, clip,width=0.49\linewidth]{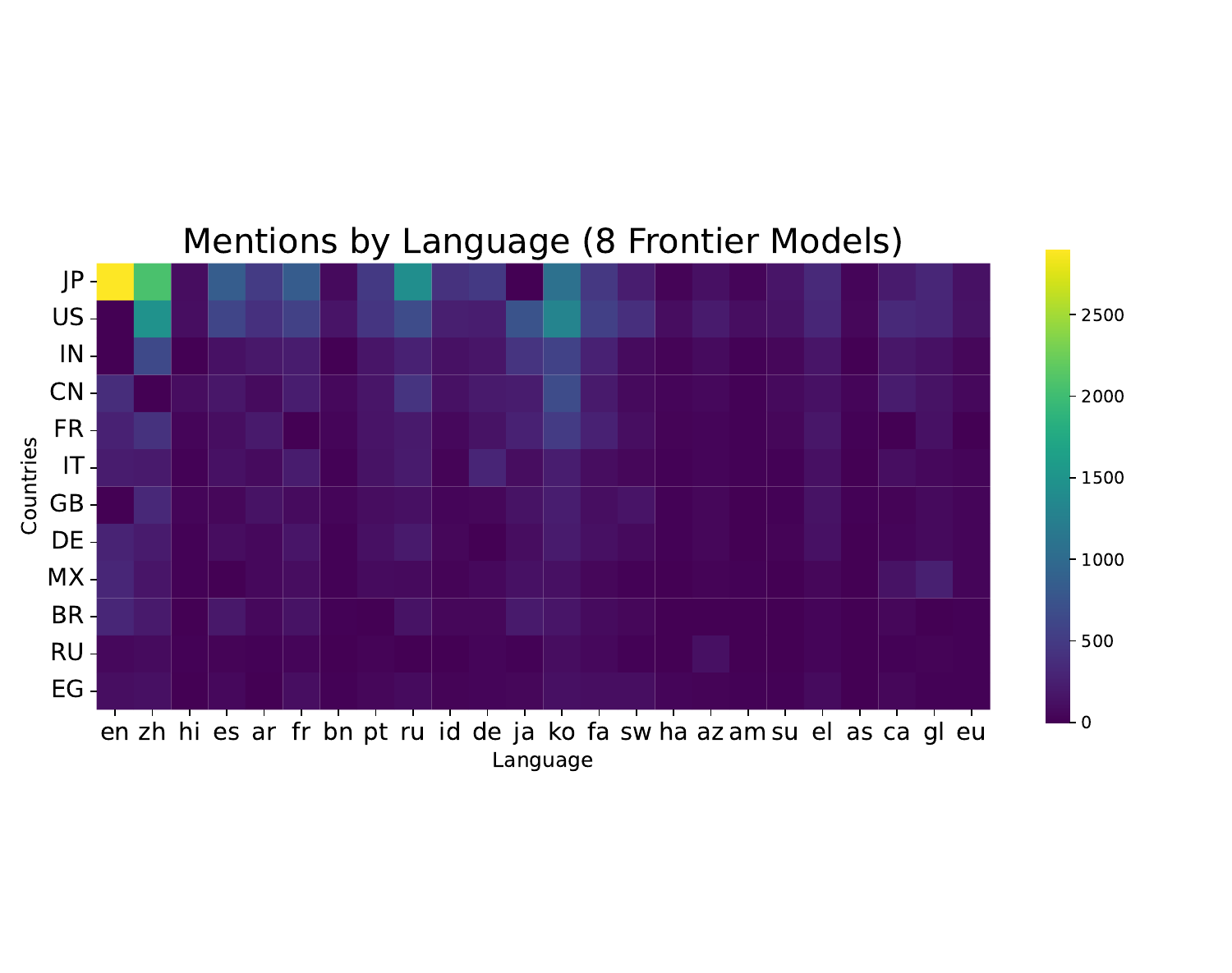} \hfil 
    \caption{Outputs (without \textit{own} references) from gpt-4o-mini, gemini-2.5-flash, claude-3.5-haiku, llama-4-maverick, command-r-08-2024, magistral-small-2506, qwen3-next-80b-a3b-instruct and deepseek-v3.2-exp variants over the full dataset of 24 languages. Countries: Japan (JP), USA (US), India (IN), China (CN), France (FR), Italy (IT),  UK (GB), Germany (DE), Mexico (MX), Brazil (BR), Russia (RU), Egypt (EG). Data from Tables \ref{tab:frontier_out_bymodel} and \ref{tab:frontier_out_bylang} respectively.}
    \label{fig:front-out-hm} \hfil
\end{figure*}

Figure \ref{fig:front-out-hm} presents the outputs (excluding models’ own references) generated by the gpt-4o-mini, gemini-2.5-flash, claude-3.5-haiku, llama-4-maverick, command-r-08-2024, magistral-small-2506, qwen3-next-80b-a3b-instruct, and deepseek-v3.2-exp model variants across the complete dataset covering 24 languages. The evaluation spans content associated with twelve countries: Japan (JP), United States (US), India (IN), China (CN), France (FR), Italy (IT), United Kingdom (GB), Germany (DE), Mexico (MX), Brazil (BR), Russia (RU), and Egypt (EG). The left figure reports country references aggregated by model, highlighting how different model variants distribute references across countries. The right figure reports country references aggregated by language, illustrating how country mentions vary across the linguistic dimension.

\subsection{Analysis by language}
\label{sec:front-model-analysis-anex_lang}

\setlength{\tabcolsep}{3.8pt}%{4.5pt}

\begin{table*}[h!]
\centering
\scriptsize
\begin{tabular}{@{}lrrrrrrrrrrrrrrr|rr@{}}
\toprule
&&&& \flag{JP} & \flag{US} & \flag{IN} & \flag{CN} & \flag{FR} & \flag{IT} & \flag{GB} & \flag{DE} & \flag{MX} & \flag{BR} & \flag{RU} & \flag{EG} & & \\
\textbf{Lang.} & \textbf{Own} (\%) & \textbf{NA} & O+N & JP & US & IN & CN & FR & IT & GB & DE & MX & BR & RU & EG & \textbf{Div} & \textbf{Ent}\\
\midrule
en & 3,223 (.31)& 192   &\textcolor{green}{.32} &\textbf{2,893} & \textit{2,028}* & 695* & 367 & 265 & 223 & \textit{323}* & 277 & 309 & 314 & 57 & 104 & 157 & 0.69\\
zh & 5,110 (.48)& 258   &\textcolor{green}{.51} & \textbf{2,064} & 1,460 & 642 & \textit{4,781}* & 407 & 193 & 319 & 210 & 165 & 202 & 80 & 119 & 142 & 0.57\\
hi & 7,430 (.70)& 1,706 &\textcolor{red}{.87} & 94 & \textbf{104} & \textit{7,423}* & 92 & 37 & 13 & 38 & 21 & 14 & 11 & 14 & 11 & 56 & 0.14\\
es & 6,935 (.66)& 152   &.67 & \textbf{843} & 588 & 130 & 173 & 110 & 128 & 48 & 100 & \textit{2,852}* & 186 & 23 & 58 & 136 & 0.65\\
ar & 7,325 (.69)& 499   &.74 & \textbf{504} & 396 & 182 & 85 & 197 & 82 & 136 & 63 & 59 & 64 & 21 & \textit{1,645}* & 108 & 0.71\\
fr & 5,956 (.56)& 136   &\textcolor{green}{.58} & \textbf{832} & 557 & 215 & 236 & \textit{4,656}* & 225 & 85 & 162 & 95 & 137 & 44 & 104 & 149 & 0.54\\
bn & 8,219 (.79)& 1,189 &\textcolor{red}{.89} & 73 & \textbf{151} & \textit{2,415*} & 59 & 35 & 17 & 41 & 20 & 13 & 16 & 10 & 13 & 68 & 0.20\\
pt & 7,719 (.73)& 358   &.76 & \textbf{479} & 442 & 167 & 161 & 147 & 146 & 94 & 124 & 85 & \textit{6,199}* & 30 & 50 & 114 & 0.42\\
ru & 3,831 (.36)& 173   &\textcolor{green}{.38} & \textbf{1,423} & 657 & 265 & 431 & 199 & 214 & 118 & 196 & 71 & 143 & \textit{3,827}* & 90 & 146 & 0.60\\
id & 8,055 (.76)& 870   &.85 & \textbf{411} & 246 & 129 & 133 & 59 & 24 & 35 & 46 & 29 & 48 & 10 & 30 & 78 & 0.26\\
de & 7,359 (.70)& 71    &.70 & \textbf{485} & 237 & 169 & 203 & 145 & 304 & 48 & \textit{6,927}* & 62 & 53 & 45 & 39 & 122 & 0.38\\
ja & 6,213 (.59)& 375   &.62 & \textit{5,584}* & \textbf{736} & 434 & 219 & 267 & 98 & 144 & 98 & 133 & 193 & 21 & 47 & 114 & 0.36\\
ko & 6,191 (.59)& 1,580 &.74 & 1,064 & \textbf{1,308} & 574 & 663 & 508 & 227 & 229 & 214 & 123 & 160 & 97 & 124 & 112 & 0.47\\
fa & 6,015 (.57)& 722   &.64 & 473 & \textbf{544} & 263 & 194 & 266 & 101 & 107 & 124 & 55 & 89 & 59 & 105 & 117 & 0.48\\
sw & 8,196 (.78)& 144   &.79 & 230 & \textbf{389} & 85 & 79 & 106 & 55 & 156 & 72 & 19 & 48 & 17 & 106 & 89 & 0.50\\
ha & 5,385 (.51)& 3,270 &.82 & 30 & \textbf{99} & 30 & 44 & 25 & 13 & 21 & 16 & 3 & 9 & 1 & 37 & 49 & 0.10\\
az & 7,960 (.75)& 253   &.78 & 123 & \textbf{212} & 81 & 64 & 43 & 34 & 47 & 48 & 23 & 10 & 119 & 29 & 78 & 0.43\\
am & 8,088 (.77)& 1,133 &\textcolor{red}{.87} & 43 & \textbf{106} & 17 & 19 & 13 & 19 & 9 & 6 & 12 & 5 & 8 & 22 & 38 & 0.13\\
su & 8,331 (.79)& 995   &\textcolor{red}{.88} & \textbf{167} & 146 & 49 & 76 & 48 & 14 & 15 & 32 & 10 & 7 & 6 & 18 & 62 & 0.11\\
el & 7,393 (.70)& 253   &.72 & \textbf{330} & 307 & 161 & 135 & 173 & 124 & 141 & 132 & 46 & 36 & 34 & 83 & 106 & 0.36\\
as & 8,732 (.83)& 140   &.84 & 41 & \textbf{53} & \textit{2,900}* & 42 & 18 & 6 & 17 & 8 & 2 & 4 & 4 & 6 & 41 & 0.30\\
ca & 7,622 (.72)& 158   &.74 & 213 & \textbf{339} & 170 & 224 & \textit{306}* & 103 & 31 & 45 & 142 & 50 & 17 & 51 & 121 & 0.49\\
gl & 7,967 (.75)& 88    &.76 & \textbf{314} & 297 & 129 & 143 & 135 & 58 & 77 & 68 & 259 & 292 & 23 & 21 & 113 & 0.37\\
eu & 8,266 (.78)& 903   &\textcolor{red}{.87} & 132 & \textbf{144} & 48 & 66 & \textit{674}* & 37 & 39 & 40 & 38 & 15 & 15 & 13 & 68 & 0.37\\
\midrule

Avg. & 6,980 (.66)& 651 &.72 & \textbf{577} & 414 & 197 & 170 & 153 & 102 & 87 & 92 & 77 & 82 & 33 & 56 & 99 & 0.40\\

\bottomrule
\end{tabular}
\caption{Frontier Model outputs for 24 languages. \textbf{Own}: counts references to countries in which the language is an official language. \textbf{NA}: denotes missing responses. \textbf{O+N}: percentage of the sum of Own and NA references relative to the total for the language. Referenced countries: Japan (JP), USA (US), India (IN), China (CN), France (FR), Italy (IT), United Kingdom (GB), Germany (DE), Mexico (MX), Brazil (BR), Russia (RU), Egypt (EG). \textbf{Div}: diversity scores. \textbf{Ent}: normalized entropy values. Bold values indicate the highest count for each language, excluding self-references. An asterisk (*) marks references to own countries.}
\label{tab:frontier_out_bylang}
\end{table*}

\setlength{\tabcolsep}{6pt}

The results in Table \ref{tab:frontier_out_bylang} illustrate consistent patterns in how frontier models associate languages with countries. Across all 24 languages, references to countries where the language is an official language (\textit{Own}) dominate the model’s outputs. This behaviour suggests a heavy reliance on language–country co-occurrence signals, likely reflecting strong correlations within the training data. While such associations are linguistically plausible, their prevalence indicates limited decoupling between a language and its primary national context.

Beyond these self-referential patterns, the model exhibits a stable hierarchy regarding non-official country references. Japan and United States are the most frequently referenced nations across all languages, followed by India, France, and China. Importantly, these references appear even in languages with limited cultural or geographic ties to these nations, suggesting that cross-country mentions are driven more by general global prominence than by language-specific relevance. This finding aligns with prior observations that multilingual models tend to overrepresent culturally salient entities regardless of the linguistic context.

Examining the \textit{Own} and \textit{NA} columns reveals distinct behaviours based on language resource availability. The \textit{Own} count is highest for low-resource languages such as Assamese, Sundanese, and Bengali, indicating a strong tendency for the model to ground these languages almost exclusively in their home countries. In contrast, high-resource languages like English, Chinese, and Russian show comparatively lower \textit{Own} counts, reflecting the model’s ability to distribute references more broadly. Additionally, while \textit{NA} (missing response) counts are generally low, spikes in languages like Hausa and Hindi indicate occasional gaps in model generation for specific low-resource contexts.

The \textit{O+N} column, which aggregates \textit{Own} and \textit{NA} percentages, further highlights this divide. Languages such as Bengali, Sundanese, Amharic, Basque, and Hindi exhibit the highest \textit{O+N} values, reflecting a concentration of outputs that are either biased toward the home country or result in non-responses—traits characteristic of low-resource processing. Conversely, high-resource languages like English, French, and Russian display lower \textit{O+N} values. This indicates a more diversified output distribution, consistent with languages that possess richer cross-country associations in the training corpus. Overall, a clear trend emerges: low-resource languages tend to yield highly concentrated or missing responses (high \textit{O+N}), while high-resource languages yield more balanced references reflecting broader cultural associations.

Finally, the analysis of the relationship between normalized diversity and entropy scores may indicate a correlation with language resource availability. In this context, diversity measures how many unique countries are referenced at all, while normalized entropy captures the evenness of the distribution of mentions across those referenced entities. High-resource languages (e.g., English, Chinese, French, Russian) exhibit both high diversity and moderate-to-high entropy, indicating that they not only reference a wider array of nations but also distribute their focus more uniformly. In contrast, low-resource languages (e.g., Hindi, Bengali, Amharic, Sundanese) demonstrate low diversity and entropy; they reference fewer unique countries, and those references are heavily concentrated on a single entity (typically \textit{own} countries) rather than being spread out. This suggests that the model’s capacity to generate broad, evenly distributed cultural references is heavily conditioned by the availability of the underlying training data.

\subsection{Measuring Cultural Diversity in Model Outputs}
\label{sec:div_anex}

Table \ref{tab:model_div} reports diversity scores, the number of distinct geographic regions referenced across 24 languages and eight frontier models. A key aspect of this metric is that it measures model diversity across different regions per language, meaning that for each language in our set, we calculate how many unique countries or regions are represented in the model's cultural references. A higher score indicates that a model's outputs for a given language draw from a broader, more globally varied set of cultural contexts, rather than being concentrated in one or two regions.

Averaged over languages, we observe large model-level differences: CommandR achieves the highest overall diversity, followed by DeepSeek and Gemini, while Mistral, GPT, and Qwen exhibit substantially lower scores. Language-level variation is also pronounced. High-resource languages such as English, French, Russian, and Spanish consistently receive broadest cultural coverage across models. In contrast, several low-resource or regionally concentrated languages (including Amharic, Assamese, Hausa, Hindi, and Bengali) remain well below the global mean, indicating narrower cultural framing in model outputs.

Beyond aggregate trends, we find substantial cross-model variance within the same language, suggesting that cultural diversity is not determined by language alone. For example, Chinese ranges from 98 (Qwen) to 188 (CommandR), Bengali from 20 (Qwen) to 116 (LLaMA), and Hindi from 18 (Claude) to 107 (DeepSeek). Such differences often exceed average inter-language gaps, underscoring the impact of model-specific training and alignment choices. Overall, these results show that cultural diversity in multilingual generation is highly uneven, favoring high-resource languages, but also that these disparities are model-dependent rather than unavoidable.

\begin{table}[h!]
\centering
\scriptsize
\setlength{\tabcolsep}{4pt} 
\begin{tabular}{@{}lrrrrrrrr|r@{}}
\toprule
Language &
\llmb{openai} &
\llmb{gemini} &
\llmb{anthropic} &
\llmb{llama} &
\llmb{Cohere} &
\llmb{mistral} &
\llmb{Qwen} &
\llmb{DeepSeek} & AVG \\

\midrule
en & \hl{160} & \textbf{\hl{171}} & \hl{159} & 135 & \hl{170} & 145 & 152 & \hl{165} & \hl{157} \\
zh & 127 & \hl{165} & 125 & \hl{157} & \textbf{\hl{188}} & 137 & 98  & \hl{143} & \hl{142} \\
hi & 33  & 69  & 18  & 58  & 52  & 60  & 47  & \textbf{107} & 56  \\
es & 112 & \hl{131} & 127 & \hl{141} & \hl{149} & 121 & 134 & \textbf{\hl{169}} & \hl{136} \\
ar & 89  & \hl{126} & 88  & 101 & \textbf{\hl{137}} & 95  & \hl{110} & 115 & \hl{108} \\
fr & 143 & \hl{154} & 135 & 130 & \textbf{\hl{174}} & 143 & 137 & \hl{173} & \hl{149} \\
bn & 36  & 75  & 28  & \textbf{\hl{116}} & 115 & 64  & 20  & 90  & 68  \\
pt & 91  & \hl{134} & 105 & 78  & \textbf{\hl{143}} & \hl{117} & 109 & \hl{139} & \hl{114} \\
ru & 141 & \hl{156} & \hl{147} & \hl{149} & \textbf{\hl{179}} & 120 & 118 & \hl{158} & \hl{146} \\
id & 72  & \textbf{\hl{118}} & 89  & 56  & 53  & 56  & 62  & \hl{115} & 78  \\
de & 94  & 113 & \textbf{\hl{150}} & 118 & \hl{137} & 119 & \hl{132} & 112 & \hl{122} \\
ja & 90  & \hl{127} & \hl{129} & \hl{127} & \textbf{\hl{130}} & 94  & 92  & \hl{123} & \hl{114} \\
ko & 73  & \hl{120} & \hl{145} & 99  & 99  & \textbf{\hl{149}} & 75  & \hl{137} & \hl{112} \\
fa & 98  & \hl{142} & 91  & 101 & \textbf{\hl{167}} & 115 & 88  & \hl{136} & \hl{117} \\
sw & 68  & 109 & 87  & \hl{110} & \textbf{\hl{134}} & 85  & 46  & 75  & 89  \\
ha & 34  & 74  & 31  & \textbf{\hl{86}}  & 34  & 11  & 35  & \textbf{\hl{86}}  & 49  \\
az & 77  & \textbf{109} & 56  & 79  & 77  & 61  & 55  & \textbf{\hl{109}} & 78  \\
am & 19  & 53  & 43  & \textbf{74}  & 43  & 11  & 25  & 33  & 38  \\
su & 41  & 77  & 64  & 30  & \textbf{100} & 82  & 26  & 77  & 62  \\
el & 87  & 100 & \hl{109} & 101 & \textbf{\hl{162}} & 91  & 75  & \hl{127} & \hl{106} \\
as & 31  & 30  & 22  & \textbf{97}  & 30  & 30  & 23  & 64  & 41  \\
ca & 102 & 115 & \hl{140} & \hl{126} & \textbf{\hl{171}} & 97  & 105 & 114 & \hl{121} \\
gl & 96  & \hl{120} & 109 & 88  & \textbf{\hl{163}} & 89  & 100 & \hl{140} & \hl{113} \\
eu & 55  & \textbf{\hl{115}} & 94  & 83  & 64  & 28  & 36  & 73  & 68  \\
\midrule
ALL & 82  & \hl{113} & 95  & \hl{102} & \hl{120} & 88  & 79  & \hl{116} & 99  \\
\bottomrule
\end{tabular}
\caption{Diversity scores by language for different Frontier API models. ALL: per-model average across languages. AVG: per-language average. Highlighted results: above average results.}
\label{tab:model_div}
\end{table}

\subsection{Topic Analysis in Model Outputs}
\label{sec:top_anex}

Table \ref{fig:to_by_taxonomy_annex} presents the countries most frequently mentioned by the models across the 11 general topics in our dataset. It reports the corresponding counts for these leading countries, as well as the models’ own-country mentions.

% Add this macro to your preamble or just above the table:
\newcommand{\flagcnt}[2]{\begin{tabular}[t]{@{}c@{}}\flag{#1}\\ \small #2\end{tabular}}

\begin{table*}[h!]
\centering
\small % Keeps the wider row layout within page boundaries
\begin{tabular}{@{}l l r@{}}
\toprule
 \textbf{Topic} & \textbf{Top-10 Countries} & \textbf{Own} \\
\midrule
 \topics{t1} 
    1. Beliefs
    & \flag{jp}\,1,506 \flag{gr}\,586 \flag{in}\,549 \flag{us}\,400 \flag{cn}\,332 \flag{sa}\,318 \flag{eg}\,282 \flag{il}\,278 \flag{it}\,189 \flag{fr}\,147 
    & 10,801 \\
    
 \topics{t2} 
    2. Social
    & \flag{jp}\,1,594 \flag{us}\,455 \flag{il}\,145 \flag{it}\,144 \flag{ps}\,126 \flag{cn}\,117 \flag{de}\,117 \flag{fr}\,104 \flag{in}\,85 \flag{gb}\,79 
    & 11,571 \\
    
 \topics{t3}
    3. Education 
    & \flag{jp}\,1,086 \flag{us}\,453 \flag{gr}\,268 \flag{cn}\,267 \flag{fr}\,169 \flag{in}\,165 \flag{de}\,157 \flag{eg}\,153 \flag{it}\,149 \flag{gb}\,136
    & 12,279 \\
    
 \topics{t4} 
    4. Arts 
    & \flag{jp}\,2,028 \flag{us}\,1,191 \flag{in}\,768 \flag{fr}\,621 \flag{it}\,516 \flag{br}\,381 \flag{cn}\,337 \flag{mx}\,304 \flag{id}\,263 \flag{es}\,262
    & 23,032 \\

 \topics{t5} 
    5. Food
    & \flag{jp}\,1,980 \flag{us}\,515 \flag{in}\,413 \flag{mx}\,344 \flag{cn}\,340 \flag{it}\,309 \flag{br}\,270 \flag{es}\,208 \flag{fr}\,175 \flag{id}\,137
    & 16,007 \\

 \topics{t6} 
    6. Geography
    & \flag{us}\,1,494 \flag{jp}\,822 \flag{in}\,563 \flag{cn}\,471 \flag{fr}\,421 \flag{it}\,368 \flag{mx}\,366 \flag{br}\,341 \flag{de}\,302 \flag{il}\,264
    & 17,591 \\

 \topics{t7} 
    7. Politics
    & \flag{us}\,1,344 \flag{cn}\,730 \flag{jp}\,589 \flag{fr}\,395 \flag{gb}\,362 \flag{de}\,360 \flag{in}\,287 \flag{sa}\,247 \flag{ru}\,174 \flag{kp}\,164
    & 17,585 \\
    
 \topics{t8} 
    8. Health
    & \flag{jp}\,1,068 \flag{us}\,361 \flag{cn}\,220 \flag{in}\,210 \flag{id}\,159 \flag{se}\,111 \flag{th}\,83 \flag{eu}\,70 \flag{br}\,64 \flag{pe}\,59
    & 10,474 \\
    
 \topics{t9} 
    9. Media
    & \flag{jp}\,1,330 \flag{us}\,1,304 \flag{kr}\,391 \flag{gb}\,272 \flag{cn}\,200 \flag{fr}\,194 \flag{de}\,167 \flag{br}\,116 \flag{in}\,109 \flag{eu}\,103
    & 17,112 \\

 \topics{t10} 
    10. History 
    & \flag{us}\,1,024 \flag{fr}\,761 \flag{in}\,521 \flag{jp}\,455 \flag{it}\,398 \flag{gb}\,362 \flag{cn}\,348 \flag{gr}\,341 \flag{de}\,332 \flag{eg}\,245
    & 14,916 \\

 \topics{t11} 
    11. Economy
    & \flag{us}\,1,388 \flag{jp}\,803 \flag{cn}\,546 \flag{eu}\,373 \flag{sg}\,323 \flag{de}\,319 \flag{in}\,270 \flag{kr}\,226 \flag{gb}\,205 \flag{fr}\,191  
    & 16,188 \\
    
\bottomrule
\end{tabular}
    \caption{Top 10 countries most frequently referenced by all Frontier API models for each topic in the taxonomy for 24 languages. Raw counts for each top country as well as own country mentions. \textbf{Own}: counts references to countries in which the language is an official language.}
    \label{fig:to_by_taxonomy_annex}
\end{table*}

On a model-by-model analysis of each topic (Table \ref{fig:topic_by_model_annex}), the systemic bias emerges once again across all eight frontier AI models, with Japan and the United States dominating as the most frequently referenced countries across the eleven evaluated topics. Specifically, Japan consistently takes the top position in culturally and socially oriented domains such as Beliefs, Education, Arts, Food, and Health, while the United States leads in structural and global topics like Geography, Politics, History, and the Economy. Notably, this US-Japan centricity represents a strong cross-model consensus, appearing almost equally in Western models like OpenAI, Gemini, and Anthropic, as well as in Chinese models like Qwen and DeepSeek. Other countries, including India, China, and France, appear only sporadically as secondary or tertiary references, highlighting a profound homogenization in the ways current LLMs exhibit biases toward particular countries and regions.

% Add this macro to your preamble or just above the table:
%\newcommand{\flagcnt}[2]{\begin{tabular}[t]{@{}c@{}}\flag{#1}\\ \small #2\end{tabular}}

\begin{table*}[h!]
\centering
\small % Keeps the wider row layout within page boundaries
\begin{tabular}{@{}l cccccccc r@{}}
\toprule
 \textbf{Topic} & 

\llmb{openai} &
\llmb{gemini} &
\llmb{anthropic} &
\llmb{llama} &
\llmb{Cohere} &
\llmb{mistral} &
\llmb{Qwen} &
\llmb{DeepSeek} \\
 
\midrule
 \topics{t1} 
    1. Beliefs
    & \flag{jp}\flag{in}\flag{gr}
    & \flag{jp}\flag{in}\flag{gr}
    & \flag{jp}\flag{in}\flag{gr}
    & \flag{jp}\flag{gr}\flag{cn}
    & \flag{gr}\flag{us}\flag{in}
    & \flag{jp}\flag{in}\flag{gr}
    & \flag{jp}\flag{sa}\flag{gr}
    & \flag{jp}\flag{gr}\flag{in}
    \\
    
 \topics{t2} 
    2. Social
    & \flag{jp}\flag{us}\flag{it}
    & \flag{jp}\flag{us}\flag{il}
    & \flag{jp}\flag{us}\flag{il}
    & \flag{jp}\flag{us}\flag{sg}
    & \flag{jp}\flag{us}\flag{gb}
    & \flag{jp}\flag{us}\flag{it}
    & \flag{jp}\flag{it}\flag{cn}
    & \flag{jp}\flag{us}\flag{it}
    \\
    
 \topics{t3}
    3. Education 
    & \flag{jp}\flag{us}\flag{br} 
    & \flag{jp}\flag{us}\flag{gr}
    & \flag{jp}\flag{cn}\flag{us}
    & \flag{jp}\flag{us}\flag{fr}
    & \flag{us}\flag{jp}\flag{cn}
    & \flag{jp}\flag{us}\flag{fr}
    & \flag{jp}\flag{cn}\flag{us}
    & \flag{jp}\flag{gr}\flag{us}
    \\
    
 \topics{t4} 
    4. Arts 
    & \flag{us}\flag{jp}\flag{in} 
    & \flag{jp}\flag{us}\flag{in}
    & \flag{jp}\flag{us}\flag{br}
    & \flag{jp}\flag{us}\flag{fr}
    & \flag{us}\flag{jp}\flag{in}
    & \flag{jp}\flag{fr}\flag{us}
    & \flag{jp}\flag{fr}\flag{cn}
    & \flag{jp}\flag{us}\flag{in}
    \\

 \topics{t5} 
    5. Food
    & \flag{jp}\flag{in}\flag{it}
    & \flag{jp}\flag{in}\flag{us}
    & \flag{jp}\flag{cn}\flag{mx}
    & \flag{jp}\flag{us}\flag{cn}
    & \flag{us}\flag{jp}\flag{in}
    & \flag{jp}\flag{us}\flag{mx}
    & \flag{jp}\flag{mx}\flag{cn}
    & \flag{jp}\flag{mx}\flag{in}
    \\

 \topics{t6} 
    6. Geography
    & \flag{us}\flag{in}\flag{br}
    & \flag{us}\flag{jp}\flag{in}
    & \flag{us}\flag{jp}\flag{in}
    & \flag{us}\flag{jp}\flag{cn}
    & \flag{us}\flag{fr}\flag{jp}
    & \flag{us}\flag{jp}\flag{fr}
    & \flag{us}\flag{jp}\flag{cn}
    & \flag{us}\flag{jp}\flag{in}
    \\

 \topics{t7} 
    7. Politics
    & \flag{us}\flag{cn}\flag{fr}
    & \flag{us}\flag{cn}\flag{jp}
    & \flag{us}\flag{jp}\flag{cn}
    & \flag{us}\flag{jp}\flag{cn}
    & \flag{us}\flag{gb}\flag{cn}
    & \flag{us}\flag{cn}\flag{fr}
    & \flag{cn}\flag{jp}\flag{us}
    & \flag{us}\flag{cn}\flag{jp}
    \\
    
 \topics{t8} 
    8. Health
    & \flag{jp}\flag{in}\flag{us}
    & \flag{jp}\flag{us}\flag{in}
    & \flag{jp}\flag{cn}\flag{in}
    & \flag{jp}\flag{us}\flag{id}
    & \flag{us}\flag{jp}\flag{cn}
    & \flag{jp}\flag{us}\flag{in}
    & \flag{jp}\flag{cn}\flag{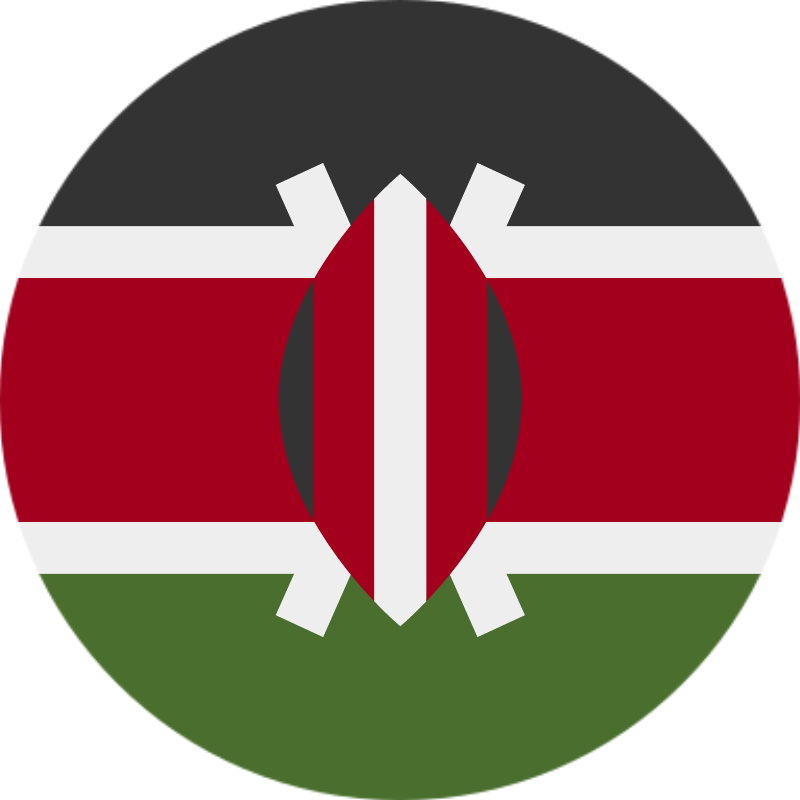}
    & \flag{jp}\flag{id}\flag{in}
    \\
    
 \topics{t9} 
    9. Media
    & \flag{us}\flag{jp}\flag{kr}
    & \flag{us}\flag{jp}\flag{kr}
    & \flag{jp}\flag{us}\flag{kr}
    & \flag{jp}\flag{us}\flag{fr}
    & \flag{us}\flag{gb}\flag{jp}
    & \flag{jp}\flag{us}\flag{gb}
    & \flag{jp}\flag{us}\flag{kr}
    & \flag{jp}\flag{us}\flag{kr}
    \\

 \topics{t10} 
    10. History 
    & \flag{us}\flag{fr}\flag{in}
    & \flag{us}\flag{fr}\flag{in}
    & \flag{us}\flag{in}\flag{gr}
    & \flag{fr}\flag{us}\flag{jp}
    & \flag{us}\flag{fr}\flag{in}
    & \flag{us}\flag{fr}\flag{jp}
    & \flag{jp}\flag{us}\flag{fr}
    & \flag{us}\flag{it}\flag{fr}
    \\

 \topics{t11} 
    11. Economy
    & \flag{us}\flag{jp}\flag{cn}
    & \flag{us}\flag{jp}\flag{eu}
    & \flag{us}\flag{jp}\flag{cn}
    & \flag{jp}\flag{us}\flag{sg}
    & \flag{us}\flag{jp}\flag{cn}
    & \flag{us}\flag{jp}\flag{cn}
    & \flag{us}\flag{jp}\flag{cn}
    & \flag{us}\flag{jp}\flag{cn}
    \\
    
\bottomrule
\end{tabular}
    \caption{Top three countries most frequently referenced by Frontier API models for each topic in the taxonomy for 24 languages, excluding own country mentions.}
    \label{fig:topic_by_model_annex}
\end{table*}

\subsection{Correlation between CC and output diversity}
\label{sec:cc-div_anex}

We examined, across all frontier models, the relationship between each language’s proportion of Common Crawl (CC) data and the diversity of outputs generated in that language for each model. Table \ref{tab:model_div_stats} reports Pearson's $r$ and Spearman's $r_s$ correlations, along with their associated p-values. 

\begin{table}[h!]
\centering
\scriptsize
\begin{tabular}{@{}lllll@{}}
\toprule
  & \multicolumn{2}{l}{\textbf{Pearson}} & \multicolumn{2}{l}{\textbf{Spearman}}\\
\textbf{Model} & \textbf{$r$} & \textbf{$p$} & \textbf{$r_s$} & \textbf{$p$} \\
\midrule
\llmb{Qwen} Qwen & 0.529 & 0.008 & 0.808 & $1.80\times10^{-6}$ \\
\llmb{gemini} Gemini & 0.483 & 0.017 & 0.798 & $3.00\times10^{-6}$ \\
\llmb{DeepSeek} DeepSeek & 0.425 & 0.038 & 0.775 & $8.90\times10^{-6}$ \\
\llmb{openai} GPT & 0.577 & 0.003 & 0.774 & $9.14\times10^{-6}$ \\
\llmb{mistral} Magistral & 0.426 & 0.038 & 0.770 & $1.07\times10^{-5}$ \\
\llmb{anthropic} Claude & 0.452 & 0.027 & 0.760 & $1.63\times10^{-5}$ \\
\llmb{llama} Llama & 0.383 & 0.064 & 0.650 & $5.84\times10^{-4}$ \\
\llmb{Cohere} Command-r & 0.329 & 0.116 & 0.629 & $9.87\times10^{-4}$ \\
Avg\_all & 0.498 & 0.013 & 0.843 & $2.31\times10^{-7}$ \\
\bottomrule
\end{tabular}
\caption{Correlation between CommonCrawl percentage and output diversity for Frontier models. Pearson's $r$ and Spearman's $r_s$ are reported along with their p-values.}
\label{tab:model_div_stats}
\end{table}

Across all models combined (Avg\_all), we observed a Pearson correlation of 0.498 ($p=0.013$), indicating a moderate positive linear relationship, and a Spearman correlation of 0.843 ($p<10^{-6}$), showing a very strong rank correlation. These results suggest that languages with a higher proportion of CC data tend to exhibit greater diversity in model outputs.

When examining individual models, the correlations varied but largely supported this trend. For example, GPT-4o-mini showed the strongest Pearson correlation (0.577, $p=0.003$) and a high Spearman correlation (0.774, $p<10^{-5}$), indicating both linear and rank-based relationships between data availability and diversity. Qwen3-next-80b-a3b-instruct also displayed high correlations (Pearson 0.529, Spearman 0.808), while Gemini-2.5-flash and Claude-3.5-haiku exhibited moderate correlations. Deepseek-v3.2-exp and Magistral-small-2506 presented slightly lower but still statistically significant correlations, reinforcing the overall pattern.  

Two models, however, showed weaker correlations. LLaMA-4-maverick had a Pearson correlation of 0.383 ($p=0.064$) and Spearman 0.650 ($p=5.84\times10^{-4}$), while Command-R-08-2024 showed Pearson 0.329 ($p=0.116$) and Spearman 0.629 ($p=9.87\times10^{-4}$). Although the Spearman correlations remain significant, the lower Pearson correlations suggest that the linear relationship is less pronounced for these models, possibly reflecting model-specific factors or limitations in handling low-resource languages. 

Overall, these results indicate that the amount of possibly available training data for each language may be an important factor influencing output diversity. Languages with less CC data tend to yield less diverse outputs, suggesting a potential reduction in creative or varied responses when data is scarce. This pattern is robust across most models, highlighting the influence of data coverage on multilingual model behaviour.

\section{Open-Weight Base and Instruct Models Analysis Over English Dataset}
\label{sec:base-ins-anex}

\begin{table*}[t]
\centering
\small
\setlength{\tabcolsep}{5pt}
\begin{tabular}{llrrrrrrrrrrrrrrr}
\toprule
&&& \flag{jp} & \flag{us} & \flag{in} & \flag{cn} & \flag{fr} & \flag{it} & \flag{gb} & \flag{de} 
& \flag{mx} & \flag{br} & \flag{ru} & \flag{eg} \\
\textbf{Model} & & \textbf{NA} & JP & US & IN & CN & FR & IT & GB & DE & MX & BR & RU & EG & \textbf{Div} & \textbf{Ent}\\
\midrule
\multirow{2}{*}{Llama-3.1-8B}
  & Base & 354 & 164 & 144 & \textbf{200} & 85 & 87 & 106 & 30 & 62 & 42 & 82 & 54 & 17 & 109 & 0.78\\
  & Inst. &  14 & \textbf{407} & 391 & 121 & 25 & 18 & 22 & 55 & 14 & 10 & 32 & 9 & 14 & 174 & 0.66\\
\multirow{2}{*}{Llama-3.1-70B}
 & Base & 783 & \textbf{237} & 194 & 199 & 116 & 85 & 61 & 36 & 80 & 37 & 85 & 51 & 40 & 117 & 0.79\\
 & Inst. & 13 & \textbf{417} & 333 & 125 & 35 & 15 & 9 & 42 & 7 & 15 & 19 & 2 & 15 & 174 & 0.65\\
 \midrule
\multirow{2}{*}{gemma-2-9b}
 & Base & 30 & 124 & \textbf{317} & 169 & 80 & 53 & 26 & 18 & 45 & 43 & 26 & 12 & 25 & 95 & 0.70\\
 & Inst. & 19 & \textbf{269} & 217 & 104 & 18 & 33 & 41 & 69 & 29 & 14 & 55 & 2 & 29 & 136 & 0.71\\
\multirow{2}{*}{gemma-2-27b}
 & Base & 28 & 100 & \textbf{202} & 128 & 71 & 38 & 43 & 75 & 31 & 23 & 33 & 24 & 29 & 80 & 0.77\\
 & Inst. & 9 & 250 & \textbf{285} & 58 & 23 & 16 & 27 & 30 & 26 & 31 & 49 & 1 & 14 & 154 & 0.73\\
\midrule
\multirow{2}{*}{Mistral-7B}
 & Base & 52 & 98 & \textbf{146} & 93 & 73 & 41 & 38 & 41 & 46 & 45 & 28 & 40 & 31 & 91 & 0.81\\
 & Inst. & 14 & 233 & \textbf{580} & 86 & 34 & 18 & 45 & 57 & 14 & 22 & 58 & 1 & 11 & 160 & 0.63\\
 \midrule
\multirow{2}{*}{Qwen2.5-7B}
 & Base & 265 & 98 & \textbf{385} & 98 & 32 & 51 & 64 & 34 & 35 & 18 & 34 & 9 & 7 & 147 & 0.71\\
 & Inst. & 16 & \textbf{321} & 304 & 74 & 310 & 15 & 50 & 20 & 21 & 5 & 11 & 5 & 10 & 150 & 0.62\\
\multirow{2}{*}{Qwen2.5-32B}
 & Base & 248 & 183 & \textbf{399} & 131 & 35 & 36 & 19 & 39 & 15 & 31 & 15 & 8 & 5 & 142 & 0.65\\
 & Inst. & 12 & \textbf{489} & 247 & 58 & 140 & 37 & 18 & 24 & 17 & 11 & 25 & 4 & 7 & 118 & 0.61\\
\multirow{2}{*}{Qwen2.5-72B}
 & Base & 182 & 180 & \textbf{606} & 150 & 67 & 31 & 28 & 55 & 29 & 34 & 40 & 6 & 25 & 144 & 0.66\\
 & Inst. & 10 & 343 & \textbf{386} & 106 & 192 & 31 & 22 & 15 & 29 & 6 & 16 & 1 & 11 & 127 & 0.60\\
\midrule
\multirow{2}{*}{Average}
  & Base     & 243 & 148 & 299 & 146 & 70 & 53 & 48 & 41 & 43 & 34 & 43 & 26 & 22 & 116 & 0.73\\
  & Instruct & 13 & 341 & 343 & 92 & 97 & 23 & 29 & 39 & 20 & 14 & 33 & 3 & 14 & 149 & 0.65\\

\bottomrule
\end{tabular}
\caption{Comparison of named base and instruct models over English questions (data used for Figure \ref{fig:base_vs_ins_hm}). NA: denotes missing responses. Referenced countries: Japan (JP), USA (US), India (IN), China (CN), France (FR), Italy (IT), United Kingdom (GB), Germany (DE), Mexico (MX), Brazil (BR), Russia (RU), Egypt (EG). Bold values indicate the highest count for each country per model. Div: diversity scores. Ent: normalized entropy values.}
\label{tab:base_instruct_models_anex}
\end{table*}

 % Base vs Ins

Table \ref{tab:base_instruct_models_anex} presents a comparative analysis of cultural leanings exhibited by base and instruction-tuned Open Weighted LLMs when responding to open-ended cultural questions in English. Responses are categorized by their alignment with references to specific countries, allowing us to examine how pretraining and instruction tuning affect the geographic distribution of model outputs. We additionally report diversity (Div) and normalized entropy (Ent) scores to quantify the spread and balance of cultural references.

\paragraph{Base models exhibit broader cultural distributions.}
Across all model families, base models demonstrate relatively more balanced distributions of cultural references. While the United States is frequently the most represented region, other countries (particularly Japan, India, China, and several European nations) appear with substantial frequency. This pattern is reflected in higher normalized entropy values, indicating a comparatively more diverse set of cultural alignments than their instructed counterparts. 

\paragraph{Instruction tuning induces strong cultural concentration.}
In contrast, instruction-tuned variants exhibit a pronounced shift toward a narrow set of cultural references. Across all examined model families, instruction tuning results in a sharp increase in references to Japan and the United States, often making these the two dominant regions by a large margin. On average, references to Japan and the US increase in instruction-tuned models, while references to India, China, Russia, Egypt, and Latin American countries decline substantially. This concentration is accompanied by a consistent reduction in normalized entropy, indicating a loss of cultural diversity. Notably, this trend holds across different architectures and parameter scales, including models developed in non-Western contexts (e.g., Qwen with China), suggesting that instruction tuning exerts a homogenizing effect that overrides differences introduced during pretraining.

\paragraph{Diversity metrics corroborate qualitative trends.}
The observed redistribution is quantitatively supported by the diversity and entropy metrics. While diversity scores (Div) increase for instruction-tuned models due to higher total counts concentrated in fewer regions, entropy consistently decreases, reflecting a more peaked and less balanced distribution. 

\paragraph{Possible implications.}
Taken together, these results demonstrate that instruction tuning systematically reduces cultural diversity in model outputs, steering responses toward a limited set of culturally dominant perspectives, particularly those associated with the United States and Japan. This finding has important implications for the deployment of instruction-tuned models in cross-cultural or global applications, where preserving diverse cultural viewpoints may be critical.

\section{Models' Analysis Across Post-training Stages}
\label{sec:olmo-anex}

\begin{table*}[t]
\centering
\small
\setlength{\tabcolsep}{4pt}
\begin{tabular}{llrrrrrrrrrrrrrrr}
\toprule
&&& \flag{jp} & \flag{us} & \flag{in} & \flag{cn} & \flag{fr} & \flag{it} & \flag{gb} & \flag{de} 
& \flag{mx} & \flag{br} & \flag{ru} & \flag{eg} \\
\textbf{Model} & & \textbf{NA} & JP & US & IN & CN & FR & IT & GB & DE & MX & BR & RU & EG & \textbf{Div} & \textbf{Ent}\\
\midrule
\multirow{3}{*}{Olmo-3-7B}
 & Base  & 608 & 35 & \textbf{186} & 43 & 27 & 25 & 18 & 28 & 13 & 12 & 11 & 7  & 9 & 114 & 0.74 \\
 & SFT   & 85 & \textbf{395} & 212 & 97 & 26 & 109 & 64 & 39 & 30 & 28 & 22 & 26 & 10 & 145 & 0.66 \\
 & Inst. & 41 & \textbf{478} & 171 & 89 & 19 & 86 & 78 & 43 & 39 & 44 & 11 & 27 & 5 & 124 &  0.64 \\
  \midrule
\multirow{3}{*}{Olmo-3-7B-Think}
 & Base  & 608 & 35 & \textbf{186} & 43 & 27 & 25 & 18 & 28 & 13 & 12 & 11 & 7  & 9 & 114 & 0.74 \\
 & SFT   & 88  & \textbf{453} & 213 & 105 & 40 & 18 & 32 & 32 & 22 & 32 & 22 & 2 & 13 & 121 & 0.66 \\
 & Inst. & 200 & \textbf{386} & 289 & 179 & 52 & 32 & 46 & 44 & 36 & 47 & 29 & 2 & 25 & 127 & 0.68 \\
 \midrule
\multirow{3}{*}{OLMo-2-7B}
 & Base  & 109 & 64 & \textbf{97} & 84 & 76 & 46 & 57 & 38 & 35 & 30 & 29 & 26 & 35 & 102 & 0.84 \\
 & SFT   & 252 & 130 & \textbf{520} & 67 & 29 & 54 & 40 & 52 & 13 & 18 & 35 & 4 & 13 & 151 & 0.63 \\
 & Inst. & 53 & 163 & \textbf{566} & 79 & 34 & 65 & 62 & 31 & 20 & 20 & 29 & 11 & 18 & 158 & 0.64 \\
  \midrule
\multirow{3}{*}{OLMo-2-32B}
 & Base  & 155 & 61 & \textbf{216} & 119 & 54 & 38 & 44 & 34 & 34 & 28 & 26 & 17 & 9 & 88 & 0.77 \\
 & SFT   & 33 & 361 & \textbf{390} & 51 & 20 & 62 & 36 & 34 & 29 & 16 & 31 & 2 & 16 & 151 & 0.65 \\
 & Inst. & 44 & 221 & \textbf{368} & 81 & 32 & 74 & 40 & 42 & 33 & 18 & 29 & 8 & 6 & 161 & 0.71 \\
\bottomrule
\end{tabular}
\caption{Comparison of Base, SFT, and Instruct variants over English questions. 
NA: denotes missing responses. Referenced countries: Japan (JP), USA (US), India (IN), China (CN), France (FR), Italy (IT), United Kingdom (GB), Germany (DE), Mexico (MX), Brazil (BR), Russia (RU), Egypt (EG). Bold values indicate the highest count for each country per model. Div: diversity scores. Ent: normalized entropy values.}
\label{tab:base_sft_instruct_models_anex}
\end{table*}

Table \ref{tab:base_sft_instruct_models_anex} reports a controlled comparison between base, supervised fine-tuned (SFT), and instruction-aligned (Inst.) variants of open-weight models that provide multiple alignment stages. The analysis focuses exclusively on English prompts, enabling us to isolate the effect of alignment from multilingual prompting effects. By including OLMo-2 (7B, 32B) and OLMo-3 (7B, 7B-Think), which expose intermediate SFT checkpoints, the table allows us to disentangle the impact of supervised fine-tuning from that of final instruction alignment.

\paragraph{Base models: higher entropy and weaker geographic priors.}
Across all architectures and scales, base models exhibit the highest entropy values and comparatively balanced country distributions. While some geographic preferences are already present (notably toward the US), these preferences remain moderate and are spread across a broader set of countries. For example, OLMo-2-7B Base achieves the highest entropy in the table, indicating minimal concentration on any single country. However, base models also show higher rates of missing or underspecified responses (NA), particularly in OLMo-3, reflecting lower controllability rather than alignment-induced bias.

\paragraph{Supervised fine-tuning sharply amplifies cultural bias.}
The transition from Base to SFT introduces the largest shift in geographic skew across all models. SFT variants consistently display dramatic increases in counts for a small number of dominant countries (United States and Japan) accompanied by substantial drops in normalized entropy. For instance, OLMo-2-7B SFT increases US-associated outputs, while entropy drops. Similar patterns appear in both OLMo-3 variants, where Japan becomes the dominant country after SFT. These results indicate that supervised alignment data strongly reinforces cultural priors present in instruction datasets, amplifying bias beyond what is observed in pretraining alone.

\paragraph{Instruction tuning moderates but does not eliminate alignment bias.}
Instruction-tuned models partially attenuate the extreme peaks introduced during SFT, but do not recover the diversity of base models. In several cases, entropy remains close to or below SFT levels, suggesting that dominant-country preferences become stabilized during alignment. Overall, instruction tuning acts as a smoothing mechanism rather than a corrective one.

\paragraph{Diversity and entropy: misleading signals.}
While diversity scores (Div) often increase after SFT and instruction tuning, this increase does not correspond to more balanced outputs. Instead, higher diversity frequently co-occurs with lower entropy, reflecting broader but highly uneven country coverage dominated by a few regions.

\section{Prompt Analysis for RQ3}
\label{sec:prompt-analysis-anex}

\subsection{Prompting for Base Models}

Since foundational Large Language Models (LLMs) are originally trained with a self-supervised next-token prediction objective, they are not inherently designed to perform specific QA tasks without guidance. Unlike instruction-tuned models, base models often lack the alignment necessary to interpret a raw query as a command to provide a factual answer. To address this limitation, we employ In-Context Learning (ICL), a method that leverages explicit reasoning over background knowledge to guide the model’s inference process \cite{NEURIPS2020_1457c0d6}. By providing the model with relevant context and demonstrations within the input window, we effectively transform the generative completion task into a structured QA format, allowing the base model to infer the desired output pattern without the need for parameter updates.

\begin{figure}[ht]
    \centering

        \begin{tcolorbox}[left=.5em]
\begin{Verbatim}[fontsize=\footnotesize]

Question: Who wrote '1984'?
Answer: George Orwell.

Question: What is the chemical symbol for 
water?
Answer: H2O.

Question: What is the dominant religion 
(in region/place)?
Answer: I'll choose {Country1}. The dominant 
religion in {Country1} is {Country1_Religion}.

Question: What sacred texts are studied 
(in region/place)?
Answer: I'll choose {Country2}. In {Country2}, 
the sacred texts studied depend primarily on 
the religious community, as the country has 
a diverse religious landscape.

Question: {Input Question}
Answer:
\end{Verbatim}
    \end{tcolorbox}
    \caption{Prompt given to the Base models in order to follow instructions.}
    \label{fig:base_prompt}
\end{figure}

We operationalize this approach using the prompt structure illustrated in Figure \ref{fig:base_prompt}. To ensure the model adheres to the correct output format, the prompt includes both general QA demonstrations and task-specific examples that explicitly condition the model to respond with a country or geographic region. The template is constructed with strategic placeholders to facilitate this: two slots are reserved for preselected countries—including a specific attribute for religion in the first instance to demonstrate cultural reasoning, while a final placeholder is designated for the target culture-related question. This few-shot setup serves to constrain the model's generation, ensuring it focuses on the relevant cultural entity.

To maintain a rigorous evaluation, we implement a post-processing filter once the judge model extracts the mentioned regions or countries from the output. We specifically identify instances where the LLM simply replicates the example regions provided in the few-shot prompt rather than generating an original response to the cultural query. These cases are classified as "Non-Answered," as they represent a failure of the model to generalize beyond the immediate context. Consequently, all reported counts and statistics in our results are derived exclusively from countries that were not introduced as examples in the prompt, ensuring our findings reflect the model's internal cultural associations.

\begin{figure}[t]
\centering
    \includegraphics[width=0.325\linewidth]{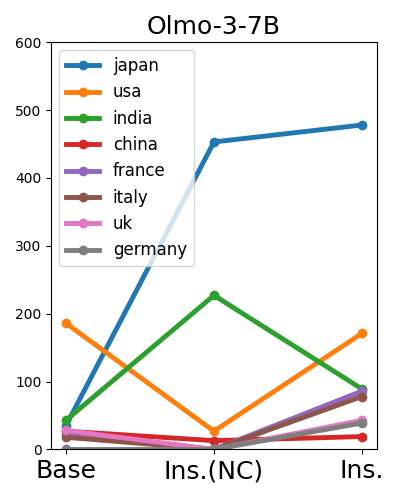}
    \includegraphics[width=0.325\linewidth]{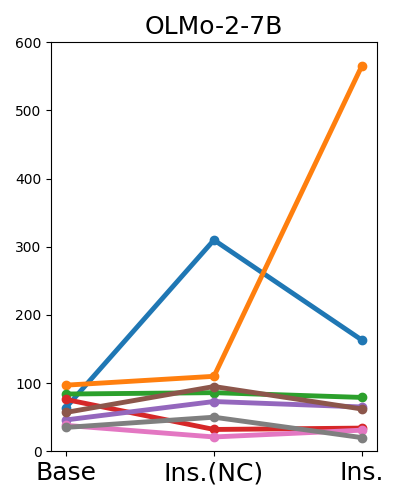}
    \includegraphics[width=0.325\linewidth]{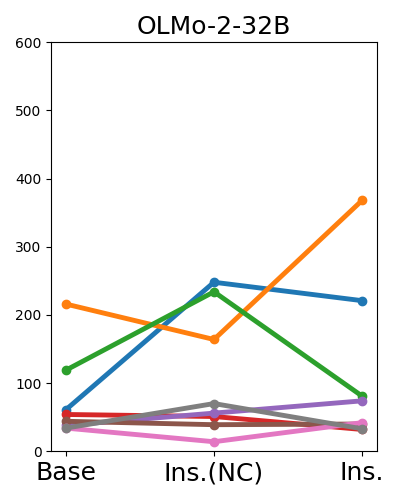} \hfil 
    \caption{Outputs from the Base and Instruct variants of \textit{Olmo3-7B}, \textit{Olmo2-7B} and \textit{Olmo2-32B} models. The Instruct models are shown in two modes: \textit{Ins.(NC)}, which uses the same prompt as the Base model, and \textit{Ins.}, which uses the standard chat template.}
    \label{fig:prompt_analysis} \hfil
\end{figure}

\begin{table*}[h]
\centering
\scriptsize
\setlength{\tabcolsep}{4pt}
\begin{tabular}{llrrrrrrrr}
\toprule
\textbf{Model Family} & \textbf{Post-Training Stage / Dataset} & \textbf{JP} & \textbf{US} & \textbf{IN} & \textbf{CN} & \textbf{FR} & \textbf{IT} & \textbf{GB} & \textbf{DE} \\
\midrule
\multirow{3}{*}{OLMo-2} 
 & SFT (\emph{tulu-3-sft-olmo-2-mixture}) & 8,816 & \textbf{10,188} & 6,046 & 8,123 & \textbf{9,668} & 5,407 & 6,655 & 4,235 \\
 & DPO (\emph{olmo-2-1124-7b-preference-mix}) & 7,389 & \textbf{7,839} & 4,591 & 7,140 & \textbf{8,110} & 3,986 & 4,742 & 4,147 \\
& DPO (\emph{olmo-2-0325-32b-preference-mix}) & 14,754 & \textbf{15,660} & 9,090 & 14,192 & \textbf{16,128} & 7,957 & 9,525 & 8,227 \\
 & RLVR (\emph{RLVR-GSM}) & 252 & \textbf{466} & 261 & 331 & \textbf{622} & 250 & 209 & 242 \\
\midrule

\multirow{3}{*}{OLMo-3-7B-Think} 
& SFT (\emph{Dolci-Think-SFT-7B}) & 21,009 & 10,261 & 22,002 & \textbf{133,953} & \textbf{29,418} & 12,408 & 19,987 & 9,525 \\
& DPO (\emph{Dolci-Think-DPO-7B}) & \textbf{7,837} & 2,315 & 2,588 & \textbf{4,399} & 3,478 & 1,595 & 1,921 & 1,441 \\
& RLVR (\emph{Dolci-Think-RL-7B}) & 743 & 298 & 703 & \textbf{1,359} & \textbf{903} & 443 & 477 & 373 \\
\midrule

\multirow{3}{*}{OLMo-3-7B} 
 
& SFT (\emph{Dolci-Instruct-SFT}) & 20,883 & 24,325 & 17,243 & \textbf{52,509} & 26,091 & 12,153 & 41,682 & \textbf{42,349} \\
& DPO (\emph{Dolci-Instruct-DPO}) & 3,031 & \textbf{7,011} & 2,753 & 3,684 & \textbf{4,006} & 2,152 & 2,874 & 1,687 \\
& RLVR (\emph{Dolci-Instruct-RL}) & 338 & \textbf{365} & 194 & \textbf{386} & 298 & 204 & 200 & 120 \\

\bottomrule
\end{tabular}
\caption{Frequency of explicit country mentions across the SFT, DPO, and RLVR datasets for OLMo-2 and OLMo-3 models available instructions. Referenced countries: Japan (JP), USA (US), India (IN), China (CN), France (FR), Italy (IT), United Kingdom (GB), Germany (DE). \textbf{Bold}: Two highest counts per dataset}
\label{tab:dataset_country_counts}
\end{table*}

\subsection{Impact of the prompt}

To assess the role of prompt formulation in eliciting instruction-following behaviour, we conduct controlled experiments in which prompts designed for base models are applied to instruction-tuned models. We examine whether such prompt mismatches lead to systematic differences in generation, focusing on the instruction-tuned variants of OLMo-2 (7B, 32B) and OLMo-3 (7B).

To evaluate how prompting affects Base versus Instruct models, we prompt the Instruct variants with the same input as the Base models (\textit{Ins.(NC)}) across \textit{Olmo3-7B}, \textit{Olmo2-7B} and \textit{Olmo2-32B} variants (see Figure \ref{fig:prompt_analysis}). The 32B model consistently shows more stable, yet also more biased, behaviour than the 7B models, suggesting greater robustness of bigger models to prompt-formatting differences. Across settings, the USA and Japan frequently receive higher values across all settings, reflecting biases similar to those observed in prior models. Overall, while the prompt can influence responses, the main conclusion holds: cultural biases remain largely consistent, particularly toward Japan, the USA, and India, reinforcing the hypothesis that such biases stem from the models’ training data.

\section{Surface Analysis of the instructions}
\label{sec:instruction-analysis-anex}

To understand how post-training instructions affect cultural bias, we investigated whether a model's tendency to reference specific countries correlates with the geographic distribution of its training data. An initial qualitative analysis using OLMo-trace \cite{liu2025olmotrace} yielded no significant insights. Therefore, we conducted a broader, quantitative experiment leveraging all publicly available post-training data for the OLMo-2 and OLMo-3 model families \cite{olmo2,olmo3}. Using a regular-expression-based approach, we quantified the raw appearances of our top eight countries across the entire post-training dataset (Table~\ref{tab:dataset_country_counts}). This alignment allows us to examine whether country distributions in the instruction-tuning data are associated with regional patterns in the model responses (Table~\ref{tab:base_sft_instruct_models_anex}).

For the OLMo-2 models, the data reveals an inconsistent relationship between dataset frequency and model outputs. A direct correlation is observable regarding the United States, which maintains the highest (or near) representation across the SFT and DPO datasets, such as 15,660 mentions in the 32B preference mix, and consistently emerges as the most referenced region in both the SFT and Instruct model outputs. However, this correlation breaks down entirely for other highly represented regions. France exemplifies this disconnect, as it is frequently the most heavily represented country (of our selection) within the OLMo-2 instruction sets, even surpassing the US in the 7B and 32B DPO sets. Although France appears frequently in the training data, it is underrepresented in the model-generated outputs (Table~\ref{tab:base_sft_instruct_models_anex}), yielding only 74 mentions in the OLMo-2-32B Instruct variant compared to 368 for the US and 65 vs 566 in OLMo-2-7B. Conversely, Japan shows a distinct increase in generation behaviour, consistently ranking as the second most referenced country in the final outputs despite having only a modest presence in the post-training datasets, where it consistently appears less frequently than France and similar to China.

This divergence between dataset composition and generation behaviour is particularly evident in the OLMo-3 models. Within the SFT datasets, China accounts for the largest number of mentions, reaching 133,953 in \textit{Dolci-Think-SFT-7B} and 52,509 in \textit{Dolci-Instruct-SFT-7B}, exceeding the counts for any other country in the analysis. However, this prominence is not reflected in the model outputs, with China receiving only 26 mentions in the SFT variant and 19 in the Instruct variant, as in the generation outputs. At the same time, Japan and India become substantially more prominent in generated content. Although both countries appear relatively infrequently in the OLMo-3 post-training datasets and rank among the lower-frequency countries in several training stages, they receive some of the highest levels of representation in model outputs. For example, in OLMo-3-7B Instruct, Japan records the highest output count (478 mentions), considerably exceeding the US count (171 mentions), despite the US appearing more frequently in the DPO data and both China and several European countries, including Germany and the UK, appearing more frequently in the SFT data.

Overall, these empirical results demonstrate that raw frequency count within post-training instructions are not sufficient to explain the cultural and regional biases observed in model outputs. While certain heavy priors (like the US in OLMo-2) appear to persist, massive influxes of specific country data (such as China in OLMo-3) can be entirely attenuated during training, while countries (such as Japan) can become amplified. We must conclude by noting that this experiment did not clarify the main \textbf{\textit{Why}} question of the paper, which continues to remain open for future investigation.

\end{document}